\documentclass{article}
\usepackage{iclr2027_conference,times}
\usepackage{microtype}
\usepackage{amsmath,amssymb,mathtools,amsthm}
\usepackage{booktabs}
\usepackage{siunitx}
\usepackage{textcomp}
\usepackage{array}
\usepackage{longtable}
\usepackage{graphicx}
\usepackage{placeins}
\usepackage{float}
\usepackage{multirow}
\usepackage{xcolor}
\usepackage{hyperref}
\usepackage{url}
\usepackage{enumitem}
\hypersetup{hidelinks,pdftitle={Same Winners, Different Success Rates: Evaluating How LLM Agents Recover from Failures},pdfauthor={Dong Xu, Zhangfan Yang, Jiantao Wu, Shipeng Zhang, Jiayu Zhang, Zexuan Zhu, Jiangjiang Li, Jun Zhang, Junkai Ji},pdfsubject={Evidence sufficiency for agent checkpoint recovery}}
\iclrfinalcopy

\newcommand{\prefcert}{\textsc{Preference-Cert}}

\definecolor{orchblue}{HTML}{2F5D7C}
\definecolor{orchteal}{HTML}{5B8D78}
\definecolor{orchgold}{HTML}{D09A2B}
\definecolor{orchcoral}{HTML}{B85F49}

\title{Same Winners, Different Success Rates: Evaluating How LLM Agents Recover from Failures}

\author{\textbf{Dong Xu}$^{1,2}$, \textbf{Zhangfan Yang}$^3$, \textbf{Jiantao Wu}$^1$, \textbf{Shipeng Zhang}$^1$, \textbf{Jiayu Zhang}$^1$, \textbf{Zexuan Zhu}$^1$,\\[-1pt]
\textbf{Jiangjiang Li}$^1$, \textbf{Jun Zhang}$^1$, \textbf{Junkai Ji}$^{1,2}$\\[-1pt]
$^1$\textnormal{School of Artificial Intelligence, Shenzhen University}\\
$^2$\textnormal{EasternDawn}\\
$^3$\textnormal{School of Computer Science, University of Nottingham Ningbo}}

\date{}

\begin{document}
\maketitle
\lhead{Preprint}

\begin{abstract}
Evaluating how LLM agents recover from mid-task failures is central to deploying reliable agentic systems. Existing checkpoint-based benchmarks measure recovery by comparing which action is selected as best across independent runs, a quantity known as set agreement. However, set agreement is a purely ordinal measure that records which action ``wins'' without reflecting the absolute level of performance. When all actions fail, they tie at zero reward, and independent runs produce the same tied set with high probability, creating an illusion of stability that masks near-zero recovery success.
We formalize this limitation through a \emph{set-path symmetry} result, proving that for equal-cost Bernoulli actions the success probabilities $(0.9,0.8)$ and $(0.2,0.1)$ yield identical best-action-set distributions at every sample size. No procedure based solely on which action wins can distinguish these two regimes. We further prove that certifying exact population ties is impossible in finite time, and that the assignment of outcomes to checkpoints carries information beyond marginal outcome distributions. The pooled success probability is the missing scalar that resolves the ordinal ambiguity.
Experiments on 864 frozen RecoveryBench episodes and two planning cohorts totaling 3{,}456 responses confirm the theoretical predictions. Agreement and held-out quality can move in opposite directions, and permuting checkpoint-to-action bindings changes 8 to 13\% of cell-level conclusions. Based on these findings, we propose reporting four diagnostic quantities (agreement, all-zero fraction, held-out success, and pooled success) that expose this failure mode with no additional data collection.
\end{abstract}

\section{Introduction}\label{sec:introduction}

As LLM-based agents are deployed in increasingly complex workflows, the ability to recover from mid-task failures has become a critical capability. When an agent fails at an intermediate step, a workflow may retry the operation, verify the current result, or replace the agent entirely. Evaluating these recovery strategies requires benchmarks that can reliably identify which action leads to successful task completion. Recent work has introduced checkpoint-based benchmarks such as RecoveryBench~\citep{tan2025recoverybench} and ToolMaze~\citep{zhu2026toolmaze} that replay failed agent states and measure recovery outcomes under controlled conditions.

A widely used stability metric in these benchmarks is \emph{set agreement}, the frequency with which independent runs select the same best action. High agreement is typically interpreted as evidence that the evaluation is stable and the selected action is reliably superior. However, this interpretation assumes that agreement reflects action quality. We show that it does not, because set agreement conflates ordinal rankings with cardinal performance, and the same high agreement can arise from universal success or universal failure.

The core problem is that set agreement is a purely ordinal measure. It records which action ``wins'' but not the absolute level of performance. When all actions fail with near-certainty, they tie at zero reward, and independent runs produce the same tied set with high probability. The resulting agreement is an artifact of shared failure, not a signal of quality. We formalize this observation through a \emph{set-path symmetry} result (Proposition~\ref{prop:set-path-symmetry}). For two equal-cost Bernoulli actions, the parameters $(0.9, 0.8)$ and $(0.2, 0.1)$ produce identical distributions over the observed best-action set at every sample size. No procedure based solely on which action ``wins'' can distinguish these two regimes. The missing information is a single scalar, the pooled success probability, which requires observing actual outcomes. In addition, we prove that certifying exact population ties is impossible in finite time (Eq.~\eqref{eq:exact-tie-release-bound}), and that even when outcomes are available, the assignment of outcomes to checkpoints carries information beyond marginal distributions. Permuting checkpoint-to-action bindings while preserving every action's outcome multiset changes 8 to 13\% of cell-level conclusions.

We confirm this disconnect empirically across two regimes. On RecoveryBench, we evaluate 864 frozen episodes across three GPT-5.6 models (Luna, Terra, Sol) and three recovery actions (retry, verify, replace), with eight draws per action. Agreement reaches 85.4\% at $k{=}1$, yet held-out success is only 1.4\%, with the entire agreement driven by all-zero comparisons. Selecting models by agreement attains only 1.3\% held-out success, compared to 6.5\% when selecting by mean outcome. In two planning cohorts totaling 3{,}456 responses across 24 frozen checkpoints, agreement drops five-fold as the sample budget grows from 1 to 4 per action, while held-out success remains stable at approximately 76\%. Decomposing agreement by outcome quality reveals that the decline is driven entirely by tie resolution. Multi-action matches account for 0.310 of 0.318 total agreement at $k{=}1$ and fall to 0.125 at $k{=}4$, while singleton agreement rises only slightly. A separate panel of Qwen3-8B across rank, path, and filter tasks confirms that the pattern generalizes beyond the GPT-5.6 family.

We additionally validate the Hoeffding certificate for statistical certification of best actions, showing that certification is practical only when margins are large (97.9\% at 128 branches for a 0.40 gap) but remains below 0.1\% even at 512 branches when the margin is 0.06. At the margins typically observed in agent recovery experiments (5 to 10 percentage points), certifying a unique best action would require hundreds or thousands of executions per action, far beyond current evaluation budgets.

Our contributions are as follows.
\vspace{-2mm}
\begingroup
\setlength{\leftmargini}{0.5\leftmargini}
\begin{itemize}
\item \textbf{Identification theory.} We prove that best-action-set sequences identify the preference gap but not the absolute success probabilities (a property we call \emph{set-path symmetry}). The pooled success probability resolves this ambiguity. We also prove that certifying exact population ties is impossible in finite time, and characterize the identified set as a two-element parameter family (Eq.~\eqref{eq:set-path-identified-set}).
\item \textbf{Empirical evidence.} We evaluate set agreement alongside held-out success on 864 frozen RecoveryBench episodes and two planning cohorts (3{,}456 responses across GPT-5.6 and Qwen3-8B models), showing that agreement and quality can move in opposite directions. Agreement decomposition, tolerance sensitivity, and cross-task analyses confirm that tie structure, not action quality, drives the observed agreement levels.
\item \textbf{Record provenance.} We show that checkpoint-to-action bindings carry information beyond marginal outcome distributions. Permuting these bindings while preserving every action's outcome multiset changes 8 to 13\% of cell-level conclusions, an error rate that exceeds typical measurement noise and is undetectable from outcome data alone.
\item \textbf{Practical recommendations.} We propose reporting four diagnostic quantities (agreement, the fraction of all-zero comparisons, held-out success, and pooled success) that make the failure mode visible with no additional data collection. We provide statistical certification bounds and demonstrate their sample requirements under realistic margins.
\end{itemize}
\endgroup

\section{Related Work}\label{sec:related}

\noindent\textbf{Agent recovery.}
ReAct~\citep{yao2023react}, Reflexion~\citep{shinn2023reflexion}, and AutoGen~\citep{wu2023autogen} provide agent architectures for reasoning, self-improvement, and multi-agent coordination.
AgentBench~\citep{liu2024agentbench} and WorkflowLLM~\citep{fan2025workflowllm} evaluate agents across environments. Replay and recovery studies~\citep{gonuguntla2026replaygap,shah2026causalreplay,qi2026r2act} examine interventions from partial executions. Execution-edit safety work studies checkpoint, fork, restore, and merge semantics~\citep{zheng2026executionedits}.
RecoveryBench~\citep{tan2025recoverybench} derives model comparisons from recovery success on replayed failed states. We show that its set-agreement readout does not distinguish successful recovery from shared failure.

\noindent\textbf{Identification and best-arm selection.}
Partial identification~\citep{manski2003partial,christensen2022partial} represents targets compatible with observations, and set-valued decisions~\citep{fuentes2026setvalued} retain unresolved alternatives.
Fixed-confidence best-arm identification~\citep{garivier2016optimal,kaufmann2013subset} returns optimal arms, while multiple-answer formulations~\citep{degenne2019multiple} permit any correct answer.
Our setting differs because the maximizer set's cardinality can change under arbitrarily small perturbations, yielding finite-time impossibility of certifying exact ties (Eq.~\eqref{eq:exact-tie-release-bound}).
Selective prediction~\citep{geifman2017selective,liu2026agentabstain} allows abstention, and our certification rule shares this spirit.
EC$^2$~\citep{golovin2010noisy} serves as our retrieval baseline.

\noindent\textbf{Repeated evaluation and provenance.}
Stable Ranking Probability~\citep{riley2024replicable} and repeated LLM evaluations~\citep{alvarado2025repetitions} motivate reporting stability and quality separately.
Provenance methods~\citep{wang2026cava,liao2026auditingprovenance,zhu2026claimreceipt} connect actions to supporting records. We add the formal association between outcome laws and checkpoint assignments.
Appendix~\ref{sec:positioning-appendix} compares targets, observations, and assumptions across these settings.

\section{Checkpoint Setting and Decision Targets}\label{sec:problem}

\begin{figure}[t]
\centering
\includegraphics[width=\linewidth]{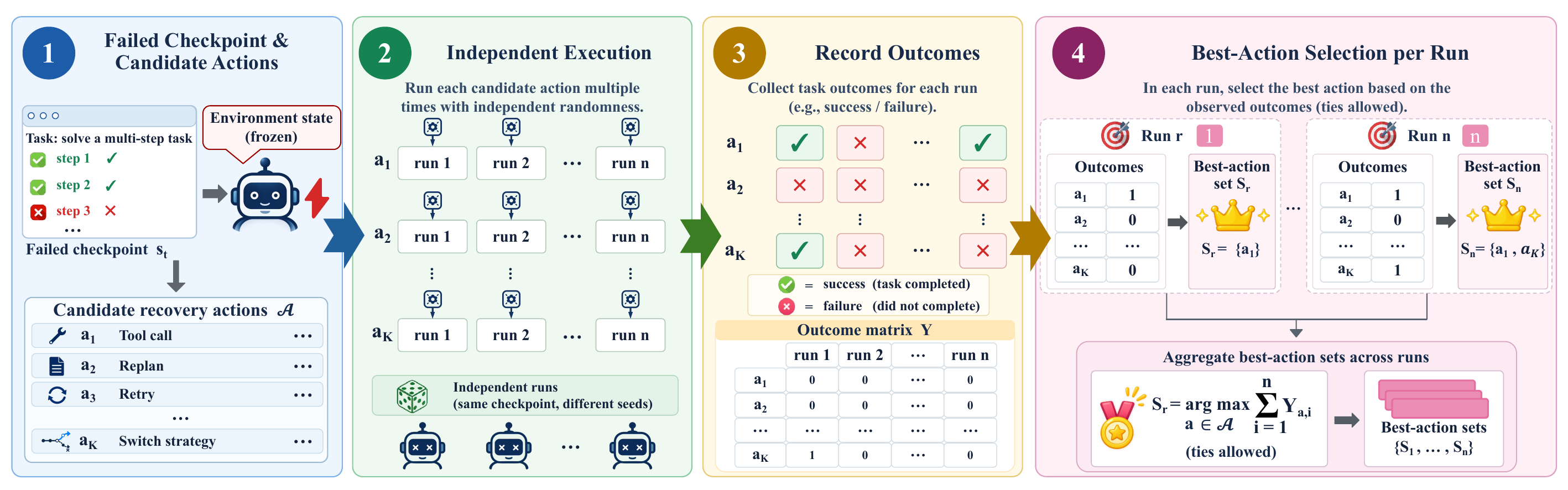}
\caption{Overview of the checkpoint evaluation pipeline.  \textbf{Step~1.} A failed checkpoint state is frozen and paired with multiple candidate recovery actions (e.g., tool call, replan, retry, switch strategy).  \textbf{Step~2.} Each action is executed multiple times independently from the same checkpoint with different random seeds.  \textbf{Step~3.} Binary outcomes (success or failure) are recorded for every action--run pair, forming an outcome matrix.  \textbf{Step~4.} For each run, the best-action set is computed by selecting all actions with the highest total score (ties retained), yielding a collection of best-action sets that serves as the basis for agreement and quality analysis.}
\label{fig:overview}
\end{figure}

\noindent\textbf{Setup and notation.}
We formalize the evaluation pipeline shown in Figure~\ref{fig:overview}. Fix a checkpoint state $s$ (e.g., a failed task with its current partial result), a finite set of recovery actions $A_0$, and a scoring contract.
Each action $a$ has cost $C(a)$ weighted by $\lambda\geq0$.
After executing action $a$, we observe a binary outcome. We write $\widehat q_s(a)$ for its sample mean minus the cost penalty $\lambda C(a)$.
A reference model provides the expected score $Q_{\mathrm{sc}}(s,a)$ under the same cost adjustment.
The key sets are
\begin{equation}
R_s = \operatorname{arg\,max}_{a\in A_0} Q_{\mathrm{sc}}(s,a),\qquad
\widehat L_s^{\mathrm{obs}} = \operatorname{arg\,max}_{a\in A_0}\widehat q_s(a),\qquad
I_s^{\mathrm{obs}}=R_s\cap\widehat L_s^{\mathrm{obs}}.
\label{eq:observed-sets}
\end{equation}
Here $R_s$ is the reference-best set, $\widehat L_s^{\mathrm{obs}}$ is the \emph{observed best-action set} (retaining all ties), and $I_s^{\mathrm{obs}}$ is their intersection.
\emph{Exact-set agreement} compares $\widehat L_s^{\mathrm{obs}}$ across independent execution blocks at the same checkpoint. Two blocks agree when they yield the same set.

\noindent\textbf{Expected vs.\ observed preference.}
Let $P$ be the true outcome law for binary success $\overline Y_{s,a}\in\{0,1\}$.
The \emph{expected} best-action set is
\begin{equation}
A^\star(P)=\operatorname{arg\,max}_{a\in A_0}
\{\mathbb E_P[\overline Y_{s,a}]-\lambda C(a)\}.
\label{eq:preference-overlap}
\end{equation}
The observed set $\widehat L_s^{\mathrm{obs}}$ estimates $A^\star(P)$ but can differ due to sampling noise and ties.
The central question is whether agreement on $\widehat L_s^{\mathrm{obs}}$, the fact that independent runs select the same set of ``winners,'' implies anything about the \emph{quality} of those winners, that is, whether the actions in $A^\star(P)$ actually succeed.

\noindent\textbf{Why ties dominate at small budgets.}
Consider three recovery actions each executed once ($k{=}1$) with binary outcomes.
If all three fail (a common scenario on hard checkpoints), every action scores zero and the observed best-action set is $A_0$ itself.
A second independent run that also produces all failures yields the same set, giving perfect agreement despite zero information about relative action quality.
As $k$ grows, empirical ties become rarer, agreement drops, but action quality is unchanged.
This mechanism, driven by ties, is the main source of inflated agreement in our experiments.

\noindent\textbf{Record provenance.}
Beyond outcomes, the assignment of outcomes to checkpoints matters.
Archived results may be grouped by action without retaining which checkpoint produced which execution.
We show in Section~\ref{sec:pairing-results} that permuting these bindings, while preserving every action's outcome distribution, changes 8 to 13\% of cell-level conclusions.
Formal definitions of record eligibility and the complete audit contract are in Appendix~\ref{sec:formal-method-details}.

\section{Why Set Agreement Alone Is Insufficient}\label{sec:method}

\subsection{Best-action sets do not identify success probabilities}\label{sec:set-path-method}

We formalize the core problem. Knowing which action ``wins'' does not, in general, reveal whether recovery actually works.
Consider two equal-cost recovery actions with independent Bernoulli outcomes $Y_{a,t}\sim\operatorname{Bernoulli}(p_a)$, $p_a\in(0,1)$, across actions $a\in\{1,2\}$ and rounds $t$.
After $n$ rounds, we retain only the observed best-action set (as is standard in set-agreement reporting):
\begin{equation}
S_n=\operatorname{arg\,max}_{a\in\{1,2\}}\sum\nolimits_{t=1}^{n}Y_{a,t},\qquad n\ge1.
\label{eq:set-path-observation}
\end{equation}

The following result shows that the entire sequence $(S_1, S_2, \ldots)$ cannot distinguish between high-quality and low-quality recovery.

\noindent\textbf{Proposition (set-path symmetry).}\label{prop:set-path-symmetry}
Let $P=(p_1,p_2)$ and $P^\dagger=(1-p_2,1-p_1)$.
The full labeled path $(S_n)_{n\ge1}$ has the same distribution under $P$ and $P^\dagger$.
For an action selected based only on this path, its success probability on a fresh execution shifts by $1-p_1-p_2$ between the two laws.

\begin{proof}
Set $Y^\dagger_{1,t}=1-Y_{2,t}$ and $Y^\dagger_{2,t}=1-Y_{1,t}$.
These are independent Bernoulli with law $P^\dagger$ and satisfy $Y^\dagger_{1,t}-Y^\dagger_{2,t}=Y_{1,t}-Y_{2,t}$.
Since cumulative differences are equal at every round, the labeled paths coincide, including ties.
\end{proof}

\noindent\textbf{Example.}
Take $(p_1,p_2)=(0.9,0.8)$ and $(p_1^\dagger,p_2^\dagger)=(0.2,0.1)$.
In both cases, Action~1 is uniquely best with the same gap of 0.1.
Yet $p_1=0.9$ in the first case and $p_1^\dagger=0.2$ in the second.
The minimax estimation error for the best success probability from the path alone is $|1-p_1-p_2|/2$ (Eq.~\eqref{eq:set-path-quality-bound}).
Flipping all outcomes (swapping success and failure) preserves which action is better, since an action that succeeds more often also fails less often. However, this flip transforms a high-success regime into a low-success one.
Any statistic that depends only on relative performance is blind to this flip.

\noindent\textbf{Identified set.}
To characterize all parameters compatible with the path law, let $u=\Pr(S_1=\{1\})$, $v=\Pr(S_1=\{2\})$, $d=u-v$, and $h=\sqrt{1-2(u+v)+d^2}$.
Solving yields exactly two compatible parameter vectors:
\begin{equation}
\Theta(u,v)=\left\{\left(\frac{1+d+\eta h}{2},\frac{1-d+\eta h}{2}\right):\eta\in\{-1,1\}\right\}.
\label{eq:set-path-identified-set}
\end{equation}
Both share the gap $d=p_1-p_2$ and the expected best-action set, but differ by $h=|1-p_1-p_2|$ in each coordinate.

\noindent\textbf{Resolution.}
If the pooled success probability $m=(p_1+p_2)/2$ is known, both means are identified: $p_1=m+d/2$, $p_2=m-d/2$.
Thus the missing information is a single scalar, the overall success level, which requires observing outcomes, not just winners.
Appendix~\ref{sec:set-path-augmentation} treats finite-sample estimation.

\subsection{Certification limits: exact ties and separated margins}\label{sec:expected-preference}

Even with outcomes in hand, certifying the expected best-action set faces fundamental limits.

\noindent\textbf{Exact ties cannot be certified.}
If two or more actions are exactly tied at the population level, an arbitrarily small perturbation changes the set.
Any procedure that controls false certification at level $\delta$ under all admissible laws can certify such a tie in finite time with probability at most $\delta$, even with unlimited sampling (Eq.~\eqref{eq:exact-tie-release-bound}; proof in Appendix~\ref{sec:exact-tie-boundary}).

\noindent\textbf{Unique winners can be certified with large margins.}
When a unique best action exists, meaning one action's expected score strictly exceeds all others, it can in principle be certified from finite data.
For $n$ executions per action, simultaneous Hoeffding bounds~\citep{hoeffding1963} yield a confidence set.
Let $\widehat q_n(a)$ be the sample mean of action $a$ minus its cost penalty $\lambda C(a)$.
For $|A_0|\ge2$ actions and error level $0<\delta<1$:
\begin{equation}
\rho_n(\delta)=\sqrt{\frac{\log(2|A_0|/\delta)}{2n}},\quad
W^{\mathrm{rob}}_{\delta,n}=
\left\{a\in A_0:
\widehat q_n(a)>\max\nolimits_{b\ne a}\widehat q_n(b)+2\rho_n(\delta)
\right\}.
\label{eq:finite-certificate}
\end{equation}
With probability ${\ge}1{-}\delta$, any action in $W^{\mathrm{rob}}_{\delta,n}$ is the unique expected best action.
The rule abstains when $W^{\mathrm{rob}}_{\delta,n}$ is empty, which happens whenever margins are too small relative to $n$.
In practice, a gap of 0.40 reaches near-complete certification at $n=128$, but a gap of 0.06 remains unresolved at $n=512$ (Section~\ref{sec:budget-results}).

\subsection{Record assignments change checkpoint conclusions}\label{sec:paired}

A checkpoint comparison depends not only on outcomes but on which outcomes are assigned to which checkpoints.
Archived results may be grouped by action without retaining their checkpoint bindings.
This creates ambiguity even when every action's outcome multiset is fully known.

\noindent\textbf{How pairing matters.}
Suppose two outcome laws $P^+, P^-$ have distinct best-action sets, and two execution records $z^+, z^-$ have different eligibility (one passes provenance checks, the other does not).
The same laws and records can be paired in two ways: \emph{diagonally} ($P^+$ with $z^+$, $P^-$ with $z^-$) or \emph{crosswise} ($P^+$ with $z^-$, $P^-$ with $z^+$).
Both pairings are consistent with the same marginal observations, but they yield opposite conclusions about which best-action set is backed by an eligible record.
The assignment of outcomes to checkpoints therefore carries information beyond what marginal outcome distributions contain. We test this prediction experimentally in Section~\ref{sec:pairing-results}.
The formal construction (compatible law--record relations and their target images) is in Appendix~\ref{sec:formal-method-details}.

\noindent\textbf{Statistical certification.}
For execution blocks without checkpoint--action identifiers, we construct simultaneous confidence regions using Clopper--Pearson intervals~\citep{clopper1934} with a Bonferroni correction~\citep{bonferroni1936} over the action set.
The certified best-action set is the intersection of the confidence region with the set of possible maximizers (Appendix~\ref{sec:statistical-release}).

\section{Experiments}\label{sec:results}\label{sec:experimental-design}

We evaluate exact-set agreement alongside held-out success on a frozen RecoveryBench panel of failed checkpoints and two planning cohorts.
Execution records come from Qwen3-8B~\citep{yang2025qwen3} and the GPT-5.6 family~\citep{openai2026gpt56} (Luna, Terra, Sol).
The central goal is to determine whether high set agreement predicts successful recovery.
All experiments use frozen checkpoints in which the agent state is replayed from a saved snapshot, ensuring that the only source of variation is the recovery action and decoding randomness.
Each experiment splits draws into a selection block (used to determine the best-action set) and a held-out block (used to evaluate the selected action's quality).
Full protocols, statistical units, and uncertainty analyses are provided in Appendix~\ref{sec:experimental-details}.

\subsection{RecoveryBench: High Agreement, Low Success}\label{sec:recoverybench-main}

We apply the complete-set readout to 12 frozen failed Terminal-Bench~\citep{merrill2026terminalbench} checkpoints from RecoveryBench~\citep{tan2025recoverybench}.
Each checkpoint is evaluated with three GPT-5.6 models and the actions \texttt{retry}, \texttt{verify}, and \texttt{replace\_agent}, with eight repeats per action (864 total episodes).
Repeats 0--3 form the selection block and repeats 4--7 the held-out block.
For each $k\in\{1,2,4\}$ draws per action, the selection block yields a complete native-reward maximizer set, which is compared with the held-out set.
Success is defined as native verifier reward ${>}0$.

\begin{table}[ht]
\centering\small
\caption{RecoveryBench validation on 12 failed checkpoints. Each model row reports agreement, all-zero pairs, and held-out success for $k=1,2,4$ draws per action. Agreement compares native-reward maximizer sets across blocks; all-zero pairs is the fraction of matching comparisons where every action scores zero in both blocks. Held-out success evaluates the selected set; intervals are in Appendix Table~\ref{tab:recoverybench-intervals}.}
\label{tab:recoverybench-validation}
\setlength{\tabcolsep}{3.2pt}
\begin{tabular*}{\linewidth}{@{\extracolsep{\fill}}lrrrrrrrrr@{}}
\toprule
 & \multicolumn{3}{c}{Agreement} & \multicolumn{3}{c}{All-zero pairs} & \multicolumn{3}{c}{Held-out success} \\
\cmidrule(lr){2-4}\cmidrule(lr){5-7}\cmidrule(lr){8-10}
Model & $k=1$ & $k=2$ & $k=4$ & $k=1$ & $k=2$ & $k=4$ & $k=1$ & $k=2$ & $k=4$ \\
\midrule
Luna & 0.854 & 0.750 & 0.667 & 0.854 & 0.750 & 0.583 & 0.014 & 0.014 & 0.014 \\
Terra & 0.776 & 0.688 & 0.583 & 0.776 & 0.688 & 0.583 & 0.045 & 0.035 & 0.014 \\
Sol & 0.734 & 0.627 & 0.417 & 0.708 & 0.590 & 0.417 & 0.083 & 0.075 & 0.063 \\
\bottomrule
\end{tabular*}
\end{table}

At $k{=}1$, Luna's agreement reaches 85.4\% while held-out success is only 1.4\% (Table~\ref{tab:recoverybench-validation}).
The all-zero pairs fraction equals the agreement for Luna at $k{=}1$ (Table~\ref{tab:recoverybench-validation}), indicating that every matching comparison consists entirely of zero-reward attempts.
The models agree perfectly on which action is best because none of them ever succeed.

The pattern holds across all three models and three budgets (Table~\ref{tab:recoverybench-validation}), no cross-block pair has positive reward for every action in both blocks (the all-success frequency is exactly~0).
At the checkpoint level, agreement and held-out success are negatively correlated, with Spearman estimates ranging from $-0.796$ to $-0.443$ at $k{=}1$ (Appendix Table~\ref{tab:recoverybench-rank}).
Checkpoints where recovery sometimes works produce less agreement, because stochastic successes break the universal-failure ties.
Under the assumption that agreement signals quality, high agreement should predict high success, yet the opposite holds, and the checkpoints with the most agreement are those where no action ever succeeds.

\noindent\textbf{Selection by agreement vs.\ by outcome.}
Selecting models and actions by agreement attains only 1.3\% held-out success at $k{=}1$, compared with 6.5\% when selecting by mean outcome (Appendix~\ref{sec:recoverybench-protocol}).
A practitioner who chooses actions by set agreement would therefore perform worse than one who simply picks the highest mean score, even when that mean is low.
Of the 864 episodes, 364 have a native verifier observation, 487 carry a model-protocol failure, and 13 carry an invalid adapter attempt (Appendix Table~\ref{tab:recoverybench-status}).
The prevalence of protocol failures explains why success is so low. The majority of episodes never reach the verifier because the model fails to follow the recovery protocol, yet these failures still generate zero-reward outcomes that contribute to agreement.

\noindent\textbf{Increasing $k$ resolves ties but lowers agreement.}
As $k$ increases from 1 to 4, agreement drops for all three models, e.g., from 0.854 to 0.667 for Luna and from 0.734 to 0.417 for Sol (Table~\ref{tab:recoverybench-validation}).
As the set-path symmetry result predicts, more observations make empirical ties less likely, producing smaller and more variable best-action sets.
Held-out success, however, is essentially unchanged across budgets. Luna stays at 1.4\% regardless of $k$ (Table~\ref{tab:recoverybench-validation}), confirming that the agreement decline reflects tie resolution, not quality degradation.

\subsection{Planning: Agreement Drops While Quality Holds}\label{sec:repeated-main}

The RecoveryBench results demonstrate the disconnect in a regime of near-universal failure. The planning experiments below test whether it persists when actions have moderate to high success rates.

We evaluate two frozen 24-checkpoint planning cohorts with GPT-5.6 Terra, Luna, and Sol.
Each model--cohort panel has three recovery actions and eight draws per action (576 responses per panel, 3{,}456 total across six panels).
The study cohort was designed first, and the replication cohort was drawn independently to test whether findings reproduce.
Draws are split into selection and evaluation blocks. For each $k\in\{1,2,4\}$, all size-$k$ subsets in the selection block form a complete public-score maximizer set.
Fixed verification, selected on a separate development panel, serves as the one-call baseline.
Full protocols are detailed in Appendix~\ref{sec:prospective-protocol}.

\begin{table}[ht]
\centering\small
\caption{GPT-5.6 planning results on two frozen cohorts (point estimates; intervals in Appendix Tables~\ref{tab:prospective-family-intervals} and~\ref{tab:prospective-family-success-intervals}). Agreement compares public-score maximizer sets across blocks. Selected strict success evaluates the chosen action mixture on held-out draws.}
\label{tab:gpt56-family}
\begin{tabular*}{\linewidth}{@{\extracolsep{\fill}}llrrrrrrr@{}}
\toprule
 & & \multicolumn{3}{c}{Set agreement} & \multicolumn{3}{c}{Selected strict success} & \\
\cmidrule(lr){3-5}\cmidrule(lr){6-8}
Cohort & Model & $k=1$ & $k=2$ & $k=4$ & $k=1$ & $k=2$ & $k=4$ & Fixed verify \\
\midrule
Study & Terra & 0.318 & 0.204 & 0.167 & 0.699 & 0.695 & 0.710 & 0.771 \\
Study & Luna & 0.234 & 0.150 & 0.083 & 0.575 & 0.568 & 0.538 & 0.635 \\
Study & Sol & 0.424 & 0.331 & 0.167 & 0.722 & 0.729 & 0.750 & 0.729 \\
Replication & Terra & 0.404 & 0.238 & 0.083 & 0.765 & 0.754 & 0.762 & 0.771 \\
Replication & Luna & 0.299 & 0.212 & 0.250 & 0.727 & 0.728 & 0.719 & 0.760 \\
Replication & Sol & 0.497 & 0.422 & 0.458 & 0.808 & 0.814 & 0.825 & 0.844 \\
\bottomrule
\end{tabular*}
\end{table}

Across all six panels, exact agreement drops from $k{=}1$ to $k{=}4$ (Table~\ref{tab:gpt56-family}), with declines ranging from 0.039 to 0.320.
For Terra in the replication cohort, agreement falls from 40.4\% to 8.3\%, a five-fold reduction.
Yet held-out strict success is largely stable. The selected-minus-fixed differences at $k{=}1$ range from $-0.072$ to $-0.006$, with every 95\% bootstrap interval containing zero in the replication cohort.
Agreement is thus primarily tracking tie structure, not action quality.
The pattern also holds across models with different overall capabilities. Sol, the strongest model, shows the lowest agreement at every budget because its higher success rate creates more variation in outcomes, breaking ties that weaker models preserve through shared failure.

\noindent\textbf{Sources of the agreement decline.}
Table~\ref{tab:agreement-decomposition-main} decomposes agreement on the original Terra panel into contributions from singleton matches (both blocks select a unique winner) and multi-action matches (both blocks retain the same tied set).
At $k{=}1$, nearly all agreement comes from multi-action matches (0.310 of 0.318). The blocks agree because most checkpoints produce ties among all three actions.
As $k$ grows, the mean set size shrinks from 2.29 to 1.52, multi-action agreement drops from 0.310 to 0.125, and singleton agreement rises only slightly.
Success is unchanged throughout. The tie resolution does not affect which actions work, only which ones are declared ``best.''

A Bernoulli control confirms the mechanism analytically.
With three equal-cost actions each having success probability 0.9, one-draw agreement is 0.553, and at four draws it falls to 0.183.
Yet expected success remains 0.9.
With success probability 0.1, one-draw agreement is also 0.553, illustrating the set-path symmetry in action (Appendix~\ref{sec:set-agreement-mechanism}).

\begin{table}[ht]
\centering\small
\caption{Decomposition of set agreement on the original 24 planning checkpoints (Terra). Single and Multiple are disjoint contributions to Agreement by the size of the matching sets. Mean size averages both blocks; Success evaluates the selected action on the held-out block.}
\label{tab:agreement-decomposition-main}
\begin{tabular*}{\linewidth}{@{\extracolsep{\fill}}rrrrrr@{}}
\toprule
$k$ & Agreement & Single & Multiple & Mean size & Success\\
\midrule
1 & 0.318 & 0.008 & 0.310 & 2.286 & 0.699\\
2 & 0.204 & 0.039 & 0.164 & 1.913 & 0.695\\
4 & 0.167 & 0.042 & 0.125 & 1.521 & 0.710\\
\bottomrule
\end{tabular*}
\end{table}

The replication cohort independently confirms the agreement decline (Table~\ref{tab:gpt56-family}), with the same pattern of stable success despite falling agreement.
Recomputing benefit without requiring correct self-reported totals preserves the decline in both panels (Appendix~\ref{sec:semantic-set-sensitivity}).
The 12-checkpoint Qwen3-8B panel, spanning rank, path, and filter tasks, shows 84.0\% strict agreement at $k{=}1$ with 37.9\% selected-action success (Appendix Table~\ref{tab:cross-task-set-agreement}). Here too, matching sets include both all-failure and all-success blocks.
In this panel, multi-action matching sets contribute 0.831 of the 0.840 agreement at $k{=}1$, and 58.3\% of agreeing pairs have maximum score zero in both blocks (Appendix~\ref{sec:cross-task-set-agreement}).
Strict successes concentrate in the rank and filter families. The path family contributes none, yet all three families produce high agreement.

\noindent\textbf{The role of task difficulty.}
Across both cohorts, the disconnect between agreement and success is most pronounced in two regimes: (i)~easy tasks where all actions succeed (producing agreement via universal success) and (ii)~hard tasks where all actions fail (producing agreement via universal failure).
Only tasks of intermediate difficulty, where some actions succeed and others fail, generate informative disagreement. These are precisely the cases where agreement would be useful for selecting actions.
Set agreement is therefore least informative where it is needed most.

\noindent\textbf{Empirical near ties.}
A natural attempt to boost agreement is to soften the exact-set comparison by retaining all actions within a tolerance $\epsilon U_s$ of the best mean score, where $U_s$ is the score range.
This converts near-winners into co-winners, absorbing small fluctuations.
Table~\ref{tab:set-tolerance} evaluates this at $k{=}4$ for two tolerances alongside the exact comparison.
The results are mixed. In one cohort agreement increases slightly (from 0.083 to 0.250), but in the other it actually decreases (from 0.167 to 0.125).
Held-out success changes negligibly in both cases.
The tolerance does not reliably improve agreement, and when it does, the improvement reflects the absorption of near-ties rather than a better signal about action quality.
All tolerances and budgets are reported in Appendix~\ref{sec:set-tolerance}.

\begin{table}[ht]
\centering\small
\caption{Empirical tolerance sensitivity at $k=4$ on 24 checkpoints per cohort. Strict is selected-action success on the held-out block; Public is normalized feasible benefit.}
\label{tab:set-tolerance}
\begin{tabular*}{\linewidth}{@{\extracolsep{\fill}}rrrrrrr@{}}
\toprule
 & \multicolumn{3}{c}{Original Terra panel} & \multicolumn{3}{c}{Independent Terra panel}\\
\cmidrule(lr){2-4}\cmidrule(lr){5-7}
$\epsilon$ & Agreement & Strict & Public & Agreement & Strict & Public\\
\midrule
0.000 & 0.167 & 0.710 & 0.668 & 0.083 & 0.762 & 0.695 \\
0.020 & 0.167 & 0.707 & 0.669 & 0.208 & 0.748 & 0.689 \\
0.050 & 0.125 & 0.714 & 0.670 & 0.250 & 0.748 & 0.689 \\
0.100 & 0.125 & 0.714 & 0.670 & 0.250 & 0.748 & 0.689 \\
\bottomrule
\end{tabular*}
\end{table}

\subsection{Record Assignments Change Comparisons}\label{sec:pairing-results}

We test the prediction from Section~\ref{sec:paired} that changing checkpoint assignments, while preserving each action's outcomes, can change conclusions.
On 48 local archive cells (Qwen3-8B, 12 prefixes $\times$ 4 faults), we fix all outputs and permute only their checkpoint assignments.

\begin{table}[ht]
\centering
\caption{Permutation controls on 48 archive cells. ``Changed'' counts cells where the permuted relation differs from the exact keyed reference. Columns report median cell proportions.}
\label{tab04_archive}
\begingroup
\small
\setlength{\tabcolsep}{4.5pt}
\renewcommand{\arraystretch}{1.0}
\begin{tabular}{@{}lrrrrr@{}}
\toprule
 & \multicolumn{4}{c}{Cell relation} &  \\
\cmidrule(lr){2-5}
Association & {Direct} & {Overlap} & {Discordant} & {Changed} & {Mean set size} \\
\midrule
Independent permutation & 0.042 & 0.417 & 0.542 & 0.125 & 1.813 \\
Shared permutation & 0.042 & 0.438 & 0.521 & 0.083 & 1.875 \\
Exact keyed & 0.042 & 0.458 & 0.500 & 0.000 & 1.875 \\
\bottomrule
\end{tabular}\par
\endgroup
\end{table}

The shared and independent permutations change 4 and 6 of 48 cell conclusions at the median (Table~\ref{tab04_archive}), despite preserving every action's outcome multiset.
The result confirms the prediction from Section~\ref{sec:paired}, namely that the bindings between checkpoints and actions carry information beyond what the marginal outcome distributions contain.
Only 4.2\% of cells retain a singleton joint target under permutation (Table~\ref{tab04_archive}), meaning that for most cells, multiple checkpoint--action assignments are consistent with the same outcomes, and different assignments lead to different conclusions.
If an evaluation pipeline fails to log which checkpoint produced which execution (e.g., due to batched processing or asynchronous dispatch), the resulting permuted bindings silently corrupt 8 to 13\% of cell-level conclusions, exceeding typical measurement noise in agent evaluations and remaining undetectable from the outcome data alone.

In a separate retrieval audit, EC$^2$~\citep{golovin2010noisy} requires 26 lookups to identify all action--eligibility conclusions, while eligibility-first settling completes in 19 (Appendix~\ref{sec:archive-joint-target}).
A restricted candidate class covers only 14 of 192 canonical tuples, causing all three selectors to stop incorrectly in the same three cells (Appendix~\ref{sec:archive-target-refinement}).

\subsection{Certification Requires Large Margins}\label{sec:budget-results}

\begin{figure}[t]
\centering
\includegraphics[width=\linewidth]{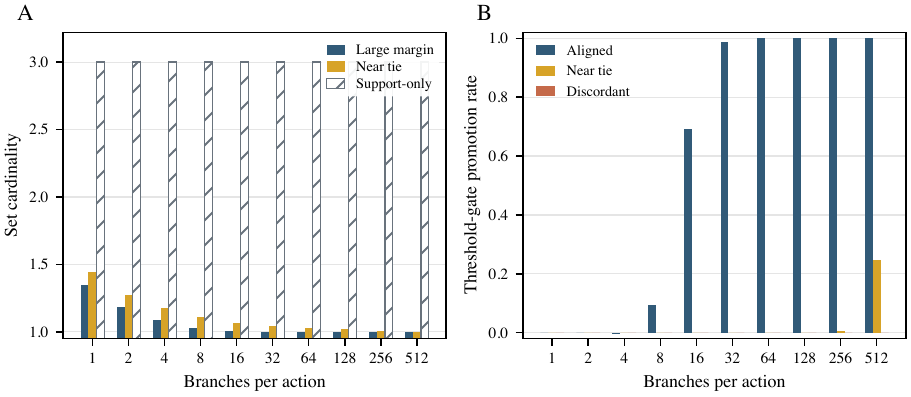}
\caption{\textbf{(A)}~Mean certified-set cardinality as the sample budget grows.  With a large margin (0.40), the set shrinks to a singleton by ${\sim}32$ branches; a near tie (0.06) remains unresolved at 512.  The support-only envelope (hatched) always retains all three actions.  \textbf{(B)}~Threshold-gate promotion rate.  Large margins reach near-complete certification by 128 branches; near ties stay below 0.1\% even at 512.  Zero false certificates were observed across all conditions.}
\label{fig:stability}
\end{figure}

We validate the Hoeffding certificate (Eq.~\eqref{eq:finite-certificate}) on controlled distributions with known margins (Figure~\ref{fig:stability}).
When the margin is large (0.40), valid certification rises from 40.5\% at 64 branches to 97.9\% at 128. In contrast, when the margin is small (0.06), certification stays below 0.1\% even at 512 branches per action. No false certificates were observed across all 36 nested cells and 10 budgets (Appendix Table~\ref{tab05_budget}).
At the margins typically observed in agent recovery experiments (success differences of 5 to 10 percentage points), certifying a unique best action would require hundreds or thousands of executions per action, far beyond current evaluation budgets.
In a 12-checkpoint development pilot with 32 draws per action, only one pointwise certificate was positive, and simultaneous correction left all unresolved (Appendix~\ref{sec:confidence-pilot}).
Most current evaluations therefore operate in a regime where statistical certification is infeasible, reinforcing the importance of reporting outcome levels alongside agreement.
The gap between what certification requires and what current budgets provide also explains why practitioners default to set agreement in the first place. It is the only metric that returns a definitive answer at small sample sizes, even though that answer is uninformative about quality.

\noindent\textbf{Changing the release target.}
On 10{,}000 controlled releases (Appendix Table~\ref{tab:release-dual-target}), requiring a replay action to be among the expected maximizers (the overlap target) admits population ties, while requiring it to be the unique winner excludes them.
Status-aware set membership admits more positives than the full gate with no observed errors under the overlap target, while the full gate makes fewer errors under the unique-winner target.
Record checks eliminate constructed provenance-gap errors in both rules (Appendix~\ref{sec:release-target-correspondence}).
The choice of release target thus has substantial practical consequences, as it determines which conclusions an evaluation can support and what evidence is needed.

\section{Discussion}\label{sec:conclusion}

A consistent pattern emerges across benchmarks, models, and sample budgets. Set agreement can be high even when recovery performance is near zero, because shared failure produces agreement just as effectively as shared success.
The set-path symmetry (Section~\ref{sec:set-path-method}) makes this precise, showing that Bernoulli parameters $(p_1,p_2)$ and $(1{-}p_2,1{-}p_1)$ yield identical best-action-set distributions. The RecoveryBench experiments confirm it empirically, with 85.4\% agreement coexisting alongside 1.4\% success.
The planning experiments extend this finding to a regime where recovery does work. Agreement drops from 40\% to 8\% as sample budgets grow, yet held-out success remains stable.

These results motivate several concrete recommendations. Every set-agreement report should be accompanied by the fraction of all-zero comparisons, held-out success, and pooled success probability, all of which can be computed from data already collected. Agreement should be decomposed by outcome quality to reveal whether it is driven by shared failure, shared success, or informative differentiation. Declines in agreement across sample budgets should not be interpreted as quality drops, since they may reflect tie resolution rather than performance degradation. Finally, evaluation pipelines should log checkpoint-to-action bindings, as our permutation experiments show that losing them silently corrupts 8 to 13\% of cell-level conclusions.

\noindent\textbf{Limitations.}
Our experiments use only binary outcomes. For continuous-valued metrics, the symmetry argument requires adaptation, though the qualitative insight that ordinal rankings do not determine absolute levels still holds.
The RecoveryBench experiments cover 12 checkpoints with three recovery actions, and richer action spaces would strengthen the empirical conclusions.
We do not propose or evaluate more sophisticated action-selection algorithms (e.g., Thompson sampling or UCB variants), which remain a natural direction for future work.
Finally, our analysis assumes independent Bernoulli outcomes across rounds. In practice, LLM outputs may exhibit autocorrelation due to temperature or seed reuse, which could affect both agreement and certification rates.

\label{main:end}

\bibliographystyle{iclr2027_conference}
\bibliography{references}

\appendix

\noindent\textbf{Guide to the supplement.} Definitions and proofs are collected in Appendix~\ref{sec:formal-method-details}; replay experiments and statistical units are specified in Appendix~\ref{sec:experimental-details}. Appendix~\ref{sec:positioning-appendix} develops identification boundaries and comparisons with prior work. Controlled evidence studies appear in Appendix~\ref{sec:observable-appendix}, with additional relation audits in Appendix~\ref{sec:supplementary-decision-audits}. Execution archives and binding retrieval are documented in Appendix~\ref{sec:live-appendix}; the fresh repeated-branch study is reported in Appendix~\ref{sec:prospective-protocol}; the frozen RecoveryBench panel and status-stratified analysis are presented in Appendix~\ref{sec:recoverybench-protocol}.

\FloatBarrier
\section{Definitions and proofs}\label{sec:formal-method-details}

This appendix formalizes the compatible law--record relation used by the checkpoint audit and proves the certification claims.  We first define the contract and compatible pairs, then derive the target images used in the main text.

\FloatBarrier
\subsection{Joint identification of preference and release}\label{sec:joint-identification}

\noindent\textbf{Audit contract.}
An audit contract $\kappa$ specifies a checkpoint state space $\mathcal S$, a recovery-action set $A_0$ with costs $C:A_0\to\mathbb R_{\geq0}$ (weighted by $\lambda\geq0$), a class of outcome laws $\mathfrak P_s$, and admissible law--record associations $\mathcal K_{\kappa,s}^{\mathrm{joint}}$.  Each state $s$ has a selector-visible signature $o=\phi(s)$.  The contract fixes the observable inputs; evaluator-side fault labels, answer keys, and unexecuted counterfactual outcomes are excluded.

\noindent\textbf{Compatible pairs.}
Given observed outcomes $y$ and record fields $x$, the set of outcome laws consistent with $y$ is $\mathcal P_s(y):=\{P\in\mathfrak P_s:y\in\operatorname{supp}(P)\}$, and the records consistent with $x$ are $\mathcal Z_{\kappa,s}(x):=\{z\in\mathcal Z_{\kappa,s}^{\mathrm{all}}:\psi_\kappa(z)=x\}$.  The \emph{compatible relation} intersects their product with the admissible associations:
\begin{equation}
\label{eq:joint-fiber-law}
\mathcal H_{\kappa,s}(y,x):=\bigl(\mathcal P_s(y)\times\mathcal Z_{\kappa,s}(x)\bigr)\cap\mathcal K_{\kappa,s}^{\mathrm{joint}}.
\end{equation}
When this equals the full product $\mathcal P_s(y)\times\mathcal Z_{\kappa,s}(x)$ (i.e., the association imposes no additional constraint), we say the relation is \emph{raw-rectangular}.
\label{eq:rectangular-bridge}

\noindent\textbf{Source keys and candidate bridges.}
A source key $\ell^\star$ (e.g., a model identifier) can restrict the relation before any target is computed: $\mathcal B_{\kappa,s}^{[\ell^\star]}$ retains only pairs whose record maps to key $\ell^\star$.
\label{eq:key-restricted-bridge}
A \emph{candidate bridge} $\mathcal B$ is a declared subset of law--record pairs representing a hypothesized data-generating process.  The set of checkpoint states consistent with observed data under bridge $\mathcal B$ is
\begin{equation}
\label{eq:candidate-supported-observable-fiber}
\mathcal S_{\kappa\mid o}^{\mathcal B}(\ell^\star;y,x):=\{s:\phi(s)=o,\ \mathcal B_{\kappa,s}^{[\ell^\star]}\cap\bigl(\mathcal P_s(y)\times\mathcal Z_{\kappa,s}(x)\bigr)\ne\varnothing\}.
\end{equation}
Two bridge systems are \emph{observationally equivalent} when they yield the same set of compatible state--law--record tuples.
\label{eq:bridge-equivalence-observable}
The collection of all candidate bridges consistent with the data is $\mathfrak B_{\kappa\mid o}(y,x,\ell^\star):=\{\mathcal B\in\mathcal B_\kappa^{\mathrm{cand}}:\mathcal S_{\kappa\mid o}^{\mathcal B}(\ell^\star;y,x)\ne\varnothing\}$.
\label{eq:candidate-supported-bridge-class}

\subsection{Population records and fixed-state targets}

\noindent\textbf{From single checkpoints to populations.}
When the evaluation covers a population of checkpoints $\mathcal D\subseteq\mathcal S$, the record and law spaces are products over individual checkpoints: $\mathfrak Z_{\kappa,\mathcal D}^{\mathrm{all}}:=\prod_{s\in\mathcal D}\mathcal Z_{\kappa,s}^{\mathrm{all}}$ and $\mathfrak P_{\kappa,\mathcal D}(\mathbf y):=\prod_{s\in\mathcal D}\mathcal P_s(y_s)$.
The population-level compatible relation requires each per-checkpoint pair to be admissible, and its target image collects the per-checkpoint overlap sets alongside a population-level eligibility bit.
For a single checkpoint ($|\mathcal D|=1$) this reduces to the fixed-state case; for larger populations it is a vector-valued target that cannot be replaced by a scalar summary.

\noindent\textbf{Record eligibility.}
Three binary checks determine whether a record supports reliable conclusions: \emph{completeness} $P_{\mathrm{rec}}$ (all required fields present), \emph{linkage} $P_{\mathrm{link}}$ (traceable to the generating model), and \emph{observability} $O_\kappa$ (scoring contract can be recomputed).  We write $C_\kappa:=P_{\mathrm{rec}}P_{\mathrm{link}}O_\kappa$ for the combined check.  The per-checkpoint overlap status $v_s(z)\in\{A,B,D,\textsc{NA}\}$ classifies whether the observed and reference best-action sets are equal ($A$), overlapping ($B$), disjoint ($D$), or missing ($\textsc{NA}$).  Direct-transfer eligibility requires all checks to pass and no disjoint or ambiguous cells:
\begin{equation}
\label{eq:truth-separation-bit}
D_\kappa^{\mathcal D}(z):=\mathbf1\{P_{\mathrm{rec}}(z)=P_{\mathrm{link}}(z)=O_\kappa(z)=1,\ d_\mathcal D(z)=b_\mathcal D(z)=0\},
\end{equation}
where $d_\mathcal D(z)=\mathbf1\{\exists s:v_s(z)=D\}$ flags any disjoint cell and $b_\mathcal D(z)$ flags ambiguity in the absence of disjointness.  An evidence update is \emph{truth-separating}\label{eq:truth-separation-map} when records with different eligibility can always be distinguished by some observable evidence.

\noindent\textbf{Target images.}
The joint target of the compatible relation maps each surviving pair $(P,z)$ to its overlap set $I_s(P)=R_s\cap A^\star(P)$ and eligibility $D_\kappa(z)$:
\begin{equation}\label{eq:joint-pair}
\Gamma_{\kappa,s}^{\mathcal I}(y,x):=\{(I_s(P),D_\kappa(z)):(P,z)\in\mathcal H_{\kappa,s}(y,x)\}.
\end{equation}
The binary version $\Gamma_{\kappa,s}(y,x)$ replaces $I_s(P)$ with $T_s(P)=\mathbf1\{I_s(P)\ne\varnothing\}$ (whether the overlap is nonempty).  We also write the coordinate projections as $\Gamma_s^{\mathrm{pref}}$ (preference images) and $\Gamma_{\kappa,s}^{\mathrm{rel}}$ (eligibility images).

\subsection{Pairing, singleton certification, and candidate coverage}

\noindent\textbf{Exact image vs.\ marginal envelope.}
For a relation $\mathcal B$ between laws and records, the \emph{exact} joint image $(I_s,D_\kappa)(\mathcal B)$ collects only actually paired overlap-set/eligibility values, while the \emph{marginal envelope} $I_s(\operatorname{pr}_P\mathcal B)\times D_\kappa(\operatorname{pr}_Z\mathcal B)$ takes the Cartesian product of coordinate projections and can therefore include unsupported combinations.
The exact image is always contained in the envelope; when equality holds the relation is \emph{target-rectangular}.  However, singleton certification is immune to this distinction.
\begin{equation}
\label{eq:pairing-singleton-invariance}
|\Gamma^{\mathcal I}(\mathcal B)|=1\ \Longleftrightarrow\ |I_s(\operatorname{pr}_P\mathcal B)|=|D_\kappa(\operatorname{pr}_Z\mathcal B)|=1.
\end{equation}
\begin{proof} A singleton joint image forces both coordinates to singletons; conversely, singleton coordinates produce exactly one Cartesian pair, which the nonempty relation attains.
\end{proof}

\noindent\textbf{Why the pairing matters.}
The distinction is sharp beyond the singleton case.  Two relations with identical coordinate projections but different law--record pairings (diagonal vs.\ crossed) produce different joint images:
\begin{equation}
\label{eq:pairing-sharpness-witness}
\Gamma^{\mathcal I}(\mathcal B_{\mathrm{diag}})=\{(I^+,1),(I^-,0)\},\qquad \Gamma^{\mathcal I}(\mathcal B_{\mathrm{cross}})=\{(I^+,0),(I^-,1)\}.
\end{equation}
Any readout that sees only coordinate projections returns the same answer for both relations and cannot distinguish these cases (Table~\ref{tab03_exact}).

\begin{table}[ht]
\centering
\caption{Exact joint-image queries on finite constructions. $N$ is the number of queries; Correct, FP, and Miss are counts for each readout.}
\label{tab03_exact}
\begingroup\small\setlength{\tabcolsep}{3pt}\renewcommand{\arraystretch}{1.0}
\begin{tabular}{@{}lrrrrrrr@{}}
\toprule
 &  & \multicolumn{3}{c}{Joint readout} & \multicolumn{3}{c}{Marginal readout} \\
\cmidrule(lr){3-5}\cmidrule(lr){6-8}
Query & $N$ & Correct & FP & Miss & Correct & FP & Miss \\
\midrule
Possible image & 15 & 15 & 0 & 0 & 13 & 2 & 0 \\
Singleton & 15 & 15 & 0 & 0 & 15 & 0 & 0 \\
Hidden pairing & 2 & 2 & 0 & 0 & 1 & 1 & 0 \\
\bottomrule
\end{tabular}
\endgroup
\end{table}

\noindent\textbf{Full-set certification.}
A gate that returns $\textsc{Continue}(I^\dagger)$ is \emph{sound} if and only if the robust candidate image $\Gamma_{\kappa\mid o}^{\mathcal I,\mathrm{rob}}$ (the joint image over all surviving candidates, possibly restricted by a source key $\ell^\star$) equals the singleton $\{(I^\dagger,1)\}$.
The canonical full-set query combines this singleton check with the record eligibility check:
\begin{equation}
\label{eq:gate-target-separation}
\mathsf{JointImage}_{\kappa}^{\mathrm{rob}}(I^\dagger\mid y,x)\Longleftrightarrow\Gamma_{\kappa\mid o}^{\mathcal I,\mathrm{rob}}(y,x;\ell^\star)=\{(I^\dagger,1)\}\ \land\ C_\kappa=1.
\end{equation}
If the image equals the displayed singleton, every surviving candidate shares the same nonempty overlap set and has an eligible record, so promotion is sound.  Any additional value in the image (an ineligible record, empty overlap, or different overlap set) makes promotion unsound.

\noindent\textbf{Active-bridge coverage.}
The certification above is relative to the declared candidate class.  To extend it to the \emph{active} relation (the true generating process), we require that every admissible active pair is covered by some candidate bridge:
Coverage ensures that a positive candidate singleton is also sound for the active relation.  A source key $\ell^\star$ that separates observable bridge classes ($\operatorname{Sep}_{\kappa}^{\mathrm{obs}}=1$) further refines the candidate fiber, so that a singleton image under the key implies the canonical full-set gate returns $\textsc{Continue}(I^\dagger)$.

\FloatBarrier
\subsection{Action-set bridge witness}\label{sec:action-set-bridge-witness}

This minimal example shows why the bridge (law--record pairing) must be observable.  Consider two actions $\{r,v\}$ with equal costs, both yielding success with probability 0.90, but differing in which action wins.  Under law $P_r$ action $r$ is the unique expected winner; under $P_v$ it is $v$:
\begin{equation}
P_r=0.90\,\delta_{(1,1)}+0.10\,\delta_{(1,0)},\qquad P_v=0.90\,\delta_{(1,1)}+0.10\,\delta_{(0,1)}.
\label{eq:action-set-bridge-witness}
\end{equation}
Each law is paired with its own eligible record ($z_r$ or $z_v$), producing two candidate bridges $\mathcal B_r=\{(P_r,z_r)\}$ and $\mathcal B_v=\{(P_v,z_v)\}$.
\label{eq:action-set-bridge-finite-domain}
Their joint images are singletons pointing to opposite winners: $\Gamma^{\mathcal I}(\mathcal B_r)=\{(\{r\},1)\}$ and $\Gamma^{\mathcal I}(\mathcal B_v)=\{(\{v\},1)\}$.
\label{eq:action-set-bridge-image}
Without a key that identifies which record generated the observation, the robust image contains both positive action sets, so the gate cannot authorize Continue.  A source key resolves the ambiguity only if the candidate contract associates it injectively with the corresponding law.

\subsection{Safety boundary and evidence requests}\label{sec:joint-decision-boundary}

The dispatcher chooses among $\textsc{Promote}$, $\textsc{Hold}$, and $\textsc{Reject}$ by minimizing worst-case loss over the surviving target image $\Gamma$.  With $\theta=(T_s(P),D_\kappa(z))$ encoding whether the overlap is nonempty and the record is eligible, the loss penalizes false promotion more heavily than missed promotion ($c_{\mathrm{FP}}>c_H>0$):
\begin{equation}
\label{eq:safety-dominant-loss}
\ell(a,\theta)=c_{\mathrm{FP}}\mathbf1\{a=\textsc{Promote},\theta\ne(1,1)\}+c_H\mathbf1\{a\ne\textsc{Promote},\theta=(1,1)\}.
\end{equation}
The minimax decision $\mathcal G^\star(\Gamma):=\arg\min_{d}\sup_{\theta\in\Gamma}\ell(d,\theta)$ contains only Promote when $\Gamma=\{(1,1)\}$ and excludes Promote whenever any nonpositive state survives.
\label{eq:safety-dominant-gate}

When the image is not yet a singleton, the dispatcher requests additional evidence along each unresolved coordinate:
\begin{equation}
\label{eq:next-evidence}
\mathsf U_{\kappa,s}(\mathcal H)=\{\mathsf{Repeat}:\Gamma_s^{\mathrm{pref}}=\{0,1\}\}\cup\{\mathsf{Collect}:\Gamma_{\kappa,s}^{\mathrm{rel}}=\{0,1\}\}.
\end{equation}
$\mathsf{Repeat}$ requests another task attempt to resolve the preference coordinate; $\mathsf{Collect}$ requests record evidence to resolve the eligibility coordinate.  Truth-separating evidence (Eq.~\eqref{eq:truth-separation-map}) collapses the corresponding coordinate to a singleton on the updated fiber.

\FloatBarrier
\section{Experimental protocols and replay controls}\label{sec:experimental-details}

\FloatBarrier
\subsection{Evaluation populations and statistical units}\label{sec:eval-populations}

The coordinate-stress audit draws 2,000 independent outcome vectors in each of four regimes (eligible, near-tie, missing-linkage, and insufficient-observability).  The separate release comparison has a fifth, discordant regime.  The acquisition study uses 4,000 controlled states with truthful informative responses and 64 paired seeds; calibrated selection uses 32 calibration and 32 disjoint evaluation seeds.  The preference study uses 10,000 independent vectors per operating point over 36 nested cells.  Finite relation queries use enumerated truth, and the dependence study reuses 4,000 latent specifications under product and coupled laws.

The replay population has development, in-distribution (ID), and structural out-of-distribution (OOD) splits.  The Qwen3-8B archive~\citep{yang2025qwen3} contains 12 prefixes, four faults, three actions, and two repetitions (48 cells and 288 branches); assignment controls and uncertainty cluster by prefix.  The strict score-aware filter retains 13 of 24 development prefixes, 28 of 48 ID prefixes, and 27 of 48 OOD prefixes (68 prefixes and 220 held-out prefix--fault cells).

\subsection{Workflow actions and execution records}\label{sec:replay-archive-design}

Replay compares fixed, status-only, and learned trace-only policies under a shared objective; transfer freezes the selector while the replay kernel is shifted, and completion, replay utility, and intervention cost are reported separately.  Luna, Terra, and Sol each provide two executions per action on the same 48-cell surface (288 branches per model); Qwen3-8B provides a separate local archive with the same two-repetition budget.  All 144 branches in each GPT-5.6 second execution satisfy the strict predicate.  The pooled score averages the two strict outcomes per action before subtracting cost.  Strict validity requires semantic correctness and public-score agreement; the local key binds outputs to checkpoint/action assignments, provider linkage is separate, and source groups remain distinct for artifact validity, record eligibility, and observed relations.  The provider archive has 12 held-out prefixes, nine satisfying the strict score-aware contract, and a 108-branch evaluation population; its records lack complete linkage.

\subsection{Comparison rules and outcome measures}\label{sec:design-controls}

Release comparisons fix branches and, for status-aware rules, supplied record information.  The full gate combines observed action relations with record checks; A-only omits record checks; set membership and serialized top-1 differ in their tie handling; Robust-95 requires statistical separation.  Acquisition fixes the budget cap and final eligibility criterion while comparing exact/calibrated selection with fixed local/joint request patterns; request cost is reported with coverage and excludes calibration acquisition from calibrated evaluation cost.

Three controls isolate associations: finite constructions vary law--record pairings while preserving coordinate projections; the dependence study varies joint outcome distributions while preserving marginal action probabilities; archive controls reassign retained outputs to fault cells while preserving each action-specific within-prefix multiset.  Promotion and acquisition coverage use eligible cases; errors and misses use each table's stated denominator.  Exact queries report correct, unsupported, and missed counts, archive comparisons report observed cell-relation changes, and statistical certification reports valid and false certificates at fixed budgets.

\subsection{Task construction and split composition}

The benchmark contains ranking, path-selection, and constrained-filtering tasks on typed directed acyclic graphs (DAGs), varying depth, breadth, information horizon, and parallelism.  Workers produce structured artifacts evaluated against deterministic answer keys.  For a task graph $D=(V,E,\tau,\delta)$, $\tau$ maps nodes to finite types and $\delta$ stores typed payloads and dependency annotations.  A faulted checkpoint is $s=\Psi_f(s_0)$, and for action $a$ and generator noise $u\sim\nu_{s,a}$,
\begin{equation}
\label{eq:typed-dag-contract}
\omega=(\mathsf B,g)=\mathsf{Branch}(s,a,u),\qquad
K_{\mathrm{rep}}^{Y,\kappa}(\cdot\mid s,a)=(Y_{s,a}^{\mathrm{rep},\kappa})_{\#}K_{\mathrm{rep}}^{\mathrm{raw}}(\cdot\mid s,a).
\end{equation}
Here $\mathcal F$ contains four injected fault surfaces; $\mathsf{Branch}$ covers typed transition, artifact materialization, and propagation draws, while $\mathsf{Clean}$ checks schema, evidence, and semantics.  The contract fixes the shared checkpoint/action interface while keeping replay law and endpoint record distinct.

\subsection{Trace representation and action contract}

The selector receives the task question, transcript prefix, intermediate artifact, dependency context, validator-visible checks, and public candidate table.  Artifact checks cover status/schema/version, role compatibility, candidate membership, evidence validity, score presence, and score recomputation; structural features record depth, breadth, information horizon, and parallelism.  Injected faults, answer keys, counterfactual branches, and future outcomes remain evaluator-side.

The action vocabulary is \texttt{continue}, \texttt{retry}, \texttt{verify}, and \texttt{replace\_agent}, with protocol costs $0$, $1$, $1$, and $2$; materialization uses the three recovery actions.  Replay utility subtracts the cascade penalty $0.01G$, while materialization stops at the recovered artifact and uses the shared-cost comparator.  Strict score-aware success requires semantic cleanliness and public-score agreement within the declared tolerances.  The frozen trace selector is a dependency-free CART~\citep{breiman1984cart} with depth 5, balanced class weights, and seed 2026.  It is fit on one row per development checkpoint--fault after collapsing action rows, then frozen before held-out replay and endpoint materialization; invalid or low-confidence traces (maximum class probability $<0.60$) route to verification.  The split manifest, counterfactual rows, objective-alignment record, and backend remain in the release bundle.  Semantic-clean sensitivity and endpoint family composition remain in the live archive.

\subsection{Primary replay uncertainty}\label{sec:replay-uncertainty}

Declared reference selects the action with highest expected utility under the original replay kernel, using a fixed order to resolve ties. In the shifted-kernel comparisons, Oracle selects the highest-utility action under the shifted kernel. Both use evaluator knowledge of their respective kernels; Trace-only chooses from the visible checkpoint features.

All replay intervals below are central 95\% prefix-cluster bootstrap intervals; prefixes, rather than fault cells, are the resampling unit~\citep{efron1979,cameron2008cluster}.  The same frozen transition kernel supplies objective-alignment, replay-headroom, and one-shot-reference rows.  Trace-only--Verify completion differences are $0.4196$ ($[0.3571,0.4732]$) on ID and $0.4537$ ($[0.3889,0.5185]$) on structural OOD; utility differences are $0.4620$ ($[0.4537,0.4704]$) and $0.5732$ ($[0.5562,0.5911]$), respectively (Table~\ref{tab:bc-replay-ci}).  The shifted-kernel audit changes only action-outcome probabilities; states, traces, actions, costs, fault surfaces, and score contract stay fixed.

\begin{table}[!htbp]
\centering
\small
\caption{Central 95\% prefix-cluster intervals for score-aware replay success rates. Prefix is the resampling unit; endpoints are rounded outward to three decimals.}
\label{tab:bc-replay-ci}
\begin{tabular}{@{}lrrr rrr@{}}
\toprule
Policy & \multicolumn{3}{c}{ID success} & \multicolumn{3}{c}{Structural OOD success}\\
\cmidrule(lr){2-4}\cmidrule(lr){5-7}
 & Estimate & Lower & Upper & Estimate & Lower & Upper\\
\midrule
Continue & 0.000 & 0.000 & 0.000 & 0.000 & 0.000 & 0.000\\
Blind retry & 0.268 & 0.232 & 0.304 & 0.250 & 0.213 & 0.287\\
Always verify & 0.438 & 0.366 & 0.500 & 0.407 & 0.343 & 0.472\\
Always replace & 0.893 & 0.839 & 0.938 & 0.889 & 0.824 & 0.944\\
Status-only & 0.670 & 0.589 & 0.741 & 0.611 & 0.546 & 0.676\\
Declared reference & 0.857 & 0.795 & 0.920 & 0.852 & 0.796 & 0.907\\
Trace-only & 0.857 & 0.795 & 0.920 & 0.861 & 0.806 & 0.917\\
\bottomrule
\end{tabular}
\end{table}

\noindent\textbf{Utility/regret intervals.} On the same prefix-cluster plane, each entry is Estimate $[{\rm Lower},{\rm Upper}]$: Original Reference has ID utility $0.805\,[0.803,0.808]$, OOD utility $0.779\,[0.774,0.783]$, and OOD regret $0.000\,[0.000,0.000]$; Original Trace-only has $0.805\,[0.803,0.808]$, $0.778\,[0.773,0.783]$, and $0.001\,[0.000,0.002]$; shifted Oracle has $0.753\,[0.750,0.756]$, $0.720\,[0.714,0.725]$, and $0.000\,[0.000,0.000]$; shifted Trace-only has $0.735\,[0.731,0.738]$, $0.691\,[0.684,0.698]$, and $0.029\,[0.027,0.031]$ for the same three metrics.

The original-kernel Trace-only structural-OOD regret interval $[0.000,0.002]$ crosses the predeclared $0.001$ boundary; the OOD label is therefore a point-estimate status and does not support transfer beyond this split.  The shifted Trace-only interval $[0.027,0.031]$ remains above that boundary.  On the original kernel, Trace-only minus Always replace utility is $+0.0250$ on both ID and OOD.  Completion contrasts are $-0.0357$ with $[-0.0893,0.0179]$ on ID and $-0.0278$ with $[-0.0926,0.0278]$ on OOD.  These are protocol-cost contrasts on the frozen replay surface and do not estimate endpoint quality.

\noindent\textbf{Unit-matched statistical audit.} Wilson intervals~\citep{wilson1927} apply to independent draws; paired-seed results retain cluster ranges, and release-vector results use a release-vector cluster bootstrap~\citep{efron1979,cameron2008cluster}.  The generating unit is retained in every row: seed for allocation, cell for joint-law controls, draw for coordinate stress, and vector for model-class certificates.  In the near-tie certificate row the estimate is $2\times10^{-5}$ before rounding, with original endpoints $1\times10^{-5}$ and $4\times10^{-5}$; displayed 0.000 is not necessarily an observed zero.

Key adverse rows are coordinate near-tie $0.049\,[0.040,0.060]$ at $N=2000$, joint discordant FP $0.000\,[0.000,0.001]$ at $N=4000$, and near-tie certification $0.000\,[0.000,0.001]$ at $n=256$, $N=10000$; the full row-level audit remains in the release record.

\subsection{Selector-fit stability}\label{sec:selector-stability}

Two hundred strict-development-prefix bootstrap refits leave action selection invariant on ID and structural OOD: held-out utility ranges are $[0.805,0.806]$ and $[0.778,0.778]$, regret is 0.000 on both splits, and action agreement with the frozen fit is 1.000.

\subsection{Fixed-kernel replay controls}\label{sec:fixed-kernel-controls}

\begin{table*}[ht]
\centering
\small
\caption{Frozen replay policies evaluated with $U_{\mathrm{pol}}$. $C_{\mathrm{emp}}$ is strict completion, $U_{\mathrm{pol}}$ expected replay utility, and $c$ mean intervention cost over 112 ID and 108 OOD cells.}
\label{tab:main-policy-surface}
\begin{tabular}{@{}lrrrrrr@{}}
\toprule
 & \multicolumn{3}{c}{ID} & \multicolumn{3}{c}{Structural OOD}\\
\cmidrule(lr){2-4}\cmidrule(lr){5-7}
Policy & $C_{\mathrm{emp}}$ & $U_{\mathrm{pol}}$ & $c$ & $C_{\mathrm{emp}}$ & $U_{\mathrm{pol}}$ & $c$\\
\midrule
Blind retry & 0.268 & 0.145 & 1.000 & 0.250 & -0.039 & 1.000\\
Always verify & 0.438 & 0.343 & 1.000 & 0.407 & 0.205 & 1.000\\
Always replace & \textbf{0.893} & 0.780 & 2.000 & \textbf{0.889} & 0.753 & 2.000\\
Status-only & 0.670 & 0.583 & 1.000 & 0.611 & 0.499 & 1.000\\
\textbf{Trace-only} & 0.857 & \textbf{0.805} & 1.500 & 0.861 & \textbf{0.778} & 1.500\\
\bottomrule
\end{tabular}
\end{table*}

Trace-only has the highest replay utility, exceeding Always replace by 0.025 on each split at mean cost 1.50 versus 2.00, while Always replace has the highest completion rate (Table~\ref{tab:main-policy-surface}).  The propagation-aware control is
\begin{equation}
\label{eq:main-policy-surface}
U_{\mathrm{pol}}(s,a)=\mathbb E_{\omega\sim K_{\mathrm{rep}}^{\mathrm{raw}}(\cdot\mid s,a)}
\left[Y^{\mathrm{rep},\kappa_{\mathrm{pol}}}(s,a;\omega)-\lambda C(a)-0.01G(s,a;\omega)\right].
\end{equation}
It is a workflow control and does not alter the shared-cost preference or release estimands.  Replay headroom has a minimum top-two margin of 0.010 on ID, structural OOD, and the combined strict test; replacement-ratio sensitivity is retained in the release bundle.

\section{Related work and identification boundaries}\label{sec:positioning-appendix}

Best-arm identification supplies the repeated-arm reference~\citep{audibert2010best,garivier2016optimal,jamieson2014lilucb,kaufmann2016complexity,truong2025multipleoptima}; partial-identification work supplies compatibility-set language~\citep{manski2003partial,fan2017joint,yata2021partial,christensen2022partial}; selective and set-valued prediction supply abstention and alternative-action interfaces~\citep{fuentes2026setvalued,geifman2017selective,liu2026agentabstain}.  Replay and recovery studies measure scalar utility, repair, or trajectory validity~\citep{gonuguntla2026replaygap,shah2026causalreplay,qi2026r2act}; execution-edit safety work studies checkpoint, fork, restore, and merge semantics~\citep{zheng2026executionedits}; provenance and action-certificate studies verify source or execution objects~\citep{liao2026auditingprovenance,she2026provenanceguard,wang2026cava,wang2026pcaa,zhu2026claimreceipt}.  OrchLoop adds a contract-relative action-set relation paired with release status at one checkpoint.  

The live endpoint receives task and candidate documents, the faulted artifact, dependency context, and retained prior-artifact score; validator-visible checks are available to the selector but are not additionally serialized to the endpoint.  Fault labels, answer keys, future outcomes, and unexecuted counterfactual results are excluded from both inputs.  The reconstructed-input manifest hashes all 108 recovery inputs; missing provider request identifiers, complete prompts, and source-complete response records set provider linkage to zero.

\subsection{Serialization sensitivity of top-1 transfer}\label{sec:readout-appendix}

For a fixed action order $\sigma$, let $\operatorname{Pref}_{\sigma}(k)$ be its first $k$ actions, $H_{\sigma}(s;L)$ the earliest-action readout, and $M_s(L)=\mathbf 1\{\mathcal R_s^{\mathrm{sc}}\cap L\ne\varnothing\}$. Then
\begin{equation}
\label{eq:main-prefix-criterion}
\left[\forall\,\varnothing\ne L\subseteq\mathcal A_{\mathrm{rec}}:\ H_{\sigma}(s;L)=M_s(L)\right]\Longleftrightarrow
\mathcal R_s^{\mathrm{sc}}=\operatorname{Pref}_{\sigma}\!\left(\left|\mathcal R_s^{\mathrm{sc}}\right|\right).
\end{equation}
Thus materialization is set-valued and order-invariant, while top-1 can change under serialization.  The primary readout retains ties; unique-winner sensitivity is diagnostic.

For a nonempty live-best set $L$, the replay-relative status is
\begin{equation}
\label{eq:admissible-status-map}
\nu_{\mathcal R}(L)=
\begin{cases}
A,&L=\mathcal R,\\
B,&L\cap\mathcal R\ne\varnothing,\ L\ne\mathcal R,\\
D,&L\cap\mathcal R=\varnothing.
\end{cases}
\end{equation}
For a nonempty cell population $\mathcal D$, write $d_{\mathcal D}=\mathbf 1\{\exists s:v_s=D\}$, $b_{\mathcal D}=\mathbf 1\{d_{\mathcal D}=0,\exists s:v_s=B\}$, and let $P_{\mathrm{rec}},P_{\mathrm{link}},O_\kappa$ be record, linkage, and observability checks.  On the declared three-status domain,
\begin{equation}
\label{eq:population-status-bits}
M_{\mathcal D}=\mathbf 1\{\forall s:v_s\in\{A,B\}\}=1-d_{\mathcal D}.
\end{equation}
The admissible interface class and its sufficient quotient are
\begin{equation}
\label{eq:admissible-release-class}
\mathfrak A_{\kappa}(\mathcal R;\mathcal D)=\mathcal V_{\mathcal R}^{\mathcal D}\times\{0,1\}^{3}\times
\operatorname{Bij}(\mathcal A_{\mathrm{rec}},[m]).
\end{equation}
\begin{equation}
\label{eq:evidence-quotient}
Q_{\kappa}(\mathcal D)=
(M_{\mathcal D},d_{\mathcal D},b_{\mathcal D},P_{\mathrm{rec}},P_{\mathrm{link}},O_\kappa).
\end{equation}
The reduced interface drops the redundant $M_{\mathcal D}=1-d_{\mathcal D}$ coordinate.  A two-bit card summary is insufficient: Partial and Provider-gap cards can share $(\min M,\min H_\sigma)=(1,1)$ while the contract returns Hold-Ambiguity and Hold-Record-Completeness, respectively.  The proof is by toggling one status or record coordinate at a time; serialization affects $H_\sigma$ but none of the gate coordinates.  If missing-information status \textsc{NA} is admitted, $M_{\mathcal D}$ must explicitly exclude both $D$ and \textsc{NA}, and a required \textsc{NA} forces a completeness hold.

\subsection{Observational non-identifiability of a trace-only gate}

The masked probe replaces the five marker keys while recomputing public derived fields.  Let $\mathcal U_T$ be the transformed observable classes and $C_u$ the corresponding collision class.  If a class contains different recovery labels, no deterministic selector using only that observable input can match all labels.  With singleton replay labels,
\begin{equation}
\label{eq:collision-accuracy}
B_{\mathrm{acc}}=\frac{1}{N_{\mathrm{obs}}}\sum_{u\in\mathcal U_T}\max_{a\in\mathcal A_{\mathrm{rec}}}n_{u,a}.
\end{equation}
\begin{equation}
\label{eq:collision-utility}
B_U=\frac{1}{N_{\mathrm{obs}}}\sum_{u\in\mathcal U_T}\max_{a\in\mathcal A_{\mathrm{rec}}}
\sum_{z\in C_u}Q^{\mathrm{replay}}_{\kappa_{\mathrm{pol}}}(z,a).
\end{equation}
where $n_{u,a}$ counts labels in $C_u$ and
\begin{equation}
\label{eq:collision-replay-value}
Q^{\mathrm{replay}}_{\kappa_{\mathrm{pol}}}(z,a)
=\mathbb E_{\omega\sim K_{\mathrm{rep}}^{\mathrm{raw}}(\cdot\mid z,a)}
\left[\Pi_{\mathrm{pol}}(z,a;\omega)\right].
\end{equation}
The five-key mask yields same-prefix collisions with distinct recovery-optimal labels in a held-out ID witness.  The field vector and witness identifier remain in the release artifact; the construction establishes non-identifiability under the mask, while model-specific inference of the hidden mechanism remains unmeasured.  The unit is a single-worker checkpoint; worker-pool independence and a persistent process are outside the measured object.

\subsection{One-shot expected-ranking identification audit}\label{sec:one-shot-identification}

\noindent\textbf{Why one observation is never enough.}
Let the rich class $\mathfrak P_{\mathrm{rich}}$ contain all independent-Bernoulli product laws over recovery actions, with cost-adjusted utility $q_a(P)=p_a-\lambda C(a)$ and best-action set $A^\star(P)=\arg\max_a q_a(P)$.
\label{eq:rich-recovery-class}
Under this class, every outcome vector $y$ has positive probability under some law whose best actions exclude any given target set $\mathcal R_{\mathrm{sc}}$ (set the target actions' success rates near zero and competitors' near one).  Therefore any gate with zero false-promotion rate must abstain on every single observation:
\begin{equation}
\label{eq:zero-false-promotion}
\sup_{P\in\mathfrak P_{\mathrm{rich}}:A^\star(P)\cap\mathcal R_{\mathrm{sc}}=\varnothing}
P\{g(Y)=1\}=0\quad\Longrightarrow\quad g\equiv 0.
\end{equation}
The same argument extends to any fixed finite number of independent branches. Controlled-risk release requires either repeated evidence or structural restrictions on the law class.

\noindent\textbf{Observable hidden-state mixtures.}
When multiple hidden checkpoints share the same observable signature $o$, the observed outcome is a mixture over checkpoint-specific kernels.  The mixture utility set $\mathcal U_o^{\mathrm{obs}}(y)$ collects all possible expected-utility vectors consistent with $y$ and the declared mixing family, and the transfer bit $\mathcal T_R^{\mathrm{obs}}(y)$ records whether the target action set $R$ overlaps the best-action set for each such utility vector.
A full-support bad law in the fiber restores the one-shot abstention boundary.

\noindent\textbf{Repeated support and matched coupling.}
With repeated independent outcome vectors, a bad-law class whose support covers the entire sample space still leaves no safe promotion region.
However, the support geometry depends on the \emph{coupling} between actions within each vector.  Consider two laws with identical per-action marginals $\operatorname{Bernoulli}(\theta)$ and $\operatorname{Bernoulli}(1-\theta)$, a shared-shock law (actions succeed or fail together via a common uniform draw $U$) and a product law (actions are independent).
\begin{equation}
\label{eq:matched-marginal-laws}
P_{\theta}^{\mathrm{ss}}=\operatorname{Law}\bigl(\mathbf 1\{U\leq\theta\},\mathbf 1\{U\leq1-\theta\}\bigr),\quad
P_{\theta}^{\times}=\operatorname{Bernoulli}(\theta)\otimes\operatorname{Bernoulli}(1-\theta).
\end{equation}
The product law has full support on $\{0,1\}^2$, so its bad subfamily leaves no safe region.  The shared-shock law for $\theta<1/2$ cannot produce $(1,0)$, making that outcome a safe promotion witness for target $\{1\}$:
\begin{equation}
\label{eq:coupling-blind-witness}
\mathcal X_{\{1\}}^{\mathrm{safe}}(\{P_\theta^\times\})=\varnothing,\qquad
\mathcal X_{\{1\}}^{\mathrm{safe}}(\{P_\theta^{\mathrm{ss}}\})=\{10\}.
\end{equation}
If the coupling is undeclared, the safe region is the intersection across admissible classes, so a gate that cannot distinguish couplings must be conservative wherever any class has a bad law.

\subsection{Finite-branch certification and statistical release}

\noindent\textbf{Hoeffding-based certificates.}
Given $n$ branches per action, the empirical cost-adjusted utility $\widehat q_{a,n}=n^{-1}\sum_i Y_{a,i}-\lambda C(a)$ and Hoeffding radius $\varepsilon_n(\delta)=\sqrt{\log(2m/\delta)/(2n)}$ define a singleton-or-abstain certificate:
\begin{equation}
\label{eq:certificate-radius}
\widehat{\mathcal R}_{n,\delta}=
\left\{a:\widehat q_{a,n}-\max_{b\ne a}\widehat q_{b,n}>2\varepsilon_n(\delta)\right\},
\quad
\Pr\{\widehat{\mathcal R}_{n,\delta}\subseteq A^\star\}\ge1-\delta.
\end{equation}
The proof is a union bound; within-vector dependence is allowed.  With a unique target gap $\Delta>0$, certification succeeds with probability $\ge 1-\delta$ once $n>8\log(2m/\delta)/\Delta^2$, and the misidentification probability decays exponentially:
\begin{equation}
\label{eq:branch-budget-bound}
\Pr\{\widehat A_n\ne\{a^\star\}\}\le 2(m-1)\exp(-n\Delta^2/2).
\end{equation}
The near-tie gap 0.06 remains largely unresolved at $n=512$, while the large-margin certificate rises from 40.5\% at $n=64$ to 97.9\% at $n=128$ with no false certificates.

\noindent\textbf{Statistical release.}\label{sec:statistical-release}
If the confidence relation covers the true law--record pair (failure $\le\delta$) and the decoder is correct (failure $\le\eta$), then a singleton release must equal the true target:
\begin{equation}
\label{eq:statistical-release-risk}
\Pr\{\textsc{StatisticalContinue}\text{ outputs an incorrect target}\}\le\delta+\eta.
\end{equation}
An additional bridge-omission term $\alpha$ gives $\alpha+\delta+\eta$; a coverage bound already including bridge omission must not double-count.  The conservative maximizer image from simultaneous utility intervals $[l_a,u_a]$ is:
\begin{equation}
\label{eq:box-maximizer-sets}
\mathcal S(l,u)=\{\varnothing\ne S\subseteq A_0:\max_{a\in S}l_a\le\min_{a\in S}u_a,\
\max_{b\notin S}l_b<\min_{a\in S}u_a\}.
\end{equation}

\noindent\textbf{Calibration boundary.}
For a model class where the winner is unknown but the gap is at least $\Delta$, the Le Cam--Pinsker lower bound~\citep{lecam1986,pinsker1964} gives the minimax error for identifying the best action from $n$ branches with gap $2\zeta$:
\begin{equation}
\label{eq:two-action-lower-bound}
\inf_{\widehat a}\sup_{\theta\in\{+,-\}}\Pr_\theta\{\widehat a\ne a^\star_\theta\}
\ge\tfrac12\left(1-\sqrt{2n\zeta\log\tfrac{1+2\zeta}{1-2\zeta}}\right)_+.
\end{equation}
This also lower-bounds set certification, since a correct singleton induces a correct selector.
Fixed-budget resolution follows best-arm analyses~\citep{audibert2010best,kaufmann2016complexity,jamieson2014lilucb,lattimore2020bandit}; stopping-aware evidence requires time-uniform confidence sequences~\citep{howard2021timeuniform}.

\noindent\textbf{Constructing a relation from anonymous execution blocks.}
Suppose $m$ stored blocks of $n$ observations each are available, one block per action, but the action bindings are hidden.  Let $\mathcal M(x)$ enumerate all block-to-action bijections consistent with the visible metadata.  For each block $j$, construct a Clopper--Pearson interval $[L_j,U_j]$ at level $\delta/m$~\citep{clopper1934,bonferroni1936}.  The confidence relation is:
\begin{equation}
\widehat H_\delta=\bigcup_{\beta\in\mathcal M(x)}\{(P,z_\beta): L_j\leq\mathbb E_P[Y_{\beta(j)}]\leq U_j\ \forall\, j\}.
\label{eq:block-confidence-relation}
\end{equation}
Coverage $\ge 1-\delta$ holds because only the true binding's $m$ means need to be covered, and there is no multiplicity penalty for $|\mathcal M(x)|$.  If metadata or a source receipt can exclude the true binding with failure probabilities $\alpha_M$ and $\alpha_R$, coverage becomes $1-\delta-\alpha_M-\alpha_R$ by a union bound.  This construction separates the checkable source-assignment contract from the sampling assumptions needed for preference inference.

\noindent\textbf{Exact ties under controlled error.}\label{sec:exact-tie-boundary}
Full-set certification faces a fundamental limit at exact ties.  If $p_0$ is a parameter where two or more actions are exactly tied and nearby parameters $p_k\to p_0$ break the tie, then any uniformly valid certifier satisfies:
\begin{equation}
\sup_p\Pr_p\{\text{release an incorrect set}\}\leq\delta\quad\Longrightarrow\quad\Pr_{p_0}\{\text{release }I_0\text{ in finite time}\}\leq\delta.
\label{eq:exact-tie-release-bound}
\end{equation}
\begin{proof} Releasing $I_0$ is incorrect at every $p_k$, so $\Pr_{p_k}(E\cap\{T\le N\})\le\delta$.  Finite-horizon TV continuity gives the same bound at $p_0$; taking $N\to\infty$ proves the claim.
\end{proof}

Thus a controlled-error rule cannot reliably certify an exact multi-action tie from finite outcomes over the unrestricted class.  A declared gap or tolerance-based target changes this problem; standard unique-best-arm formulations impose uniqueness explicitly~\citep{garivier2016optimal}.

\noindent\textbf{Compatibility boundary.}
With one branch vector, the realized overlap is exact, but the rich class allows every action to be a possible unique expected winner. The expected-transfer bit therefore has support-compatible range $\{0,1\}$ in every cell.  Simultaneous 95\% Clopper--Pearson intervals~\citep{clopper1934,bonferroni1936} are reported as model-class diagnostics, separate from the realized readout.

\FloatBarrier

\section{Controlled evidence and decision studies}\label{sec:observable-appendix}\label{sec:decision-audits}

\subsection{Generators for the main release and request comparisons}\label{sec:main-simulation-protocols}

\noindent\textbf{Release population.}
Table~\ref{tab:release-dual-target} uses \texttt{independent\_gate\_decision\_comparison.py}: five equally weighted regimes, 2,000 independent releases each, and four cells per release. The action order is retry, verify, replacement, with normalized costs $(1,1,2)$, $\lambda=0.05$, and deductions $c_a=\lambda C(a)$. Each release chooses replay action $r$ uniformly; $b$ is the first other action in that order. In the order $(r,b,\text{remaining})$, the Bernoulli mean triples are $(0.76,0.44,0.44)$ for eligible and both record-gap regimes, $(0.62+c_r,0.62+c_b,0.44)$ for the tied regime, and $(0.46,0.76,0.44)$ for discordance. All four cells use that triple and 32 draws per action. Conditional on these choices, outcomes are independent across draws, actions, cells, and releases; every rule sees the same draws. The separate eight-draw output is not included in this table.

Completeness is zero only for provider gap, observability is zero only for observability gap, and linkage is fixed at one. Thus $h_i=P_{\mathrm{rec},i}O_{\kappa,i}$ incorporates all varying supplied checks. The eligible regime has $(u_i,o_i)=(1,1)$, the tied regime $(0,1)$, and the other three $(0,0)$, using the definitions in Section~\ref{sec:paired-release-contrasts}. These are design weights rather than estimated task frequencies. Each observed set maximizes the empirical mean minus cost, with numerical tie tolerance $10^{-12}$. Membership requires $r$ in every cell's observed set; Full gate requires each set to be exactly $\{r\}$ and supplied checks to pass. Top-1 uses the fixed action order. Robust-95 requires the replay lower bound to exceed every competitor upper bound with radius $\sqrt{\log(6/0.05)/(2\cdot32)}$. The generator seed is 20260825.

\noindent\textbf{Request population and responses.}
Table~\ref{tab02_acquisition} uses 4,000 states: 250 per pair of family (rank, path, filter, delegation) and unresolved-coordinate regime (preference only, release only, both, neither). Within each family--regime group, index states by $k=0,\ldots,249$. An unresolved coordinate starts at $\{0,1\}$ and a resolved coordinate at $\{1\}$. Truths $(t_P,t_R)$ are $(k\bmod2,1)$, $(1,k\bmod2)$, $(\lfloor k/2\rfloor\bmod2,k\bmod2)$, and $(1,1)$ in those four regimes. The population contains 2,248 eligible and 1,752 ineligible states, including 248 eligible states with both coordinates unresolved.

The five requests are Repeat, Collect, JointCertificate, RepeatReplica, and CollectReplica, with costs $(0.700,0.800,1.100,0.850,0.650)$ and coverage sets $\{P\},\{R\},\{P,R\},\{P\},\{R\}$ in that order. Budget is 1.50. Every one of the 32 subsets is considered, subject to its total cost and coverage. Family-specific $(q_R,q_C)$ are $(0.90,0.84)$, $(0.82,0.88)$, $(0.86,0.84)$, and $(0.78,0.92)$. The remaining informative probabilities are $q_J=q_Rq_C+0.04$, $q_{R'}=\max(0.05,q_R-0.06)$, and $q_{C'}=\max(0.05,q_C-0.08)$. The five response indicators are independent conditional on family/state; the joint request has its own indicator. An informative response reveals the true covered coordinates; otherwise the fibers remain unchanged. Exact selection maximizes complete informative coverage over feasible subsets.

Potential responses keyed by seed, state, and request are shared across policies. Exact and fixed rules use 64 seeds; calibrated selection uses 32 calibration seeds and 32 disjoint evaluation seeds, choosing from response coverage without truth labels. Evaluation cost in Table~\ref{tab02_acquisition} excludes calibration; its separate cost analysis is in Section~\ref{sec:evidence-coverage}. The source files and frozen response summaries are listed by table label in the supplementary code index.

\noindent\textbf{Observable contract and utility.} The observable perturbation fixes the strict score-aware replay states, action kernel, labels, prefix-cluster bootstrap, and eight nested deterministic noise draws per prefix. It changes only the trace. The five masked keys are \texttt{status\_timeout}, \texttt{version\_current}, \texttt{role\_extractor}, \texttt{schema\_v1}, and \texttt{score\_matches\_answer}, with neutral values $(0,1,1,1,1)$; derived status and score fields are recomputed. The unmasked view has no cross-fault collision, whereas the five-key view creates the declared conflicting classes and supplies the $O=1$ versus $O=0$ audit. The public scheduler bundle contains 90 task contexts and 392 opaque decision states. Codebook controls remap only the four fault-signature codes, evaluating all 24 within-prefix bijections with frozen and transformed-view refits (Table~\ref{tab:codebook-summary-appendix}).

\begin{table}[!htbp]
\centering\small
\caption{Observable-codebook shift on held-out replay. Agreement is with the held-out replay action; fixed and refit selectors receive the same transformed cells.}
\label{tab:codebook-summary-appendix}
\begin{tabular}{lrrrr}
\toprule
Split & Original & Fixed & Refit & Fixed regret \\
\midrule
ID & \textbf{1.000} & 0.375 & \textbf{1.000} & 0.293 \\
Structural OOD & 0.880 & 0.405 & 0.880 & 0.361 \\
\bottomrule
\end{tabular}
\end{table}

The collision audit uses the full transformed selector input and permits dependence across fault cells. Its best common-action accuracy and utility ceilings are reported in Table~\ref{tab:collision-ceilings}.
\begin{table}[!htbp]
\centering\small
\caption{Collision-class ceilings and observed agreement under the five-key view. Rates and counts use separate columns; $B_U$ is the best common-action replay-utility ceiling and Floor is its collision-regret floor.}
\label{tab:collision-ceilings}
\begin{tabular}{lrrrrrrr}
\toprule
Split & Obs. rate & Obs. $n$ & $B_{\mathrm{acc}}$ & $B_{\mathrm{acc}}$ $n$ & $B_U$ & Floor \\
\midrule
ID & 0.500 & 56 & 0.750 & 84 & 0.804 & 0.002 \\
Structural OOD & 0.380 & 41 & 0.870 & 94 & 0.779 & 0.000 \\
\bottomrule
\end{tabular}
\end{table}
The frozen selector has shifted-law regret $0.019$ on 112 ID cells and $0.029$ on 108 structural-OOD cells; its choices agree with the primary selector on all 220 cells. Interpolating the original and transfer kernels as $p_\alpha=(1-\alpha)p_{\mathrm{original}}+\alpha p_{\mathrm{transfer}}$ for $\alpha\in\{0,.25,.50,.75,1\}$ gives frozen regret $(0.000,0.001,0.004,0.008,0.019)$ on ID and $(0.001,0.004,0.008,0.015,0.029)$ on OOD. Condition-refit regret is zero at every displayed interpolation point except $0.001$ at $\alpha=0$ on OOD.

For checkpoint $s$, action $a$, and replay branch $\omega$, define the semantic outcome $Y^{\mathrm{sem}}(s,a;\omega)$ and
\begin{equation}
\label{eq:outcome-contract}
U^{\kappa_{\mathrm{pol}}}(s,a)=\mathbb E_{\omega\sim K_{\mathrm{rep}}^{\mathrm{raw}}(\cdot\mid s,a)}[\Pi_{\mathrm{pol}}(s,a;\omega)],\qquad
\Pi_{\mathrm{pol}}=Y^{\kappa_{\mathrm{pol}}}-0.05C(a)-0.01G(s,a;\omega).
\end{equation}
Here $Y^{\kappa_{\mathrm{pol}}}=Y^{\mathrm{sem}}S_{\mathrm{score}}$ requires semantic validity and score agreement; score comparison uses relative tolerance $10^{-4}$ and absolute tolerance 0.05. The replay readout evaluates this predicate on $K_{\mathrm{rep}}^{\mathrm{raw}}$, while the live readout evaluates retained endpoint artifacts without a propagation coordinate. The charge $G$ uses the frozen \texttt{cascade\_mass} draw. The policy reference is
\begin{equation}
\label{eq:policy-reference}
\mathcal R_{\mathrm{pol}}(s)=\operatorname{arg\,max}_{a\in\mathcal A_{\mathrm{full}}}U^{\kappa_{\mathrm{pol}}}(s,a).
\end{equation}
The one-shot score-aware relation removes $G$ and restricts to $\mathcal A_{\mathrm{rec}}$; the collision audit uses the recovery-only projection. All 462 semantic-clean counterfactual branches are score-consistent. The 90-prefix semantic-clean run is a separate sensitivity analysis, and its live utility omits cascade mass because execution stops after recovery.

\FloatBarrier
\subsection{Contract-level decision rule}\label{sec:status-contract}

The release gate maps a six-dimensional binary input (population disjointness $d_{\mathcal D}$, ambiguity $b_{\mathcal D}$, record completeness $P_{\mathrm{rec}}$, linkage $P_{\mathrm{link}}$, observability $O_\kappa$, and membership $M_{\mathcal D}$) to a disposition via the first applicable case.
\begin{equation}
\label{eq:population-gate}
\widetilde G_\kappa=\begin{cases}
\mathsf H_R,&P_{\mathrm{rec}}=0,\\
\mathsf H_L,&P_{\mathrm{rec}}=1,\ P_{\mathrm{link}}=0,\\
\mathsf H_O,&P_{\mathrm{rec}}=P_{\mathrm{link}}=1,\ O_\kappa=0,\\
\mathsf R_D,&P_{\mathrm{rec}}=P_{\mathrm{link}}=O_\kappa=1,\ d_{\mathcal D}=1,\\
\mathsf H_A,&C_\kappa=1,\ d_{\mathcal D}=0,\ (b_{\mathcal D}=1\lor M_{\mathcal D}=0),\\
\mathsf P_{\mathrm{obs}},&C_\kappa=M_{\mathcal D}=1,\ d_{\mathcal D}=b_{\mathcal D}=0.
\end{cases}
\end{equation}
A repeat request is issued only for target-separated ambiguity with positive margins and available budget; it cannot bypass record or observability holds.

\noindent\textbf{Coordinate necessity.}
Each coordinate is necessary. Starting from the all-pass state $(0,0,1,1,1)=\textsc{Promote}$, flipping any single coordinate changes the disposition (to rejection, ambiguity hold, record hold, linkage hold, or observability hold respectively).  No coordinate can be removed without collapsing distinct dispositions.

An endpoint scalar seeing only the replay-best set and observed best set cannot reproduce the gate, since two records with identical scalars but $P_{\mathrm{rec}}=0$ vs.\ $1$ receive \textsc{Hold} vs.\ \textsc{Promote}.

\FloatBarrier
\subsection{Model-class branch-budget audit}\label{sec:model-class-branch-budget}

The finite-branch audit varies repeated evidence while the evaluator knows the unique expected winner and cost-adjusted gap: 0.40 for the large-margin class, 0.06 for the near tie, and 0.36--0.40 for the heterogeneous class. Each of 36 nested cells uses budgets $n\in\{1,2,4,8,16,32,64,128,256,512\}$ and 10,000 independent releases (seed 20260822). The same Hoeffding rule in Eq.~\eqref{eq:finite-certificate} is used at every budget. The sufficient gap thresholds $4\rho_n(0.05)$ at $n=32,64,128,256,512$ are $1.094,0.774,0.547,0.387,0.274$.

\begin{table}[ht]
\centering\small
\caption{Finite-branch preference certification at $\delta=0.05$. Valid is the fraction of 360,000 cells with a correct unique-winner certificate; False counts incorrect certificates. Small nonzero rates may round to 0.0000; exact counts follow below.}
\label{tab05_budget}
\begin{tabular*}{\linewidth}{@{\extracolsep{\fill}}lrrrrrrrrrr@{}}
\toprule
 & \multicolumn{2}{c}{$n=32$} & \multicolumn{2}{c}{$n=64$} & \multicolumn{2}{c}{$n=128$} & \multicolumn{2}{c}{$n=256$} & \multicolumn{2}{c}{$n=512$} \\
\cmidrule(lr){2-3}\cmidrule(lr){4-5}\cmidrule(lr){6-7}\cmidrule(lr){8-9}\cmidrule(lr){10-11}
Class & Valid & False & Valid & False & Valid & False & Valid & False & Valid & False \\
\midrule
Near tie & 0.0000 & 0 & 0.0000 & 0 & 0.0000 & 0 & 0.0000 & 0 & 0.0004 & 0 \\
Heterogeneous & 0.0124 & 0 & 0.3084 & 0 & 0.9353 & 0 & 1.0000 & 0 & 1.0000 & 0 \\
Large margin & 0.0174 & 0 & 0.4054 & 0 & 0.9785 & 0 & 1.0000 & 0 & 1.0000 & 0 \\
\bottomrule
\end{tabular*}
\end{table}
Certification reaches one for the large-margin and heterogeneous classes by 256 branches, while the near tie remains at 0.0004 at 512. The exact near-tie valid counts at $n=32,64,128,256,512$ are $0,1,0,7,136$ out of 360,000, and every class--budget condition has zero false certificates. At 32 branches, set membership, serialized top-1, robust-95, and Contract-$A$-only have respectively $(\text{overlap},\mathrm{FP}_{\mathrm{all}},\mathrm{Miss}_{\mathrm{all}})$ equal to $(0.493,0.090,0.045)$, $(0.481,0.083,0.051)$, $(0.001,0.000,0.448)$, and $(0.469,0.077,0.057)$ on 12,000 draws.\label{sec:certificate-event-counts}

\FloatBarrier
\subsection{Deterministic gate control and joint-law checks}\label{sec:gate-calibration}\label{sec:joint-law-family}

The joint-law check compares observed compatibility across product and common-shock families with matched action marginals (Table~\ref{tab:main-joint-law}).
\begin{table*}[ht]
\vspace{-4mm}
\centering
\small
\caption{Observed compatibility under joint outcome laws with matched action marginals. Within each regime, both families use the same $N=4{,}000$ latent-law specifications. The two families share replay actions, costs, marginal success probabilities, and branch budgets per action. Branch outcomes are sampled separately for the two families. All rates are proportions of these specifications. OC denotes observed compatibility; FP and Miss denote false promotion and missed overlap, respectively.}
\label{tab:main-joint-law}
\begingroup
\small
\setlength{\tabcolsep}{2.2pt}
\begin{tabular*}{\linewidth}{@{\extracolsep{\fill}}lrrrrrrrrrrrr@{}}
\toprule
 & \multicolumn{6}{c}{Product} & \multicolumn{6}{c}{Coupled} \\
\cmidrule(lr){2-7}\cmidrule(lr){8-13}
 & \multicolumn{3}{c}{1 branch} & \multicolumn{3}{c}{32 branches} & \multicolumn{3}{c}{1 branch} & \multicolumn{3}{c}{32 branches} \\
\cmidrule(lr){2-4}\cmidrule(lr){5-7}\cmidrule(lr){8-10}\cmidrule(lr){11-13}
Regime & OC & FP & Miss & OC & FP & Miss & OC & FP & Miss & OC & FP & Miss \\
\midrule
Discordant & 0.372 & 0.372 & 0.000 & 0.002 & 0.002 & 0.000 & 0.427 & 0.427 & 0.000 & 0.000 & 0.000 & 0.000 \\
Near tie & 0.514 & 0.269 & 0.101 & 0.472 & 0.255 & 0.128 & 0.624 & 0.311 & 0.033 & 0.470 & 0.220 & 0.095 \\
Aligned & 0.646 & 0.000 & 0.354 & 0.992 & 0.000 & 0.008 & 0.748 & 0.000 & 0.252 & 0.999 & 0.000 & 0.001 \\
\bottomrule
\end{tabular*}
\endgroup
\vspace{-2mm}
\end{table*}

The known-by-construction control uses 36 cells, singleton replay labels cycling over three actions, five scenarios, and 20,000 Monte Carlo draws at one and eight branches per action (Table~\ref{tab:gate-calibration}). Strict requires alignment, complete provenance, and no collision; $\tau=0.95$, set intersection, and earliest serialization are baselines.
\begin{table}[!htbp]
\centering\small
\caption{Known-by-construction gate control. Acceptance rates use the predeclared truth; a displayed zero is $0/20{,}000$.}
\label{tab:gate-calibration}
\begin{tabular}{llrrrrrrrr}
\toprule
 &  & \multicolumn{4}{c}{1 branch} & \multicolumn{4}{c}{8 branches} \\
\cmidrule(lr){3-6}\cmidrule(lr){7-10}
Scenario & Truth & Strict & $\tau$ & Set & Top-1 & Strict & $\tau$ & Set & Top-1 \\
\midrule
Discordant & hold & 0.000 & 0.000 & 0.000 & 0.000 & 0.000 & 0.000 & 0.000 & 0.000 \\
Ambiguous & hold & 0.000 & 0.000 & 0.000 & 0.000 & 0.000 & 0.000 & 0.000 & 0.000 \\
Provider & hold & 0.000 & 0.000 & 0.000 & 0.000 & 0.000 & 0.000 & 0.656 & 0.553 \\
Codebook & hold & 0.000 & 0.000 & 0.000 & 0.000 & 0.000 & 0.000 & 0.654 & 0.551 \\
Aligned & promote & 0.000 & 0.000 & 0.000 & 0.000 & 0.466 & 0.622 & 0.651 & 0.549 \\
\bottomrule
\end{tabular}
\end{table}
In Table~\ref{tab:gate-calibration}, the strict implementation has $0/5$ held-card label disagreements and $1/1$ aligned promotion; replay-only promotion disagrees on $5/5$ held cards. Repeated projection can accept missing-provenance or collision cards, which is why the release gate retains both record and observability coordinates.

The joint-law check reuses 4,000 fixed-seed latent-law specifications under a product family and a common-shock family with probability 0.80 plus an independent component of probability 0.20. Per-action parameters lie in $(0,1)$, so both families have full support; repeated vectors are independent while within-vector actions may be dependent. Matched marginals, release signatures, budgets, and replay identities are held fixed. The mechanism-separation audit uses 8,000 paired cells per condition: $\gamma_i=(\text{family},\text{depth bin})$ is shared, the link condition uses $a^\star_{i,\mathrm{link}}=g(\gamma_i)$, and the permuted condition uses a balanced fold-wise action permutation. Cost-adjusted utilities use $\lambda=0.05$ and $C=(1,1,2)$, with margins $0.16$ or $0.03$ and budgets $1,8,32$. A leave-fold-out codebook predictor and set-valued, serialized, and robust-95 readouts use only the visible signature; their paired results are reported below.

\FloatBarrier
\subsection{Evidence allocation}\label{sec:evidence-allocator}\label{sec:shared-budget-allocator}

The coordinate-local audit crosses four families (rank, path, filter, delegation) with four requirement regimes (preference-only, release-only, joint, resolved), using 250 states per cell, 4,000 states, and 64 paired seeds. An informative response replaces its target fiber by truth; an uninformative response leaves it unchanged. The same potential response is paired across policies.
\begin{table}[ht]
\centering\small
\caption{Budget feasibility for coordinate-targeted requests. TP, FP, and Miss use eligible, ineligible, and eligible denominators; Joint TP is restricted to eligible states with both coordinates outstanding. Cost and Rounds are means over all 4,000 states.}
\label{tab:main-evidence-phase}
\begin{tabular}{@{}lrrrrrrr@{}}
\toprule
Rule & Budget & TP & FP & Miss & Joint TP & Cost & Rounds \\
\midrule
Scalar projection & 0 & 0.667 & 0.285 & 0.333 & 0.000 & 0.000 & 0.000 \\
\textbf{Typed-greedy} & 1 & 0.826 & \textbf{0.000} & 0.174 & 0.000 & 0.750 & 0.750 \\
\textbf{Typed-parallel} & 2 & 0.911 & \textbf{0.000} & 0.089 & 0.774 & 1.000 & 0.750 \\
\bottomrule
\end{tabular}
\end{table}

Two evidence channels are available, namely \emph{outcome evidence} (repeat the task to observe new outcomes) and \emph{record evidence} (collect provenance or observability information).  Each channel is modeled as a paired Bernoulli draw in which an informative response reveals the true coordinate value and an uninformative response leaves the fiber unchanged.
\noindent\textbf{Evidence update.}
New evidence restricts the compatible relation to pairs consistent with the observation; a \emph{separating} response collapses the corresponding coordinate image to a singleton.

\noindent\textbf{Unresolved coordinates and request vocabulary.}
Three coordinates may be unresolved: bridge identity, preference overlap, and record eligibility.  Each maps to a request type, respectively $\textsc{verify}$ (bridge), $\textsc{repeat}$ (preference), and $\textsc{collect}$ (eligibility).
Pending record checks (completeness $\sigma_R$, linkage $\sigma_L$, observability $\sigma_O$, each $\in\{\bot,0,1\}$) augment the request set: $\textsc{collect}$ is added for pending completeness or observability, and $\textsc{verify}$ for pending linkage.\label{sec:dispatcher-pending}\label{sec:requests}

\noindent\textbf{Dispatcher logic.}
The dispatcher follows a three-way rule. If the full-set certificate holds, it returns $\textsc{Continue}(I^\dagger)$; if no failure is confirmed and requests remain, it returns $\textsc{Request}(Q_{\mathrm{all}})$; otherwise it returns $\textsc{Hold}$.
\begin{equation}
\label{eq:action-set-dispatcher}
\pi_{\mathfrak C}^{\mathcal I}(\Gamma,\sigma,\mathsf U)=\begin{cases}
\textsc{Continue}(I^\dagger),&e=1,\\
\textsc{Request}(Q_{\mathrm{all}}),&e=0,\ F=0,\ Q_{\mathrm{all}}\ne\varnothing,\\
\textsc{Hold},&\text{otherwise}.
\end{cases}
\end{equation}
The budgeted version substitutes $\textsc{Hold}$ when the total request cost exceeds the remaining budget $B$.
A confirmed failure ($F=1$) is terminal; a missing check is pending, not terminal.  Truthful restriction can only shrink the target image, so Continue retains the original certificate, and a class with no positive target cannot acquire one.  With initial budget $B_0$ and minimum request cost $c_{\min}$, at most $\lfloor B_0/c_{\min}\rfloor$ rounds can occur.

\noindent\textbf{Coordinate-local guarantee.}
Under independent informative probabilities $q_{\mathrm{rep},c}$ and $q_{\mathrm{rec},c}$ for the two coordinates, the typed-parallel rule promotes an eligible case with probability equal to the product over its unresolved coordinates:
\begin{equation}
\label{eq:allocator-guarantee}
\Pr(\widehat E_c=1)=\mathbf1\{\theta_c=(1,1)\}\,q_{\mathrm{rep},c}^{u_{\mathrm{rep},c}}\,q_{\mathrm{rec},c}^{u_{\mathrm{rec},c}},
\end{equation}
where $u_{\mathrm{rep},c},u_{\mathrm{rec},c}\in\{0,1\}$ indicate unresolved coordinates.  False promotion is exactly zero because an ineligible coordinate cannot be resolved to $\{1\}$ by truthful evidence.

\noindent\textbf{Shared budget and coupled evidence.}
When the two evidence channels are positively correlated (joint informative probability $q^{\mathrm{rep}}q^{\mathrm{rec}}+\kappa$ with coupling $\kappa=0.04$), the typed-parallel rule still achieves zero false promotion:
\begin{equation}
\label{eq:shared-budget-guarantee}
\Pr(\widehat E_c=1\mid\theta_c=(1,1))=p_{U_c^{\mathrm{req}}},\qquad \Pr(\widehat E_c=1\mid\theta_c\ne(1,1))=0,
\end{equation}
where $p_{U_c^{\mathrm{req}}}$ is the joint informative probability over unresolved coordinates.  When the budget cannot cover both requests, a one-round rule must hold because one fiber remains unresolved.  A preference-only projection that omits the release coordinate has a positive false-promotion floor.

\FloatBarrier
\subsection{Evidence Acquisition under a Shared Budget}\label{sec:acquisition-results}

The acquisition comparison uses a catalog containing local requests for each coordinate, alternative local requests with different costs and reliabilities, and a request covering both coordinates.
Local-coordinate requests each outstanding coordinate separately; Joint-coordinate uses the joint request whenever any requirement remains.
Exact typed uses the declared request law to maximize feasible coverage; Calibrated typed estimates coverage from separate response observations. These four typed rules share the same final promotion condition.
Under the common budget cap $B=1.50$, Exact typed reaches 93.2\% eligible coverage, compared with 90.6\% for Local-coordinate and 87.2\% for Joint-coordinate (Table~\ref{tab02_acquisition}).
Exact typed uses more of the permitted budget than Local-coordinate; relative to Joint-coordinate, it improves coverage while reducing average expenditure.

\begin{table}[ht]
\centering
\vspace{-4mm}
\caption{Evidence acquisition under a shared request budget of $B=1.50$. Coverage and false promotion use eligible and ineligible states as denominators, respectively; mean cost is averaged over all states. The audit covers 4,000 states with 64 paired response vectors per state for exact and fixed rules; calibrated selection uses 32 calibration and 32 independent evaluation vectors per state. Informative updates are truthful, coverage and false promotion are proportions, and evaluation cost excludes calibration acquisition.}
\label{tab02_acquisition}
\begingroup
\small
\setlength{\tabcolsep}{4.5pt}
\renewcommand{\arraystretch}{1.0}
\begin{tabular}{@{}lrrr@{}}
\toprule
Rule & {Coverage} & {False promotion} & {Mean cost} \\
\midrule
Scalar projection & 0.667 & 0.285 & 0.000 \\
Joint-coordinate & 0.872 & 0.000 & 0.825 \\
Local-coordinate & 0.906 & 0.000 & 0.750 \\
\textbf{Calibrated typed} & 0.924 & 0.000 & 0.856 \\
\textbf{Exact typed} & 0.932 & 0.000 & 0.813 \\
\bottomrule
\end{tabular}\par
\endgroup
\vspace{-2mm}
\end{table}

Calibrated selection reaches 92.4\% coverage on independent evaluation draws, compared with 93.3\% for the exact selector on those same draws (Appendix Table~\ref{tab:evidence-calibration}).
In a separate simulation with independent responses and a request law shared within each case family, the cost including calibration is 9.034 per evaluation case, above the fixed policies' costs (Appendix Table~\ref{tab:calibration-transfer}).
The exact and calibrated selectors and the local and joint controls record zero false promotions, as expected from the truthful-update model and shared final gate.
Scalar projection, which omits the release requirement, promotes 28.5\% of ineligible states despite issuing no request.

Among the 248 eligible cases with both coordinates outstanding, Exact typed and Joint-coordinate each reach 77.0\% coverage, compared with 73.3\% for Local-coordinate. The aggregate comparison also includes cases with only one outstanding coordinate, where local requests can avoid the cost of a joint request. The separate budget-feasibility comparison is reported in Appendix Table~\ref{tab:main-evidence-phase}.

\subsection{Evidence coverage and calibration}\label{sec:evidence-coverage}

This audit uses a coordinate-local interface in which each binary coordinate fiber updates independently, with no cross-coordinate inference.  Each request $r$ in the catalog has cost $c_r>0$, a coverage set $S_r$ of coordinates it can resolve, and an informative indicator $Z_r$.  A coordinate is \emph{covered} when at least one informative request targeting it returns truth; uncovered fibers stay unchanged:
\begin{equation}
F_{c,u}^{A,Z}=\begin{cases}\{\theta_{c,u}\},&\exists r\in A:\ u\in S_r,\ Z_r=1,\\V_{c,u},&\text{otherwise}.\end{cases}
\label{eq:evidence-coverage-transition}
\end{equation}

\noindent\textbf{Theorem (budgeted evidence coverage).}
Under truthful updates and a one-round non-adaptive rule choosing requests $A$ within budget $B$, the optimal true-positive rate and false-positive rate satisfy:
\begin{equation}
\label{eq:evidence-coverage-optimum}
\operatorname{TP}^{\star}_c(B)=\max_{A\in\mathcal F_B(U_c)}p_c^{\mathrm{cov}}(A)\quad(c\text{ eligible}),\qquad \operatorname{FP}_c(B)=0\quad(c\text{ ineligible}),
\end{equation}
where $p_c^{\mathrm{cov}}(A)$ is the probability that all unresolved coordinates receive informative responses.  When the budget is below the minimum cover cost, $\operatorname{TP}=0$.

\begin{proof}  An eligible case promotes exactly when every unresolved coordinate is covered by an informative response.  Truthful updates cannot make an ineligible coordinate appear positive.  Maximizing over feasible covers gives the optimum; randomization over covers yields a convex combination and cannot improve it.  For two coordinates, the optimum is computed by exact enumeration.
\end{proof}

The coverage event is a conjunction over coordinates of a disjunction over requests:
With independent indicators, inclusion--exclusion gives the exact coverage probability.
For two outstanding coordinates, local requests have joint success $p_L=\Pr(Z_1=Z_2=1)$ and a coupled request has success $p_J=\Pr(Z_{12}=1)$. Compare only feasible covers: choose the local pair when $p_L>p_J$, the joint certificate when $p_J>p_L$, and hold when neither fits. If all three requests fit under independent indicators, coverage is
\begin{equation}
\label{eq:evidence-overlap-rescue}
p_c^{\mathrm{cov}}=1-(1-q_{12})(1-q_1q_2)=q_{12}+q_1q_2-q_1q_2q_{12}\geq q_1q_2q_{12}.
\end{equation}
The audit enumerates all 32 subsets of five declared request types; at $B=1.50$, expected and empirical coverage are 0.933 and 0.932.

\noindent\textbf{Exact and calibrated selection.}
Under the same coordinate-local transition, let $\Pi_B(c)$ index a finite family of feasible request sets $A_\pi\in\mathcal F_B(U_c)$ fixed before calibration, and write $p_c(\pi)=p_c^{\mathrm{cov}}(A_\pi)$.
If $\Pi_B(c)=\varnothing$, return Hold without issuing a request. Otherwise, exact selection uses the declared request law to choose $\pi_c^\star\in\operatorname{arg\,max}_{\pi\in\Pi_B(c)}p_c(\pi)$.
Enumerating all feasible sets recovers Eq.~\eqref{eq:evidence-coverage-optimum}.

For a nonempty family, calibrated selection uses $m\geq1$ independent response vectors $Z_{c,1},\ldots,Z_{c,m}$ from the conditional request law given $c$ and $\theta_c=(1,1)$. Each vector evaluates every candidate set and may contain dependent responses. Such data can come from an independently labelled eligible calibration population. Unlabelled calibration instead requires the informative-response law to be independent of eligibility conditional on the observed context; otherwise it estimates a mixture, and the bound below need not hold. Our finite simulation satisfies this independence by construction, with response probabilities determined only by the observed family.
Write $S_Z(A)=\bigcup_{r\in A:Z_r=1}S_r$ for the coordinates revealed by informative responses, so the coverage event is $\mathsf C_c(A,Z)=\mathbf1\{U_c\subseteq S_Z(A)\}$.
The empirical coverage and selected request index are
\begin{equation}
\widehat p_{m,c}(\pi)=\frac{1}{m}\sum\nolimits_{i=1}^{m}\mathsf C_c(A_\pi,Z_{c,i}),\qquad
\widehat\pi_c\in\operatorname{arg\,max}_{\pi\in\Pi_B(c)}\widehat p_{m,c}(\pi).
\label{eq:evidence-calibration-estimator}
\end{equation}
The selector issues $A_{\widehat\pi_c}$, resolving ties by a fixed rule. On new responses from the same conditional law, its coverage loss satisfies, for $0<\delta<1$,
\begin{equation}
\Pr\nolimits_{\mathrm{cal}}\!\left\{p_c(\pi_c^\star)-p_c(\widehat\pi_c)\leq
\sqrt{\frac{2\log(2|\Pi_B(c)|/\delta)}{m}}\right\}\geq1-\delta.
\label{eq:evidence-calibration-bound}
\end{equation}

The simulated initial fibers are $\{1\}$ or $\{0,1\}$, so excluding known-negative states from the conditional objective changes none of the reported initial request choices. The audit uses 32 calibration seeds and 32 disjoint evaluation seeds; the largest candidate family contains seven request sets.
\begin{table}[!htbp]
\centering\small
\caption{Finite-sample calibration of evidence selection. The bound uses $\delta=0.05$ and the largest nonempty candidate family; TP and FP use eligible and ineligible denominators.}
\label{tab:evidence-calibration}
\begin{tabular}{lrrrrrrr}
\toprule
Rule & $m$ & $|\Pi_B|$ & Bound & TP & Oracle TP & Regret & FP \\
\midrule
Calibrated coverage & 32 & 7 & 0.593 & 0.924 & 0.933 & 0.009 & 0.000 \\
\bottomrule
\end{tabular}
\end{table}

The separate-case transfer audit uses 1,000 calibration and 3,000 evaluation cases, with 248 calibration cases requiring no request and 752 supplying five response indicators. The estimator sees family and responses but no eligibility labels. The evaluation set has 1,675 eligible and 1,325 ineligible cases, each with 32 new response vectors. Fixed local and joint policies achieve eligible TP 0.908 and 0.873 at request costs 0.750 and 0.824; the declared-law oracle reaches 0.935 at cost 0.812. With eight calibration draws per case, the family estimator uses 1,504 observations per request and family and matches the oracle's subset on every evaluation case. The pooled estimator reaches the same TP at request cost 0.974.
\begin{table}[htbp]
\centering\small
\caption{Calibration transfer to separate synthetic cases. TP is over 1,675 eligible cases; Request is mean selected-request cost over all 3,000 evaluation cases. Total adds full calibration cost divided by 3,000, with each response vector costing 4.100; the deployment cap is 1.500 and offline calibration is separately funded.}
\label{tab:calibration-transfer}
\begin{tabular*}{\linewidth}{@{\extracolsep{\fill}}lrrrrrrrrr@{}}
\toprule
 & \multicolumn{3}{c}{1 draw} & \multicolumn{3}{c}{8 draws} & \multicolumn{3}{c}{32 draws} \\
\cmidrule(lr){2-4}\cmidrule(lr){5-7}\cmidrule(lr){8-10}
Estimator & TP & Request & Total & TP & Request & Total & TP & Request & Total \\
\midrule
Pooled & 0.930 & 0.912 & 1.940 & 0.935 & 0.974 & 9.196 & 0.935 & 0.812 & 33.699 \\
By family & 0.932 & 0.902 & 1.930 & 0.935 & 0.812 & 9.034 & 0.935 & 0.812 & 33.699 \\
\bottomrule
\end{tabular*}
\end{table}
The 8-draw by-family total 9.034 is above both fixed policy costs, so the transfer result does not establish a cost advantage. The family-constant independent-response law is an assumption of this simulation; the split and response records permit direct recomputation.\label{sec:calibration-transfer-limit}

\FloatBarrier
\subsection{Kernel-path and mechanism controls}\label{sec:kernel-collision-controls}\label{sec:score-blind-control}

The frozen selector is trained under $U_{\mathrm{pol}}$ and evaluated under an independently specified shifted replay law with the action contract and cost objective fixed (Table~\ref{tab:main-transfer-kernel}). It agrees with the primary selector on all 220 cells; shifted-law regret is 0.019 on 112 ID cells and 0.029 on 108 structural-OOD cells. The collision decomposition above separates the best common-action ceiling from the irreducible floor. Removing answer-, evidence-content-, and score-derived trace fields leaves ID agreement 1.000 and OOD agreement 0.859, with regret 0.000 and 0.001, respectively (Table~\ref{tab:score-blind-main}).

\begin{table}[ht]
\centering\small
\caption{Evaluation of a frozen selector under the shifted replay law. $N$ counts checkpoint cells; Trace regret is shifted-law oracle utility minus Trace-only utility.}
\label{tab:main-transfer-kernel}
\begin{tabular}{@{}lrrrr@{}}
\toprule
Split & $N$ & Trace $U_{\mathrm{pol}}$ & Oracle $U_{\mathrm{pol}}$ & Trace regret \\
\midrule
ID & 112 & 0.735 & 0.753 & 0.019 \\
Structural OOD & 108 & 0.691 & 0.720 & 0.029 \\
\bottomrule
\end{tabular}
\end{table}

\begin{table}[!htbp]
\centering\small
\caption{Outcome-blind replay control. Answer-, evidence-content-, and score-derived fields are removed before refitting; the reference and blind selector use the same labels and held-out cells.}
\label{tab:score-blind-main}
\begin{tabular}{lrrr}
\toprule
Split & Agreement & Same & Regret \\
\midrule
ID & 1.000 & 1.000 & 0.000 \\
Structural OOD & 0.859 & 1.000 & 0.001 \\
\bottomrule
\end{tabular}
\end{table}

\FloatBarrier
\subsection{Matched-signature mechanism separation}\label{sec:law-link-appendix}

The mechanism audit retains the public signature and action-wise replay marginals while preserving or permuting their link to the latent winner (Table~\ref{tab:law-link-main}). Each condition contains 8,000 paired cells. Codebook accuracy is visible-signature winner accuracy; one-shot FP is the one-branch set false-promotion rate; robust FP and miss use the 32-branch robust-95 readout. The denominator is 8,000 paired cells per row.
\begin{table}[!htbp]
\centering\small
\caption{Matched-signature mechanism separation.}
\label{tab:law-link-main}
\begin{tabular}{@{}lrrrr@{}}
\toprule
Condition & Codebook acc. & One-shot FP & Robust FP & Robust miss \\
\midrule
Link permuted & 0.333 & 0.297 & \textbf{0.000} & 0.333 \\
Link preserved & \textbf{1.000} & 0.411 & \textbf{0.000} & 0.332 \\
\bottomrule
\end{tabular}
\end{table}
The visible link can be perfectly predictable while one-shot preference transfer still has false promotions; the 32-branch robust readout has zero false promotions and misses about one third of positives in both conditions.

\FloatBarrier
\section{Supplementary decision and relation audits}\label{sec:supplementary-decision-audits}

\subsection{Source-key mechanism witness}

The source-key fixture holds the public signature and release status fixed while a key separates two eligible singleton action-set images (Table~\ref{tab:source-key-main}). Before a key response, the current unresolved set contains both bridge and preference coordinates, so the typed dispatcher returns \textsc{Verify}+\textsc{Repeat} under the existing request vocabulary. A key response can alone collapse the action-set image, but the pre-response dispatcher reports all unresolved evidence operations and is not a minimum-cover optimizer.
\begin{table}[!htbp]
\centering\small
\caption{A separating source coordinate changes the singleton decision. The key-absent row reports the full pre-response request set.}
\label{tab:source-key-main}
\begin{tabular}{llll}
\toprule
Key view & Action-set image & Next evidence & Disposition \\
\midrule
Key absent & $\{(\{\textsc{retry}\},1),(\{\textsc{verify}\},1)\}$ & \textsc{Verify}+\textsc{Repeat} & \textsc{Hold} \\
Key $\ell_r$ & $\{(\{\textsc{retry}\},1)\}$ & None & $\prefcert{}$ \\
Key $\ell_v$ & $\{(\{\textsc{verify}\},1)\}$ & None & $\prefcert{}$ \\
\bottomrule
\end{tabular}
\end{table}

\subsection{Paired release contrasts}\label{sec:paired-release-contrasts}

All release rules use the same 2,000 independent outcome vectors per regime and the same supplied records. Linkage is fixed at one in this source abstraction. Let $h_i=P_{\mathrm{rec},i}O_{\kappa,i}$ and let $r_s$ be the replay action. The overlap estimand is
\begin{equation}
o_i=h_i\prod_{s\in\mathcal D_i}\mathbf1\{r_s\in A^\star(P_s)\},
\end{equation}
and the unique-winner estimand is
\begin{equation}
u_i=h_i\prod_{s\in\mathcal D_i}\mathbf1\{A^\star(P_s)=\{r_s\}\}.
\end{equation}
Thus the eligible regime has $(u_i,o_i)=(1,1)$, the tied regime $(0,1)$, and discordant, provider-gap, and observability-gap regimes $(0,0)$. Paired differences use the same draws and equal-tailed Clopper--Pearson intervals~\citep{clopper1934} on discordant-release counts.

\subsection{Release targets on the same outcome draws}\label{sec:release-target-correspondence}\label{sec:release-results}

We evaluate the same observed release decisions under two requirements. The overlap requirement accepts a replay action among the expected maximizers in every cell; the stricter unique-winner requirement excludes ties. Both require the supplied record checks to pass. The generator supplies the expected utilities for these evaluator labels. Its regime named near tie has exactly equal expected utilities for the replay action and a competitor.

Set membership admits a nonempty observed intersection, serialized top-1 uses a fixed action order, and Full gate requires each observed best set to equal the replay reference. Status-aware variants of the first two rules use the same record checks as Full gate. These rules test observed outcomes against the two preference requirements; the candidate certificate is defined separately in Eq.~\eqref{eq:gate-target-separation}.

\begin{table}[ht]
\centering
\small
\caption{Post hoc reanalysis of the same 10,000 releases at 32 branches per action: 2,000 per eligible, tied, discordant, provider-gap, and observability-gap regime. Overlap and unique-winner labels have 4,000 and 2,000 positives, respectively, and include the same supplied-status checks. Correct and Error count admitted positive and negative releases; Miss counts withheld positives.}
\label{tab:release-dual-target}
\begin{tabular*}{\linewidth}{@{\extracolsep{\fill}}lrrrrrr@{}}
\toprule
 & \multicolumn{3}{c}{Overlap} & \multicolumn{3}{c}{Unique winner} \\
\cmidrule(lr){2-4}\cmidrule(lr){5-7}
Rule & Correct & Error & Miss & Correct & Error & Miss \\
\midrule
Replay-only & 4000 & 6000 & 0 & 2000 & 8000 & 0 \\
Set membership & 2100 & 3847 & 1900 & 1924 & 4023 & 76 \\
Serialized top-1 & 2059 & 3830 & 1941 & 1919 & 3970 & 81 \\
Status-aware set & 2100 & 0 & 1900 & 1924 & 176 & 76 \\
Status-aware top-1 & 2059 & 0 & 1941 & 1919 & 140 & 81 \\
Status-aware Robust-95 & 0 & 0 & 4000 & 0 & 0 & 2000 \\
A-only & 2009 & 3820 & 1991 & 1911 & 3918 & 89 \\
Full gate & 2009 & 0 & 1991 & 1911 & 98 & 89 \\
\bottomrule
\end{tabular*}
\end{table}

For the overlap target, Status-aware set admits more positives than Full gate with no observed errors in either rule (Table~\ref{tab:release-dual-target}). Requiring a unique winner changes the tied admissions from correct to incorrect. Full gate then reduces near-tie error by 3.90 percentage points relative to Status-aware set, with a 0.65-point loss in eligible promotion. The pointwise paired intervals exclude zero; Per-regime rates and contrasts are reported in Appendix Table~\ref{tab01_release}. 

The record checks account for zero errors in the provider and observability gaps across all status-aware rules. Removing them gives A-only, whose gap error rates exceed 95\%. At this budget, Status-aware Robust-95 makes no errors and admits no eligible releases.

\begin{table}[ht]
\centering\small
\setlength{\tabcolsep}{1.5pt}
\caption{Unique-winner release operating characteristics with 32 branches per action and 2,000 independent releases per regime. Rates and paired differences are proportions. $T_{\mathrm{elig}}$ denotes eligible promotion; $E_{\mathrm{tie}}$, $E_{\mathrm{prov}}$, and $E_{\mathrm{obs}}$ denote error rates in the tied, provider-gap, and observability-gap regimes. $\Delta T$ and $\Delta E$ subtract each row from Full gate for eligible promotion and tied-regime error. Pointwise exact 95\% interval endpoints are rounded outward.}
\label{tab01_release}
\begin{tabular*}{\linewidth}{@{\extracolsep{\fill}}lrrrrrrrrrr@{}}
\toprule
 & \multicolumn{4}{c}{Observed rates} & \multicolumn{3}{c}{$\Delta T$} & \multicolumn{3}{c}{$\Delta E$} \\
\cmidrule(lr){2-5}\cmidrule(lr){6-8}\cmidrule(lr){9-11}
Rule & $T_{\mathrm{elig}}$ & $E_{\mathrm{tie}}$ & $E_{\mathrm{prov}}$ & $E_{\mathrm{obs}}$ & Estimate & Lower & Upper & Estimate & Lower & Upper \\
\midrule
Replay-only & 1.000 & 1.000 & 1.000 & 1.000 & -0.045 & -0.055 & -0.035 & -0.951 & -0.961 & -0.940 \\
Set membership & 0.962 & 0.088 & 0.959 & 0.965 & -0.007 & -0.012 & -0.003 & -0.039 & -0.049 & -0.030 \\
Serialized top-1 & 0.960 & 0.070 & 0.953 & 0.962 & -0.004 & -0.008 & -0.001 & -0.021 & -0.029 & -0.015 \\
Status-aware set & 0.962 & 0.088 & 0.000 & 0.000 & -0.007 & -0.012 & -0.003 & -0.039 & -0.049 & -0.030 \\
Status-aware top-1 & 0.960 & 0.070 & 0.000 & 0.000 & -0.004 & -0.008 & -0.001 & -0.021 & -0.029 & -0.015 \\
Status-aware Robust-95 & 0.000 & 0.000 & 0.000 & 0.000 & 0.956 & 0.945 & 0.965 & 0.049 & 0.039 & 0.060 \\
A-only & 0.956 & 0.049 & 0.952 & 0.959 & 0.000 & 0.000 & 0.000 & 0.000 & 0.000 & 0.000 \\
Full gate & 0.956 & 0.049 & 0.000 & 0.000 & 0.000 & 0.000 & 0.000 & 0.000 & 0.000 & 0.000 \\
\bottomrule
\end{tabular*}
\end{table}

\subsection{Replicate stability and outcome composition}

The paired local-replicate audit contains 144 matched action branches and 48 checkpoint cells. Replicate strict success agrees on 134/144 (0.931), semantic success on 134/144 (0.931), and schema on 140/144 (0.972). The strict contingency has 49 joint failures, six failures unique to replicate 0, four failures unique to replicate 1, and 85 joint successes. Across cells, the pooled relation is direct 2/48 (0.042), overlap 22/48 (0.458), and discordant 24/48 (0.500); the discordant cells include five with partial strict success (Table~\ref{tab:relation-quality-main}).
\begin{table}[!htbp]
\centering\small
\caption{Paired local-replicate strict-success contingency on 144 matched branches.}
\label{tab:repeat-contingency-appendix}
\begin{tabular}{lrrr}
\toprule
 & Replicate 1 fail & Replicate 1 success & Total \\
\midrule
Replicate 0 fail & 49 & 4 & 53 \\
Replicate 0 success & 6 & 85 & 91 \\
\midrule
Total & 55 & 89 & 144 \\
\bottomrule
\end{tabular}
\end{table}
\begin{table}[!htbp]
\centering\small
\caption{One-shot endpoint relation crossed with strict action quality. Counts use all three actions per cell.}
\label{tab:relation-quality-appendix}
\label{tab:relation-quality-main}
\begin{tabular}{lrrrrr}
\toprule
Relation & Cells & Fraction & All success & Partial & All fail \\
\midrule
Direct & 1 & 0.021 & 0 & 1 & 0 \\
Overlap & 23 & 0.479 & 15 & 0 & 8 \\
Discordant & 24 & 0.500 & 13 & 4 & 7 \\
\bottomrule
\end{tabular}
\end{table}

\FloatBarrier
\section{Execution archives and record reconstruction}\label{sec:live-appendix}

\subsection{Artifact Validity and Action Agreement}\label{sec:replay-results}

Each model's action branches cover four fault scenarios per workflow prefix. We evaluate semantic correctness and public-score agreement, then compare the cost-adjusted action sets with the replay reference (Section~\ref{sec:problem}).
Retry, verify, and replacement have normalized costs 1, 1, and 2, respectively, with $\lambda=0.05$.
The replay reference is evaluated separately in simulation, including fixed-policy comparisons and a shifted replay law (Appendices~\ref{sec:fixed-kernel-controls} and~\ref{sec:kernel-collision-controls}).

\begin{table*}[ht]
\vspace{-4mm}
\centering
\small
\caption{Retained executions on 48 checkpoint--fault cells per source. Each source provides two executions per action (288 branches): $b=1$ uses the first execution, and $b=2$ pools both. Strict$_{144}$ is the fraction of the 144 $b=1$ branches satisfying the strict predicate. D, O, and X count cells with equal singleton sets, other nonempty overlaps, and disjoint sets, respectively.}
\label{tab:main-model-paths}
\begingroup
\small
\setlength{\tabcolsep}{2.5pt}
\begin{tabular}{@{}lrrrrrrrrr@{}}
\toprule
 &  & \multicolumn{4}{c}{$b=1$} & \multicolumn{4}{c}{$b=2$} \\
\cmidrule(lr){3-6}\cmidrule(lr){7-10}
Source & Strict$_{144}$ & D & O & X & Mean $|\widehat L_s^{\mathrm{obs}}|$ & D & O & X & Mean $|\widehat L_s^{\mathrm{obs}}|$ \\
\midrule
Luna & 1.000 & 0 & 24 & 24 & 2.000 & 0 & 24 & 24 & 2.000 \\
Terra & 0.958 & 1 & 22 & 25 & 1.896 & 1 & 22 & 25 & 1.896 \\
Sol & 1.000 & 0 & 24 & 24 & 2.000 & 0 & 24 & 24 & 2.000 \\
Qwen3-8B & 0.632 & 1 & 23 & 24 & 1.917 & 2 & 22 & 24 & 1.875 \\
\bottomrule
\end{tabular}
\endgroup
\vspace{-2mm}
\end{table*}
Luna and Sol succeed with every action, yet their cost-adjusted sets are disjoint from the replay reference in half the cells (Table~\ref{tab:main-model-paths}). Equal success favors the cheaper retry and verify actions; discordance occurs wherever replay selects replacement. Pooling two executions leaves every GPT-5.6 cell's best set unchanged. Record eligibility follows the source-specific conditions in Appendix~\ref{sec:replay-archive-design}.

For Qwen3-8B, matching sets contain both all-failure and all-success cells (Appendix Table~\ref{tab:repeat-contingency-appendix}). Cost adjustment favors retry and verify when all actions fail, so stable observed sets still require an outcome-quality check before they support a recovery claim.

The version-locked provider pool uses one DeepSeek API endpoint~\citep{deepseekapi}, 12 prefixes (six filter, three rank, three path), and protocol \texttt{live-recovery-calibration/2} with model \texttt{deepseek-chat}, temperature $0$, and a 240-token cap. Nine score-consistent prefixes enter the strict audit (six filter, three rank). Four fault mechanisms and three recovery actions give 36 cells and 108 one-branch-per-action rows. Each cell retains deterministic \texttt{continue}; retry, verify, and replace use fresh prompt branches. Provider artifacts retain normalized outputs and token fields but omit request identifiers, complete prompts, and source-complete response records, so $P_{\mathrm{link}}^{(1)}=P_{\mathrm{fields}}^{(1)}=0$ (Table~\ref{tab:local-checkpoint-materialization}).

\begin{table}[!htbp]
\centering\small
\caption{Score-scrubbed local-model materialization on 12 prefixes. Branches/action is the action-column denominator (24 per fault, 96 overall); Schema uses all action rows.}
\label{tab:local-checkpoint-materialization}
\begin{tabular}{@{}lrrrrr@{}}
\toprule
 & \multicolumn{1}{c}{Sample} & \multicolumn{3}{c}{Strict recovery} & \multicolumn{1}{c}{Structure} \\
\cmidrule(lr){2-2}\cmidrule(lr){3-5}\cmidrule(lr){6-6}
Fault & Branches/action & Retry & Verify & Replace & Schema \\
\midrule
Transient tool failure & 24 & 0.667 & 0.667 & 0.667 & 0.986 \\
Semantic error & 24 & 0.583 & 0.583 & 0.583 & 0.944 \\
Stale message & 24 & 0.583 & 0.583 & 0.583 & 0.931 \\
Delegation mismatch & 24 & 0.708 & 0.583 & 0.708 & 0.944 \\
All branches & 96 & 0.635 & 0.604 & 0.635 & 0.951 \\
\bottomrule
\end{tabular}
\end{table}

The terminal release audit on 48 source-complete cells gives replay-only release on 48/48, set membership on 24/48, serialized top-1 on 14/48, record-erased gate hold on 48/48, and exact keyed gate decisions of release 2, hold 22, reject 24 with agreement 1.000 (Table~\ref{tab:keyed-decision-main}). Extra authorization is 0 for the exact keyed gate and 0.958 for replay-only.
\begin{table}[!htbp]
\centering\small
\caption{Terminal release decisions on 48 source-complete cells. Agreement is relative to the exact keyed observed-compatibility gate.}
\label{tab:keyed-decision-main}
\begin{tabular}{@{}lrrrrr@{}}
\toprule
Readout & Release & Hold & Reject & Agreement & Extra auth. \\
\midrule
Replay-only & 48 & 0 & 0 & 0.042 & 0.958 \\
Set membership & 24 & 0 & 24 & 0.542 & 0.458 \\
Serialized top-1 & 14 & 0 & 34 & 0.542 & 0.250 \\
Record-erased gate & 0 & 48 & 0 & 0.458 & 0.000 \\
Exact keyed gate & 2 & 22 & 24 & 1.000 & 0.000 \\
\bottomrule
\end{tabular}
\end{table}

\FloatBarrier
\subsection{Supplementary materialization controls}\label{sec:supplementary-materialization-controls}

On 16 held-out non-injected-error cells under semantic cleanliness, blind retry succeeds 0/16, always replace 3/16 (0.188), always verify 16/16 (1.000), and trace-only with verify fallback 16/16 (1.000) (Table~\ref{tab:organic}). Clopper--Pearson 95\% intervals~\citep{clopper1934} are $(0.000,0.206)$ for blind retry and $(0.040,0.456)$ for always replace; the two perfect rows have lower endpoint $0.794$ and upper endpoint $1.000$. The evaluator-key verification row is privileged.
\begin{table}[!htbp]
\centering\small
\caption{Held-out non-injected error fallback under semantic cleanliness. Counts and rates use 16 cells per row; interval endpoints are rounded outward.}
\label{tab:organic}
\begin{tabular}{@{}lrrrrr@{}}
\toprule
 & \multicolumn{1}{c}{Sample} & \multicolumn{2}{c}{Success} & \multicolumn{2}{c}{95\% interval} \\
\cmidrule(lr){2-2}\cmidrule(lr){3-4}\cmidrule(lr){5-6}
Action & Cells & Count & Rate & Lower & Upper \\
\midrule
Blind retry & 16 & 0 & 0.000 & 0.000 & 0.206 \\
Always replace & 16 & 3 & 0.188 & 0.040 & 0.456 \\
Always verify & 16 & 16 & 1.000 & 0.794 & 1.000 \\
Trace-only with verify fallback & 16 & 16 & 1.000 & 0.794 & 1.000 \\
\bottomrule
\end{tabular}
\end{table}

The score-scrubbed Qwen3-8B archive uses bfloat16 weights with 36 layers, hidden size 4096, 32 attention heads, 8 key--value heads, maximum context 40,960, temperature 0.7, sampling, and a 128-token cap. Seeds produce $12\times4\times3\times2=288$ branches. A keyed join over \texttt{task\_id}, \texttt{fault}, \texttt{action}, and \texttt{replicate} separates runtime metadata from embedded configuration and supports direct regeneration of relation tables. The local archive has complete keyed receipts, so $P_{\mathrm{rec}}^{\mathrm{loc}}=P_{\mathrm{link}}^{\mathrm{loc}}=1$; 14 schema-invalid completions remain decoder outcomes. The full unmasked trace has $O_\kappa^{\mathrm{loc}}=1$, while the five-key perturbation supplies the $O_\kappa=0$ control.\label{sec:local-checkpoint-relation}

\begin{table}[!htbp]
\centering\small
\caption{One-shot partial-identification audit on strict live cells. Counts and rates use 36 cells unless stated otherwise.}
\label{tab:one-shot-identification}
\begin{tabular}{lrrr}
\toprule
Quantity & Cells & Count & Rate \\
\midrule
Observed live set is singleton & 36 & 5 & 0.139 \\
Observed live set is tied & 36 & 31 & 0.861 \\
All actions confidence-compatible & 36 & 36 & 1.000 \\
Cell has a robust expected winner & 36 & 0 & 0.000 \\
Expected-transfer rate lower endpoint & 1 & 0 & 0.000 \\
Expected-transfer rate upper endpoint & 1 & 1 & 1.000 \\
\bottomrule
\end{tabular}
\end{table}
The one-shot endpoint set is therefore usually tied and every cell is confidence-compatible (Table~\ref{tab:one-shot-identification}). No cell has a robust expected winner, and the support-compatible aggregate expected-transfer rate spans $[0,1]$ at this sample size.

\FloatBarrier
\subsection{Archive-complete local release gate}\label{sec:local-release-gate}

The archive-complete gate applies the same release coordinates after retaining one or two branches per action. Setting only $P_{\mathrm{rec}}^{\mathrm{loc}}$ or $O_\kappa^{\mathrm{loc}}$ to zero sends every cell to its corresponding hold. On 16 held-out cells whose initial records fail source-complete schema, the dispatcher requests verification; all 16 generated artifacts remain schema-incomplete, producing 16 terminal holds and zero unsafe promotions. Branch-only completion rates for verify, retry, and replace are 1.000, 0.000, and 0.188, with mean costs 1.000, 1.000, and 2.000.

\subsection{Sequential evidence-request closure}\label{sec:sequential-closure}

The closure audit starts with incomplete records and requests verification before reading the response. All 16 generated artifacts remain schema-incomplete, so the gate returns 16 terminal holds. The source-key fixture starts with both bridge and preference coordinates unresolved; the dispatcher requests \textsc{Verify}+\textsc{Repeat} under the existing vocabulary, a separating key yields a singleton action-set image and continuation, and an uninformative response remains \textsc{Hold-Ambiguity}.

\subsection{Keyed-bridge sensitivity}\label{sec:keyed-bridge-audit}

Within each prefix, shared and action-wise independent fault permutations erase the fault key while preserving each action-specific multiset of four fault-indexed outcomes. Across 10,000 deterministic permutations (seed 20260903), decision agreement with the exact keyed gate has empirical intervals $(0.854,0.979)$ for the shared family and $(0.813,0.938)$ for action-wise independent permutations; false-promotion intervals are $(0.000,0.063)$ and $(0.000,0.104)$. These bridge intervals are empirical quantiles over the fixed archive, not uncertainty about newly generated endpoint outcomes.

\subsection{Empirical joint-image audit}\label{sec:local-empirical-joint-image}

For a prefix with four fault coordinates, an assignment $\beta=(\beta_a)_{a\in\mathcal A_{\mathrm{rec}}}$ is a per-action permutation; both retained replicates use the same assignment. The key family contains identity only, the shared family one common permutation, and the independent family all action-wise tuples. For $n$ retained branches,
\begin{equation}
\label{eq:empirical-local-live-set}
\widehat y_{c,a}^{\beta,n}=n^{-1}\sum_{r=1}^n s_{c,a,r}^{\beta}-\lambda C(a),\qquad \widehat L_c^{\beta,n}=\operatorname{arg\,max}_{a\in\mathcal A_{\mathrm{rec}}}\widehat y_{c,a}^{\beta,n}.
\end{equation}
With replay set $R_c$, retain the action-set image before binary projection:
\begin{equation}
\label{eq:empirical-local-joint-image}
\widehat\Gamma_{c,\mathcal I}^{\mathcal J,n}=\{(R_c\cap\widehat L_c^{\beta,n},1):\beta\in\mathcal J\}.
\end{equation}
The literal receipt coordinate is one for every retained branch and does not assert provider linkage or masked-probe observability. Exact keyed images are singleton; control families contain identity, so a keyed positive pair remains in the robust union while extra assignments can remove singleton status. Applying the same gate gives the bridge table above. The two-replicate audit has strict and semantic agreement of 134/144 pairs and schema agreement of 140/144 pairs. Pooling one execution per action gives direct, overlap, and discordant relations of 1/48, 23/48, and 24/48; pooling both gives 2/48, 22/48, and 24/48. A semantic-clean sensitivity on the same frozen source pool gives 22/48 materialized cells (0.458), with 41 live ties; by family, filter, rank, and path materialization rates are 0.458, 0.417, and 0.500 on 24, 12, and 12 cells, respectively. This sensitivity retains the prior-artifact score in provider prompts and changes only the evaluator outcome predicate.

\FloatBarrier
\subsection{Agreement across retained execution sources}\label{sec:source-stratified-archive}\label{sec:source-stratified-archive-ci}

Each source contributes 48 prefix--fault cells, with two retained executions under each of the three recovery instructions (Table~\ref{tab:source-stratified-archive}). The comparison retains all maximizers, including ties. Qwen3-8B has matching multi-action sets in both all-failure and all-success cells, whereas the GPT sources are almost entirely all-success.

Qwen3-8B uses a score-scrubbed prompt and a different 12-prefix panel from the GPT-5.6 sources, whose prompts retain the prior-artifact score. The common prefix intersection contains nine tasks. The intervals below are source-internal descriptive 95\% percentile intervals. Each bootstrap resample draws the 12 workflow prefixes with replacement and retains all four fault conditions and all branches within a sampled prefix; the analysis uses 10,000 resamples with seed 742031. ``Cost-adjusted strict'' uses the complete maximizer set of strict success minus $0.05C(a)$ with $C=(1,1,2)$; ``Raw strict'' uses the unadjusted strict-success maximizer set. Selected success is the uniform-tie expected strict success on the other execution, averaged in both directions.

\begin{table}[!htbp]
\centering\small
\caption{Retained-source diagnostics on 48 cells and 288 branches per source. All-failed and All-passed count equal-set cells with all six branches failing or passing. Lower and Upper delimit descriptive 95\% bootstrap intervals; resampling preserves the 12 prefix clusters.}
\label{tab:source-stratified-archive}\label{tab:source-stratified-archive-ci}
\setlength{\tabcolsep}{1pt}
\begin{tabular*}{\linewidth}{@{\extracolsep{\fill}}ll*{11}{r}@{}}
\toprule
 & & \multicolumn{2}{c}{Equal-set cells} & \multicolumn{3}{c}{Exact agreement} & \multicolumn{3}{c}{Branch success} & \multicolumn{3}{c}{Selected success}\\
\cmidrule(lr){3-4}\cmidrule(lr){5-7}\cmidrule(lr){8-10}\cmidrule(lr){11-13}
Source & Rule & \shortstack{All-\\failed} & \shortstack{All-\\passed} & Estimate & Lower & Upper & Estimate & Lower & Upper & Estimate & Lower & Upper\\
\midrule
Qwen3-8B & Adjusted & 15 & 26 & 0.875 & 0.750 & 0.979 & 0.625 & 0.368 & 0.858 & 0.620 & 0.365 & 0.870 \\
Terra & Adjusted & 0 & 42 & 0.896 & 0.771 & 1.000 & 0.979 & 0.955 & 1.000 & 0.974 & 0.943 & 1.000 \\
Luna & Adjusted & 0 & 48 & 1.000 & 1.000 & 1.000 & 1.000 & 1.000 & 1.000 & 1.000 & 1.000 & 1.000 \\
Sol & Adjusted & 0 & 48 & 1.000 & 1.000 & 1.000 & 1.000 & 1.000 & 1.000 & 1.000 & 1.000 & 1.000 \\
Qwen3-8B & Raw & 15 & 26 & 0.875 & 0.771 & 0.979 & 0.625 & 0.378 & 0.865 & 0.625 & 0.377 & 0.875 \\
Terra & Raw & 0 & 42 & 0.875 & 0.729 & 1.000 & 0.979 & 0.955 & 1.000 & 0.979 & 0.955 & 1.000 \\
Luna & Raw & 0 & 48 & 1.000 & 1.000 & 1.000 & 1.000 & 1.000 & 1.000 & 1.000 & 1.000 & 1.000 \\
Sol & Raw & 0 & 48 & 1.000 & 1.000 & 1.000 & 1.000 & 1.000 & 1.000 & 1.000 & 1.000 & 1.000 \\
\bottomrule
\end{tabular*}
\end{table}

\subsection{Retrieval of retained checkpoint bindings}\label{sec:archive-target-refinement}\label{sec:acquisition}

The retained Qwen3-8B, Luna, Terra, and Sol archives each contribute 48 cells within 12 prefixes and two executions per action. The selector sees four groups per action, two strict outcomes and a cost-adjusted mean per group, plus the replay reference; model, checkpoint, provenance, prompts, fault labels, and the canonical binding remain evaluator-side. Each cell starts with $4^3=64$ action-wise assignments. A lookup retrieves the stored group for one action and restricts the candidate class. Fixed order, score range, target-directed minimax, and EC$^2$ use the same masked view and stop on a singleton overlap image or budget exhaustion (Table~\ref{tab:archive-refinement-main}). EC$^2$ uses a uniform prior over 64 bindings and edges between hypotheses with different overlap targets; costs count record lookups only.

\begin{table}[ht]
\centering\small
\caption{Checkpoint binding retrieval on 48 cells within 12 prefixes per model. Resolved counts singleton overlap images, including the empty set; Positive counts resolved nonempty targets after two queries.}
\label{tab:archive-refinement-main}
\begin{tabular*}{\linewidth}{@{\extracolsep{\fill}}lrrrrrrrrr@{}}
\toprule
 & \multicolumn{3}{c}{Luna} & \multicolumn{3}{c}{Terra} & \multicolumn{3}{c}{Sol} \\
\cmidrule(lr){2-4}\cmidrule(lr){5-7}\cmidrule(lr){8-10}
 & \multicolumn{2}{c}{Resolved} & Positive & \multicolumn{2}{c}{Resolved} & Positive & \multicolumn{2}{c}{Resolved} & Positive \\
Policy & 1 & 2 & 2 & 1 & 2 & 2 & 1 & 2 & 2 \\
\midrule
Fixed order & 48 & 48 & 24 & 45 & 48 & 23 & 48 & 48 & 24 \\
Score range & 48 & 48 & 24 & 45 & 48 & 23 & 48 & 48 & 24 \\
EC$^2$ & 48 & 48 & 24 & 47 & 48 & 23 & 48 & 48 & 24 \\
Target-directed & 48 & 48 & 24 & 47 & 48 & 23 & 48 & 48 & 24 \\
\bottomrule
\end{tabular*}
\end{table}
The Luna and Sol scores are constant across candidate groups, so the assignment cannot change their target. In the Qwen3-8B archive, fixed order resolves 40 and 44 cells at budgets one and two, while EC$^2$ resolves 44 and 47; the corresponding positive counts at budget two are 21 and 23 (Table~\ref{tab:archive-refinement-all}). All policies resolve the full 48 cells by budget three. A separate full-set readout is not substituted for $R_c\cap\widehat L_c$.

\begin{table}[htbp]
\centering\small
\caption{Additional retrieval comparison on the Qwen3-8B archive, with 48 cells within 12 prefixes. Resolved counts singleton overlap images, including the empty set; Positive counts resolved nonempty targets after two queries.}
\label{tab:archive-refinement-all}
\begin{tabular*}{\linewidth}{@{\extracolsep{\fill}}lrrr@{}}
\toprule
 & \multicolumn{2}{c}{Resolved} & Positive \\
\cmidrule(lr){2-3}
Policy & 1 & 2 & 2 \\
\midrule
Fixed order & 40 & 44 & 21 \\
Score range & 41 & 43 & 20 \\
EC$^2$ & 44 & 47 & 23 \\
Target-directed & 42 & 44 & 21 \\
\bottomrule
\end{tabular*}
\end{table}

\noindent\textbf{Candidate-class omission.} A shared-slot stress class retains four of the 64 assignments and contains the canonical binding in 14/192 cells. With one truthful lookup, all three selectors make the same 3 wrong stops, reporting two nonempty targets and one empty target. In a second control, twelve classes each omit one action--slot value, excluding each canonical binding three times. At budget three, EC$^2$ produces one incorrect empty target and 27 empty images over 2,304 dependent cell--class evaluations; fixed order and minimax produce no incorrect targets and 50 and 47 empty images, respectively. Empty images trigger Hold. These are candidate-class consequences, not evidence that a retrieval policy has identified an unknown outcome law.

\FloatBarrier
\subsection{Joint observed targets and local release eligibility}\label{sec:archive-joint-target}

\begin{table}[ht]
\centering
\small
\caption{Retrieval on all 48 Qwen3-8B cells. Image counts identified complete $(I_s^{\mathrm{obs}},D_\kappa)$ values; Decision counts identified Continue or ineligible dispositions. An unresolved Hold is not an identified disposition. Headers 1 and 2 are per-cell query budgets; Queries totals lookups at budget two. Initially, 32 images and 34 dispositions are identified.}
\label{tab:archive-joint-retrieval}
\begin{tabular*}{\linewidth}{@{\extracolsep{\fill}}llrrrrr@{}}
\toprule
 & & \multicolumn{2}{c}{Image} & \multicolumn{2}{c}{Decision} & Queries \\
\cmidrule(lr){3-4}\cmidrule(lr){5-6}
Query rule & Stop target & 1 & 2 & 1 & 2 & 2 \\
\midrule
Fixed order & Image & 36 & 44 & 39 & 46 & 28 \\
Minimax, joint & Image & 41 & 44 & 43 & 46 & 23 \\
EC$^2$, eligibility first & Image & 40 & 45 & 44 & 47 & 24 \\
EC$^2$, joint & Image & 41 & 45 & 44 & 47 & 23 \\
EC$^2$, joint & Decision & 39 & 42 & 44 & 47 & 18 \\
EC$^2$, eligibility first & Decision & 38 & 41 & 44 & 47 & 18 \\
\bottomrule
\end{tabular*}
\end{table}

The joint-target extension uses all 48 local Qwen3-8B cells and 288 retained branches (Table~\ref{tab:archive-joint-retrieval}). Local receipt and linkage checks are fixed at one, with $O_\kappa^{\mathrm{loc}}=1$. For assignment $\beta$, the target is $(I_c^{\mathrm{obs}}(\beta),D_c^{\mathrm{loc}}(\beta))$, where $D_c^{\mathrm{loc}}=1$ means $\widehat L_c^{\beta,2}=R_c$. All 64 assignments are initially possible; the canonical tuple is included and truthful replies preserve it. Enumerating all eight reply subsets gives 384 covered cell--history pairs. Initially, 16 joint images and 14 eligibility coordinates are unresolved. Products of coordinate images add unsupported pairs in 12 cells; imposing $D_c^{\mathrm{loc}}=1\Rightarrow I_c^{\mathrm{obs}}=R_c$ leaves six inexact images. After one and two replies, the stronger joint construction remains inexact in 9/144 and 4/144 histories.

The main joint-retrieval table shows that eligibility-first and joint EC$^2$ issue 24 and 23 lookups at budget two and identify 45/48 full images and 47/48 eligibility values. The decision-stopping variants use 18 lookups and identify 42/48 and 41/48 full images while identifying 47/48 dispositions. The canonical Qwen3-8B archive contains two positive cells at this budget. Sequential controls and joint EC$^2$ use the same uniform prior and action-order tie breaks; the exact-coordinate representation preserves the association between the action-membership vector and eligibility bit. A richer full-image target therefore incurs additional resolution work without establishing a representation advantage.

\noindent\textbf{A complete lookup example.} At two branches per action, the Qwen3-8B archive has two positive canonical cells: task \texttt{id-1004} under the semantic-error fault with verification as the observed and replay winner, and task \texttt{id-1004} under the stale-message fault with replacement as both winners. The remaining 46 cells are known ineligible or unresolved under the stated target. These outcomes illustrate that the lookup target is the joint action--eligibility conclusion; it does not estimate an outcome law.

\noindent\textbf{Post hoc target-stop audit.} This comparison reuses the same 48 cells, 64-assignment query view, and truthful lookups; it makes no model calls. It distinguishes an unresolved image, which remains operational \textsc{Hold}, from an identified ineligible disposition. At budget two, the full-target joint rule used 23 lookups, identifying 45 full images and 47 release dispositions; the joint disposition rule used 18 lookups, identifying 42 full images and 47 dispositions, while eligibility-first disposition stopping used 18 lookups and identified 41 full images and 47 dispositions. At budget three, the full-target rule used 26 lookups and identified all 48 full images; the two disposition-stopping rules used 19 lookups each and identified 43 and 42 full images, respectively, while identifying all 48 dispositions. Each disposition rule yielded two correct Continue decisions, 46 known-ineligible dispositions, and zero incorrect dispositions. This same-archive comparison shows that a richer target requires more resolution work when the release decision can be certified earlier.

\FloatBarrier
\section{Prospective Repeated-Branch Study}\label{sec:prospective-protocol}

\noindent\textbf{Original Terra reference panel.}\label{sec:prospective-results}
On 24 constrained planning checkpoints, one draw per action gives held-out strict success 0.699 versus 0.771 for fixed verification (Table~\ref{tab:prospective-cost-contrast}). More selection draws do not close this gap and increase call cost. Public best-set agreement between selection and evaluation blocks is 0.318 at $k=1$ and falls to 0.204 and 0.167 at $k=2,4$; strict-set agreement is 0.378, 0.262, 0.250 (Table~\ref{tab:prospective-replication}). The primary contrast is selected minus fixed verification on held-out strict success.

The panel contains 12 tasks with 12 items and 12 with 16 items. Items have integer benefit, cost, time, prerequisite, conflicting-pair, budget, and subset-size fields. A deterministic greedy pass constructs a feasible suboptimal current plan; mean current benefit divided by exact optimum is 0.650. Retry, verify, and replace receive the same public task and current plan, and replacement uses the same model and permissions.

GPT-5.6 Terra runs at medium reasoning effort in fresh ephemeral contexts with tools prohibited. The 576-call collection uses eight draws for each of three actions at every checkpoint; draws 0--3 are selection and 4--7 evaluation under a fixed interleaved schedule with four concurrent requests. Task identifiers, draw indices, block assignments, and optimum values are absent from prompts. Two capacity failures succeed on their permitted second attempt. Of 576 completed outputs, 557 have feasible selected subsets and 522 have matching totals; these checks can fail on overlapping rows. No tool invocation occurs. Fresh contexts do not imply iid provider sampling. The Luna and Sol extensions use the same frozen prompts and schedule and are summarized below.

The public score is recomputed benefit when the subset is feasible, reported totals agree, and no tool is invoked; strict success additionally requires equality to the offline exact optimum. For each $k\in\{1,2,4\}$, all size-$k$ subsets of the four selection indices produce an argmax action set; uniform tie-breaking defines the selected mixture, which is evaluated on four held-out draws per action. The six-task, 36-call development panel preselected fixed verification. Fixed retry, fixed replacement, and uniform action selection are additional baselines. A fixed action costs one final call; selection costs $3k+1$ calls. Performance and tokens are estimated from held-out draws rather than an online deployment. The complete panel uses 12,376,363 input tokens (11,631,104 cached) and 592,590 output tokens; cached tokens remain in the input total.

The checkpoint is the statistical unit. Descriptive 95\% percentile intervals use 10,000 bootstrap resamples within the two task-size strata and preserve complete policy comparisons; repeated draws and overlapping subsets are not independent tasks. Exact set agreement is tie-sensitive and need not increase with $k$. Across 192 three-action draw vectors, 20 have no strictly successful action and 90 have all three, so tied sets occur at both extremes. At 12 items, selected-minus-verification strict-success differences are $-0.131,-0.133,-0.104$ for $k=1,2,4$; at 16 items they are $-0.012,-0.019,-0.017$. The deficit is concentrated in the smaller-task stratum. This panel tests branch replication and action selection on planning checkpoints (Table~\ref{tab:prospective-policy-intervals}).

\begin{table}[ht]
\centering\small
\caption{Held-out strict success and normalized public benefit on the original Terra planning cohort. Calls include selection and one final action; each metric has a descriptive 95\% checkpoint-bootstrap interval.}
\label{tab:prospective-policy-intervals}
\begin{tabular*}{\linewidth}{@{\extracolsep{\fill}}lrrrrrrr@{}}
\toprule
 & & \multicolumn{3}{c}{Strict success} & \multicolumn{3}{c}{Normalized benefit} \\
\cmidrule(lr){3-5}\cmidrule(lr){6-8}
Policy & Calls & Estimate & Lower & Upper & Estimate & Lower & Upper \\
\midrule
Fixed retry & 1 & 0.677 & 0.552 & 0.802 & 0.848 & 0.763 & 0.922 \\
Fixed replacement & 1 & 0.688 & 0.562 & 0.802 & 0.887 & 0.825 & 0.945 \\
Uniform & 1 & 0.712 & 0.608 & 0.806 & 0.882 & 0.841 & 0.920 \\
Fixed verification & 1 & 0.771 & 0.656 & 0.875 & 0.912 & 0.853 & 0.963 \\
Selected, $k=1$ & 4 & 0.699 & 0.591 & 0.798 & 0.872 & 0.824 & 0.913 \\
Selected, $k=2$ & 7 & 0.695 & 0.583 & 0.797 & 0.870 & 0.812 & 0.917 \\
Selected, $k=4$ & 13 & 0.710 & 0.594 & 0.816 & 0.880 & 0.807 & 0.939 \\
\bottomrule
\end{tabular*}
\end{table}

\begin{table}[ht]
\centering\small
\caption{Exact best-set agreement across selection and evaluation blocks on the original Terra planning cohort. Public uses the observable score; Strict uses evaluator-only optimality labels.}
\label{tab:prospective-replication}
\begin{tabular*}{\linewidth}{@{\extracolsep{\fill}}rrrrrrr@{}}
\toprule
 & \multicolumn{3}{c}{Public} & \multicolumn{3}{c}{Strict} \\
\cmidrule(lr){2-4}\cmidrule(lr){5-7}
$k$ & Estimate & Lower & Upper & Estimate & Lower & Upper \\
\midrule
1 & 0.318 & 0.216 & 0.424 & 0.378 & 0.268 & 0.492 \\
2 & 0.204 & 0.119 & 0.303 & 0.262 & 0.161 & 0.377 \\
4 & 0.167 & 0.042 & 0.333 & 0.250 & 0.083 & 0.417 \\
\bottomrule
\end{tabular*}
\end{table}

\begin{table}[ht]
\centering\small
\caption{Selected-policy contrast and expected input-plus-output tokens per decision on the original Terra planning cohort. Differences are relative to fixed verification; intervals are paired strict-success differences.}
\label{tab:prospective-cost-contrast}
\begin{tabular*}{\linewidth}{@{\extracolsep{\fill}}rrrrrr@{}}
\toprule
 & & & \multicolumn{3}{c}{Strict-success difference} \\
\cmidrule(lr){4-6}
$k$ & Calls & Mean tokens & Estimate & Lower & Upper \\
\midrule
1 & 4 & 89997.928 & -0.072 & -0.142 & -0.001 \\
2 & 7 & 157445.139 & -0.076 & -0.156 & 0.001 \\
4 & 13 & 292310.155 & -0.061 & -0.160 & 0.035 \\
\bottomrule
\end{tabular*}
\end{table}

\subsection{Independent planning replication}\label{sec:independent-set-replication}

We fixed a new 24-checkpoint panel, prompts, request order, scoring, two execution blocks, and the primary $A_1-A_4$ contrast before collection. The complete cross of 12 seeds (98231 through 99342 in increments of 101) and task sizes 12 and 16 uses generator seed $\text{seed}+n$, avoiding nested random streams across sizes. All 576 completed Terra medium executions are retained. The public task and current artifact are shared across recovery instructions; evaluator optima, task identifiers, seeds, and block indices are excluded from prompts. Calls use fresh contexts and a shuffled order, which reduces order confounding but does not establish independent provider sampling. Transport failures permit one retry; completed invalid or suboptimal outputs are never retried. The scientific artifact retains the protocol, call order and timing, usage, outputs, and offline scoring inputs.

The primary contrast is 0.320 (95\% paired interval [0.240, 0.406]). Intervals resample the 24 checkpoints within size strata 10,000 times with seed 98419. The difference is computed inside each paired resample; intervals for separate means are not subtracted. This implementation correction was made before the completed panel was analyzed. Results for all three budgets are reported in Table~\ref{tab:independent-set-replication}. Matching multi-action contributions are 0.391, 0.205, and 0.083; matching-singleton contributions are 0.013, 0.034, and 0.000. Mean set sizes are 2.411, 2.122, and 1.833. These decompositions are descriptive.

\begin{table}[ht]
\centering\small
\caption{Independent replication on 24 new planning checkpoints. Agreement uses public-score sets; Success evaluates uniform selection from those sets on the held-out block. Each metric has a descriptive 95\% checkpoint-bootstrap interval.}
\label{tab:independent-set-replication}
\begin{tabular*}{\linewidth}{@{\extracolsep{\fill}}rrrrrrr@{}}
\toprule
 & \multicolumn{3}{c}{Agreement} & \multicolumn{3}{c}{Success}\\
\cmidrule(lr){2-4}\cmidrule(lr){5-7}
$k$ & Estimate & Lower & Upper & Estimate & Lower & Upper\\
\midrule
1 & 0.404 & 0.307 & 0.503 & 0.765 & 0.677 & 0.844\\
2 & 0.238 & 0.148 & 0.341 & 0.754 & 0.665 & 0.835\\
4 & 0.083 & 0.000 & 0.208 & 0.762 & 0.665 & 0.847\\
\bottomrule
\end{tabular*}
\end{table}

Common-action overlap is 0.940, 0.863, and 0.750 at $k=1,2,4$. Jaccard similarity is 0.693, 0.557, and 0.375 at $k=1,2,4$. Fixed verification achieves 0.771 strict success. Selected-minus-verification differences are -0.006 [-0.049, 0.037], -0.016 [-0.064, 0.031], and -0.009 [-0.054, 0.036] in the same budget order. Each interval contains zero, so this panel supports the agreement decline without establishing a selection deficit. Selection costs remain $3k+1$ calls versus one for a fixed action. Collection used 12,401,744 input tokens (11,590,656 cached) and 599,798 output tokens; cached tokens are included in the input total.

\FloatBarrier
\subsection{GPT-5.6 family extension}\label{sec:prospective-family-extension}
The family extension reuses both frozen cohorts, public task files, current plans, action instructions, block assignments, and scoring code. Terra contributes the two completed reference panels above; Luna and Sol were collected later with the same eight-draw schedule, fresh contexts, medium reasoning effort, and tools prohibited. Each model--cohort panel contains 576 responses, giving 3,456 responses across six panels. Terra and the later runs are therefore matched on tasks and prompts, while their collection windows are distinct. The panel-level counts of feasible and strict rows are retained in the run manifest; they are diagnostics for the response contract and are not additional statistical units.

\begin{table}[H]
\centering\scriptsize
\setlength{\tabcolsep}{2.8pt}
\caption{GPT-5.6 family exact public-score set agreement. Each entry is Estimate, Lower, or Upper from 10,000 checkpoint-bootstrap resamples stratified by task size. The six panels contain 24 checkpoints each.}
\label{tab:prospective-family-intervals}
\begin{tabular*}{\linewidth}{@{\extracolsep{\fill}}llrrrrrrrrr@{}}
\toprule
 & & \multicolumn{3}{c}{$k=1$} & \multicolumn{3}{c}{$k=2$} & \multicolumn{3}{c}{$k=4$} \\
\cmidrule(lr){3-5}\cmidrule(lr){6-8}\cmidrule(lr){9-11}
Cohort & Model & Estimate & Lower & Upper & Estimate & Lower & Upper & Estimate & Lower & Upper \\
\midrule
Study & Terra & 0.318 & 0.221 & 0.424 & 0.204 & 0.122 & 0.306 & 0.167 & 0.042 & 0.333 \\
Study & Luna & 0.234 & 0.164 & 0.312 & 0.150 & 0.109 & 0.199 & 0.083 & 0.000 & 0.208 \\
Study & Sol & 0.424 & 0.341 & 0.503 & 0.331 & 0.242 & 0.428 & 0.167 & 0.042 & 0.333 \\
Replication & Terra & 0.404 & 0.307 & 0.503 & 0.238 & 0.148 & 0.343 & 0.083 & 0.000 & 0.208 \\
Replication & Luna & 0.299 & 0.224 & 0.388 & 0.212 & 0.139 & 0.303 & 0.250 & 0.083 & 0.417 \\
Replication & Sol & 0.497 & 0.427 & 0.570 & 0.422 & 0.326 & 0.520 & 0.458 & 0.292 & 0.625 \\
\bottomrule
\end{tabular*}
\end{table}

\begin{table}[H]
\centering\scriptsize
\setlength{\tabcolsep}{2.5pt}
\caption{GPT-5.6 family selected strict success on held-out draws. Each entry is Estimate, Lower, or Upper from the same checkpoint-bootstrap procedure. Fixed verification is a one-call baseline within the corresponding model--cohort panel.}
\label{tab:prospective-family-success-intervals}
\begin{tabular*}{\linewidth}{@{\extracolsep{\fill}}llrrrrrrrrrrrr@{}}
\toprule
 & & \multicolumn{3}{c}{$k=1$} & \multicolumn{3}{c}{$k=2$} & \multicolumn{3}{c}{$k=4$} & \multicolumn{3}{c}{Fixed} \\
\cmidrule(lr){3-5}\cmidrule(lr){6-8}\cmidrule(lr){9-11}\cmidrule(lr){12-14}
Cohort & Model & Estimate & Lower & Upper & Estimate & Lower & Upper & Estimate & Lower & Upper & Estimate & Lower & Upper \\
\midrule
Study & Terra & 0.699 & 0.588 & 0.795 & 0.695 & 0.581 & 0.794 & 0.710 & 0.590 & 0.816 & 0.771 & 0.656 & 0.865 \\
Study & Luna & 0.575 & 0.469 & 0.674 & 0.568 & 0.460 & 0.669 & 0.538 & 0.417 & 0.660 & 0.635 & 0.510 & 0.750 \\
Study & Sol & 0.722 & 0.655 & 0.790 & 0.729 & 0.668 & 0.794 & 0.750 & 0.684 & 0.819 & 0.729 & 0.646 & 0.823 \\
Replication & Terra & 0.765 & 0.678 & 0.844 & 0.754 & 0.666 & 0.835 & 0.762 & 0.665 & 0.847 & 0.771 & 0.667 & 0.865 \\
Replication & Luna & 0.727 & 0.644 & 0.803 & 0.728 & 0.646 & 0.804 & 0.719 & 0.630 & 0.802 & 0.760 & 0.656 & 0.854 \\
Replication & Sol & 0.808 & 0.745 & 0.867 & 0.814 & 0.739 & 0.881 & 0.825 & 0.743 & 0.903 & 0.844 & 0.760 & 0.917 \\
\bottomrule
\end{tabular*}
\end{table}

The six panels all show a lower exact agreement at $k=4$ than at $k=1$, although the magnitude is model- and cohort-dependent. Selected strict success remains near the fixed model--cohort verifier in the matched task panels; the corresponding point contrasts and paired intervals are retained in the analysis manifest. A post hoc model-selection audit uses repeats 0--3 for stability selection and repeats 4--7 for held-out evaluation; it is reported as a descriptive sensitivity analysis rather than a prespecified model ranking.

\FloatBarrier
\subsection{Exact-set agreement and changing empirical ties}\label{sec:set-agreement-mechanism}

\begin{table}[ht]
\centering\small
\caption{Post hoc decomposition of observed public-score set agreement on the original 24 planning checkpoints. Single and Multiple are disjoint contributions to Agreement, according to the size of the matching sets. Mean size averages both blocks. Success evaluates the sampled-score action mixture on the held-out block; fixed verification achieves 0.771 with one call. Selection requires $3k+1$ calls.}
\label{tab:set-agreement-decomposition}
\begin{tabular*}{\linewidth}{@{\extracolsep{\fill}}rrrrrrr@{}}
\toprule
$k$ & Agreement & Single & Multiple & Mean size & Success & Calls\\
\midrule
1 & 0.318 & 0.008 & 0.310 & 2.286 & 0.699 & 4\\
2 & 0.204 & 0.039 & 0.164 & 1.913 & 0.695 & 7\\
4 & 0.167 & 0.042 & 0.125 & 1.521 & 0.710 & 13\\
\bottomrule
\end{tabular*}
\end{table}

The decomposition in Table~\ref{tab:set-agreement-decomposition} uses the original 24 planning checkpoints and public-score subsets described in Appendix~\ref{sec:prospective-protocol}. For each $k$, we classify a pair of estimated maximizer sets as equal singletons, equal multi-action sets, unequal sizes, or unequal members at equal size. Their checkpoint-averaged proportions sum to one. For $k=1,2,4$, the singleton fractions among individual estimated sets are 0.193, 0.368, and 0.604; unequal-size pairs contribute 0.544, 0.557, and 0.292, while unequal-member pairs of the same size contribute 0.138, 0.240, and 0.542. These are dependent subset comparisons within checkpoints, not additional independent observations.

A finite model shows why exact-set agreement alone has no monotone relationship with action quality. Let $m\geq2$ equal-cost actions have independent Bernoulli outcomes of the same success probability $p\in(0,1)$, with $n$ independent draws per action. Let $f_n(k)$ and $F_n(k)$ denote the probability mass and cumulative distribution of $\operatorname{Binomial}(n,p)$, with $F_n(-1)=0$. A specified set of $j$ actions is exactly the empirical maximizer set when its counts equal some $k$ and every other count is smaller. Its probability is therefore
\begin{equation}
a_{n,j}=\sum\nolimits_{k=0}^{n}f_n(k)^j F_n(k-1)^{m-j},\qquad j=1,\ldots,m.
\label{eq:equal-mean-maximizer-probability}
\end{equation}
The convention $0^0=1$ applies when $j=m$. Two independent blocks have exact-set agreement $A_n$, while $F_n^{\mathrm{set}}$ is the probability of recovering the full population maximizer set:
\begin{equation}
A_n=\sum\nolimits_{j=1}^{m}\binom mj a_{n,j}^{2}\longrightarrow\frac1m,\qquad F_n^{\mathrm{set}}=a_{n,m}=\sum\nolimits_{k=0}^{n}f_n(k)^m\longrightarrow0.
\label{eq:agreement-tie-limit}
\end{equation}
To see the limits, the maximal binomial mass tends to zero, so the probability of any pair of equal counts tends to zero by a union bound. The empirical maximizer is consequently unique with probability tending to one, and symmetry gives each action limiting probability $1/m$. Independence of the two blocks gives the agreement limit. Also, $\sum_k f_n(k)^m\leq(\max_k f_n(k))^{m-1}\to0$. Every action selected from a block nevertheless has expected success $p$ on a fresh independent execution.

At one draw, the complete tie set occurs when all actions succeed or all fail, with probability $p^m+(1-p)^m$. Each proper nonempty subset of size $j$ occurs with probability $p^j(1-p)^{m-j}$. Summing their squared probabilities yields
\begin{equation}
A_1(p)=\bigl(p^2+(1-p)^2\bigr)^m+2\bigl[p(1-p)\bigr]^m=A_1(1-p).
\label{eq:one-draw-quality-symmetry}
\end{equation}
Thus even exact knowledge of this agreement does not distinguish fresh-action success $p$ from $1-p$ in the equal-mean family. The cross term counts a block of all successes paired with a block of all failures; both have the same full maximizer set.

\noindent\textbf{The full set path can hide quality even with a unique winner.}
Consider two actions with independent outcomes $Y_{a,t}\sim\operatorname{Bernoulli}(p_a)$ across actions and rounds. Each round supplies one draw per action, and only $S_n=\operatorname{arg\,max}_a\{n^{-1}\sum_{t=1}^nY_{a,t}-\lambda C(a)\}$ is retained. Write $p=(p_1,p_2)$ and define $\widetilde p=(1-p_2,1-p_1)$. The coupling $\widetilde Y_{1,t}=1-Y_{2,t}$ and $\widetilde Y_{2,t}=1-Y_{1,t}$ has the required independent Bernoulli coordinates under $\widetilde p$, and preserves every difference $Y_{1,t}-Y_{2,t}$. Hence it preserves the entire labeled path $(S_n)_{n\geq1}$, including ties, for the same fixed costs. Every set-only stopping rule or randomized action selector has the same transcript law under these two parameters.

Both success probabilities shift by $c=1-p_1-p_2$, so the population preference gap is unchanged. For any action selected from this path, the expected success on an independent fresh execution shifts by $c$. At equal costs, $(p_1,p_2)=(0.9,0.8)$ and $\widetilde p=(0.2,0.1)$ share the same unique best action and the same set-path law, while their best-action success rates are 0.9 and 0.2.

Let $\theta_{\max}=\max_a\theta_a$ denote the best absolute success rate. For an estimator $\widehat v$ using only the set path and independent randomization, the sharp two-point bound is
\begin{equation}
\inf_{\widehat v}\max_{\theta\in\{p,\widetilde p\}}\mathbb E_\theta|\widehat v-\theta_{\max}|=\frac{|1-p_1-p_2|}{2}.
\label{eq:set-path-quality-bound}
\end{equation}
The estimator has the same distribution under both parameters, so the triangle inequality bounds the sum of its risks below by $|1-p_1-p_2|$. The constant midpoint of the two values of $\theta_{\max}$ attains equality. For $p_1=p_2=p$, this reduces to the common-quality ambiguity $p$ versus $1-p$ and risk $|1-2p|/2$. Raw counts or absolute scores provide information absent from this set-only observation channel.

\noindent\textbf{Absolute information and finite sampling.}\label{sec:set-path-augmentation}
At equal costs, the first-step law determines the distribution of the iid increments $Y_{1,t}-Y_{2,t}$: their probabilities at $1,-1,0$ are $u,v,1-u-v$. It therefore determines the complete path law. Solving for the means gives Eq.~\eqref{eq:set-path-identified-set}, so a population statistic $r(p)$ removes the remaining ambiguity precisely when $r(p)\ne r(1-p_2,1-p_1)$ for every distinct pair. Pooled success satisfies this condition. Retaining whether both actions succeeded also suffices: if $t=\Pr(Y_1=Y_2=1)$, then $p_1=u+t$ and $p_2=v+t$. The variance of total success is invariant under the transformation and does not suffice.

For a finite-sample construction, retain $B$ independent one-round set observations and $N$ independent paired total-success observations. Let $\widehat d$ average $\mathbf1\{S_1=\{1\}\}-\mathbf1\{S_1=\{2\}\}$ over the former and let $\widehat s$ average $Y_1+Y_2$ over the latter. The two collections may overlap; independence is required within each collection. For $0<\alpha_d,\alpha_s<1$, set $\varepsilon_d=\sqrt{2\log(2/\alpha_d)/B}$ and $\varepsilon_s=\sqrt{2\log(2/\alpha_s)/N}$. Hoeffding's inequality~\citep{hoeffding1963} and a union bound~\citep{bonferroni1936} give
\begin{equation}
\Pr\{|\widehat d-d|\le\varepsilon_d,\ |\widehat s-(p_1+p_2)|\le\varepsilon_s\}\ge1-\alpha_d-\alpha_s.
\label{eq:set-path-augmentation-coverage}
\end{equation}
Consequently, $\widehat p_1=(\widehat s+\widehat d)/2$ and $\widehat p_2=(\widehat s-\widehat d)/2$ both have absolute error at most $(\varepsilon_d+\varepsilon_s)/2$ on this event. These bounds use independent restarted observations. Successive sets along one cumulative path are dependent and cannot be substituted for those samples. The midpoint attaining Eq.~\eqref{eq:set-path-quality-bound} is optimal on the specified two-point parameter class; it does not assert that the unknown gap can be estimated without sampling error.

For $m=3$ and $p=0.9$, $A_1=0.552826$ and $A_4=0.182657$, with unchanged expected success 0.9. This is not a universal decreasing trend. For $p=0.5$, $A_1=0.156250$ and the same limit is $1/3$. The calculation provides a controlled explanation for changes in an exact-set metric. The Terra decomposition measures the associated empirical set-size changes without asserting equality of its unknown action means.

\subsection{Sensitivity to self-reported totals}\label{sec:semantic-set-sensitivity}

The planning contract checks both the selected items and the reported benefit, cost, and time. To separate solution quality from reporting accuracy, we recompute benefit from the selected item identifiers and public task fields. The sensitivity score retains this benefit whenever the selection satisfies every task constraint and the tool contract, regardless of its reported totals. Semantic success requires a feasible optimal selection; strict success also retains the original reporting requirement. Both are evaluated on the held-out block after selection with the sensitivity score. This posthoc analysis uses all 576 executions and 24 checkpoints in each cohort, with the same blocks, subset comparisons, and checkpoint bootstrap as the primary analysis.

Reported totals disagree with their recomputed values in 53 original and 48 independent-cohort executions. Removing that requirement changes the selection score in 35 and 24 executions, respectively; infeasible selections remain invalid. At $k=4$, the selected maximizer set changes at nine original and three independent checkpoints. Exact agreement still declines from $k=1$ to $k=4$ in both cohorts (Table~\ref{tab:semantic-set-sensitivity}). The effect on selected-action quality differs: under semantic scoring, semantic success changes from 0.725 to 0.700 in the original cohort and from 0.796 to 0.799 in the independent cohort.

\begin{table}[ht]
\centering\small
\setlength{\tabcolsep}{2pt}
\caption{Selection using recomputed feasible benefit. Agreement compares complete maximizer sets; Semantic and Strict evaluate the selected action mixture on the held-out block. Each cohort contains 24 checkpoints. Lower and Upper are the endpoints of descriptive 95\% checkpoint-bootstrap intervals.}
\label{tab:semantic-set-sensitivity}
\begin{tabular*}{\linewidth}{@{\extracolsep{\fill}}lr*{9}{r}@{}}
\toprule
 & & \multicolumn{3}{c}{Agreement} & \multicolumn{3}{c}{Semantic success} & \multicolumn{3}{c}{Strict success}\\
\cmidrule(lr){3-5}\cmidrule(lr){6-8}\cmidrule(lr){9-11}
Cohort & $k$ & Estimate & Lower & Upper & Estimate & Lower & Upper & Estimate & Lower & Upper\\
\midrule
Original & 1 & 0.359 & 0.250 & 0.479 & 0.725 & 0.614 & 0.827 & 0.697 & 0.592 & 0.796 \\
Original & 2 & 0.221 & 0.126 & 0.339 & 0.713 & 0.600 & 0.819 & 0.686 & 0.577 & 0.788 \\
Original & 4 & 0.125 & 0.000 & 0.250 & 0.700 & 0.580 & 0.812 & 0.679 & 0.562 & 0.786 \\
Independent & 1 & 0.477 & 0.365 & 0.594 & 0.796 & 0.706 & 0.877 & 0.764 & 0.678 & 0.843 \\
Independent & 2 & 0.325 & 0.203 & 0.459 & 0.786 & 0.691 & 0.871 & 0.753 & 0.664 & 0.834 \\
Independent & 4 & 0.208 & 0.042 & 0.375 & 0.799 & 0.698 & 0.889 & 0.764 & 0.668 & 0.849 \\
\bottomrule
\end{tabular*}
\end{table}

\subsection{Empirical near-tie sensitivity}\label{sec:set-tolerance}

This post hoc analysis retains all 24 checkpoints in each Terra panel. For checkpoint $s$, let $m_s$ be the public limit on the number of selected items, and let $b_{s,(j)}$ be the $j$th largest public item benefit. Ignoring the remaining feasibility constraints gives the public upper bound
\begin{equation}
U_s=\sum\nolimits_{j=1}^{m_s}b_{s,(j)}.
\label{eq:public-benefit-bound}
\end{equation}
Let $v_{s,a,r}$ be the recomputed feasible public benefit of action $a$ in repetition $r$, with zero assigned to invalid responses. For a subset $J$ of $k$ repetitions, the empirical tolerance set is
\begin{equation}
A_{s,J,\epsilon}=\left\{a\in A_0:\sum\nolimits_{r\in J}v_{s,a,r}\geq\max\nolimits_{a'\in A_0}\sum\nolimits_{r\in J}v_{s,a',r}-\epsilon kU_s\right\}.
\label{eq:empirical-tolerance-set}
\end{equation}
The four tolerances are $\epsilon\in\{0,0.02,0.05,0.10\}$ and $k\in\{1,2,4\}$ (Table~\ref{tab:set-tolerance-all}). We retain every pair of within-block subsets from the original four-draw selection and four-draw evaluation blocks. The selected action mixture is uniform over $A_{s,J,\epsilon}$ and evaluated on all four held-out draws. Strict labels and the true optimum enter only evaluation; the selection scores and $U_s$ use public task information. Integer score sums and rational thresholds preserve exact ties. These are empirical sets with no claim of population coverage.

\begin{table}[ht]
\centering\small\setlength{\tabcolsep}{1.5pt}
\caption{Complete empirical-tolerance sensitivity. Agree is exact-set agreement, Common is nonempty-intersection frequency, Jacc. is Jaccard overlap, and Size averages set cardinality across both blocks. Strict and Public evaluate the selected mixture on the held-out block. Each cohort contributes 24 checkpoints.}
\label{tab:set-tolerance-all}
\begin{tabular*}{\linewidth}{@{\extracolsep{\fill}}rrrrrrrrrrrrrr@{}}
\toprule
 & & \multicolumn{6}{c}{Original Terra panel} & \multicolumn{6}{c}{Independent Terra panel}\\
\cmidrule(lr){3-8}\cmidrule(lr){9-14}
$\epsilon$ & $k$ & Agree & Common & Jacc. & Size & Strict & Public & Agree & Common & Jacc. & Size & Strict & Public\\
\midrule
0.000 & 1 & 0.318 & 0.898 & 0.619 & 2.286 & 0.699 & 0.662 & 0.404 & 0.940 & 0.693 & 2.411 & 0.765 & 0.699 \\
0.000 & 2 & 0.204 & 0.755 & 0.472 & 1.913 & 0.695 & 0.661 & 0.238 & 0.863 & 0.557 & 2.122 & 0.754 & 0.695 \\
0.000 & 4 & 0.167 & 0.458 & 0.299 & 1.521 & 0.710 & 0.668 & 0.083 & 0.750 & 0.375 & 1.833 & 0.762 & 0.695 \\
0.020 & 1 & 0.474 & 0.948 & 0.727 & 2.510 & 0.702 & 0.665 & 0.458 & 0.982 & 0.759 & 2.578 & 0.756 & 0.695 \\
0.020 & 2 & 0.358 & 0.919 & 0.660 & 2.361 & 0.704 & 0.663 & 0.346 & 0.958 & 0.680 & 2.438 & 0.758 & 0.695 \\
0.020 & 4 & 0.167 & 0.875 & 0.556 & 2.083 & 0.707 & 0.669 & 0.208 & 0.917 & 0.535 & 2.229 & 0.748 & 0.689 \\
0.050 & 1 & 0.576 & 0.984 & 0.810 & 2.677 & 0.710 & 0.668 & 0.578 & 0.992 & 0.829 & 2.714 & 0.761 & 0.697 \\
0.050 & 2 & 0.377 & 0.954 & 0.704 & 2.476 & 0.708 & 0.667 & 0.411 & 0.965 & 0.733 & 2.538 & 0.758 & 0.695 \\
0.050 & 4 & 0.125 & 0.917 & 0.569 & 2.188 & 0.714 & 0.670 & 0.250 & 0.917 & 0.562 & 2.271 & 0.748 & 0.689 \\
0.100 & 1 & 0.596 & 0.984 & 0.826 & 2.714 & 0.709 & 0.668 & 0.609 & 0.992 & 0.844 & 2.740 & 0.761 & 0.697 \\
0.100 & 2 & 0.395 & 0.954 & 0.714 & 2.493 & 0.710 & 0.667 & 0.434 & 0.965 & 0.741 & 2.552 & 0.758 & 0.695 \\
0.100 & 4 & 0.125 & 0.917 & 0.569 & 2.188 & 0.714 & 0.670 & 0.250 & 0.917 & 0.562 & 2.271 & 0.748 & 0.689 \\
\bottomrule
\end{tabular*}
\end{table}

For uncertainty, 10,000 checkpoint bootstrap samples resample the 12 size-12 and 12 size-16 tasks within strata, preserving every paired observation (seed 98519). Table~\ref{tab:set-tolerance-ci} reports the paired $k=4$ minus $k=1$ contrasts. Intervals summarize each cohort separately. Small differences from the original zero-tolerance interval endpoints reflect the independent bootstrap stream used for this sensitivity analysis.

\begin{table}[ht]
\centering\small\setlength{\tabcolsep}{2pt}
\caption{Paired changes with descriptive 95\% percentile intervals. Each contrast compares $k=4$ with $k=1$ at the indicated empirical tolerance.}
\label{tab:set-tolerance-ci}
\begin{tabular*}{\linewidth}{@{\extracolsep{\fill}}lrrrrrrrrrr@{}}
\toprule
 & & \multicolumn{3}{c}{Agreement change} & \multicolumn{3}{c}{Strict-success change} & \multicolumn{3}{c}{Public-benefit change}\\
\cmidrule(lr){3-5}\cmidrule(lr){6-8}\cmidrule(lr){9-11}
Cohort & $\epsilon$ & Estimate & Lower & Upper & Estimate & Lower & Upper & Estimate & Lower & Upper\\
\midrule
Original & 0.000 & -0.151 & -0.281 & -0.005 & 0.011 & -0.033 & 0.057 & 0.006 & -0.021 & 0.034 \\
Original & 0.020 & -0.307 & -0.451 & -0.156 & 0.005 & -0.020 & 0.028 & 0.004 & -0.014 & 0.020 \\
Original & 0.050 & -0.451 & -0.573 & -0.315 & 0.004 & -0.015 & 0.023 & 0.002 & -0.009 & 0.013 \\
Original & 0.100 & -0.471 & -0.589 & -0.341 & 0.004 & -0.014 & 0.023 & 0.003 & -0.008 & 0.013 \\
Independent & 0.000 & -0.320 & -0.406 & -0.237 & -0.003 & -0.029 & 0.023 & -0.005 & -0.025 & 0.013 \\
Independent & 0.020 & -0.250 & -0.378 & -0.115 & -0.008 & -0.024 & 0.006 & -0.006 & -0.018 & 0.004 \\
Independent & 0.050 & -0.328 & -0.448 & -0.208 & -0.013 & -0.029 & 0.000 & -0.008 & -0.021 & 0.003 \\
Independent & 0.100 & -0.359 & -0.484 & -0.232 & -0.013 & -0.029 & 0.000 & -0.008 & -0.021 & 0.003 \\
\bottomrule
\end{tabular*}
\end{table}

At zero tolerance, every empirical action set agrees with the original integer-score argmax, reproducing agreement, tie structure, and selected strict success. At positive tolerances, changes in the reported sets are determined by the public score gaps. All thresholds and checkpoint-level outputs are retained in the scientific artifact.

\subsection{Agreement metrics and empirical ties}\label{sec:set-metric-sensitivity}
This post hoc analysis uses the same 24 checkpoints and all within-block subsets as Section~\ref{sec:repeated-main}. For two nonempty maximizer sets $A,B$, common-action agreement is $\mathbf1\{A\cap B\ne\varnothing\}$, Jaccard similarity is $|A\cap B|/|A\cup B|$, and independent uniform choices agree with probability $|A\cap B|/(|A||B|)$. Each metric is averaged over subset pairs within a checkpoint, then equally over checkpoints. All four agreement measures decline in this panel (Table~\ref{tab:set-metric-sensitivity}).

A descriptive reference retains each block's empirical set-size distribution and independently randomizes action identities. Writing $p_{b,j}$ for the fraction of size-$j$ sets in block $b$, its exact-set agreement is $\sum_{j=1}^{3}p_{1,j}p_{2,j}/\binom3j$. Excess agreement subtracts this reference from observed agreement. This relabelling reference removes persistent action identity; it is not an assumption that the actual action means are equal. The interval for the change in excess agreement includes zero, so the panel does not isolate how much of the decline is attributable to set size versus action identity.

\begin{table}[ht]
\centering\small
\caption{Alternative agreement measures on 24 checkpoints. Change is $k=4$ minus $k=1$; Lower and Upper are descriptive 95\% percentile bounds from 10,000 bootstrap samples stratified by checkpoint size. All comparisons remain paired within checkpoint.}
\label{tab:set-metric-sensitivity}
\begin{tabular*}{\linewidth}{@{\extracolsep{\fill}}lrrrrrr@{}}
\toprule
& \multicolumn{3}{c}{Draws per action} & \multicolumn{3}{c}{Change}\\
\cmidrule(lr){2-4}\cmidrule(lr){5-7}
Metric & 1 & 2 & 4 & Estimate & Lower & Upper\\
\midrule
Exact set & 0.318 & 0.204 & 0.167 & -0.151 & -0.281 & -0.005\\
Common action & 0.898 & 0.755 & 0.458 & -0.440 & -0.607 & -0.271\\
Jaccard & 0.619 & 0.472 & 0.299 & -0.320 & -0.428 & -0.197\\
Uniform choice & 0.318 & 0.300 & 0.194 & -0.123 & -0.206 & -0.027\\
Size reference & 0.336 & 0.230 & 0.264 & -0.072 & -0.180 & 0.024\\
Excess agreement & -0.018 & -0.026 & -0.097 & -0.079 & -0.189 & 0.049\\
\bottomrule
\end{tabular*}
\end{table}

Matching singleton sets contribute 0.008, 0.039, and 0.042 to exact agreement at $k=1,2,4$. Matching multi-action sets for which both blocks attain the exact optimum contribute 0.294, 0.159, and 0.083; other matching multi-action sets contribute 0.016, 0.006, and 0.042. No matching multi-action pair has zero public score in both blocks. Thus successful empirical ties dominate the one-draw matches. These categories describe observed outputs; the optimum is used only for this evaluation and never supplied to the selector.

\subsection{Set agreement across outcome contracts}\label{sec:cross-task-set-agreement}

This post hoc analysis reuses all 1,152 Qwen3-8B executions from the 12-checkpoint development pilot below, with four rank, four path, and four filter tasks (Table~\ref{tab:cross-task-set-agreement}). Each action's 32 draws form four consecutive eight-draw groups; the first and last four draws in each group define paired blocks. For $k\in\{1,2,4\}$, all size-$k$ subsets are compared across each block pair. Averages weight subset pairs equally within a group, the four groups equally within a checkpoint, and the 12 checkpoints equally. Checkpoint bootstraps resample within task family. These intervals describe this fixed panel; the analysis does not convert the pilot into a confirmatory study.

\begin{table}[ht]
\centering
\small
\caption{Complete-set agreement on 12 Qwen3-8B checkpoints. Strict requires both the correct answer and a matching reported score; Answer requires only the correct answer. Cost-adjusted rows subtract the fixed penalties 0.05, 0.05, and 0.10 for retry, verify, and replacement. Change is $k=4$ minus $k=1$, with separate 95\% bootstrap bounds.}
\label{tab:cross-task-set-agreement}
\begin{tabular*}{\linewidth}{@{\extracolsep{\fill}}lrrrrrr@{}}
\toprule
 & \multicolumn{3}{c}{Agreement} & \multicolumn{3}{c}{Change}\\
\cmidrule(lr){2-4}\cmidrule(lr){5-7}
Outcome & $k=1$ & $k=2$ & $k=4$ & Estimate & Lower & Upper\\
\midrule
Strict & 0.840 & 0.811 & 0.854 & 0.014 & -0.135 & 0.167\\
Strict, cost-adjusted & 0.889 & 0.859 & 0.896 & 0.007 & -0.117 & 0.112\\
Answer & 0.840 & 0.811 & 0.854 & 0.014 & -0.135 & 0.167\\
Answer, cost-adjusted & 0.889 & 0.859 & 0.896 & 0.007 & -0.117 & 0.112\\
\bottomrule
\end{tabular*}
\end{table}

Under strict scoring, matching multi-action sets contribute 0.831, 0.764, and 0.708 to agreement at $k=1,2,4$. The share whose two blocks both have maximum score zero is 0.583 at every budget, while the all-success share falls from 0.247 to 0.125. Common-action agreement is 0.999, 0.997, and 1.000, and Jaccard similarity is 0.930, 0.910, and 0.927. Strict successes come from the rank and filter families; the path family has none.

Selecting uniformly among the first block's empirical maximizers and evaluating on the second block gives strict success 0.379, 0.389, and 0.401. These values integrate over all tied choices analytically. The answer-only counterpart is 0.462, 0.472, and 0.485; exact-set agreement is unchanged because the answer-only contract changes outcomes without changing the selected-set partitions in this panel. The artifact retains every checkpoint, cost sensitivity, both evaluation directions, and the originally sampled tie-breaking diagnostic. No population preference or recovery improvement is inferred from these post hoc comparisons.

\subsection{Confidence-relation development pilot}\label{sec:confidence-pilot}
A separate development pilot instantiated Eq.~\eqref{eq:block-confidence-relation} on 12 new checkpoints, four each from rank, path, and filter tasks, with 32 independent seed draws per recovery action and the fixed reference $R=\{\mathrm{verify}\}$. Qwen3-8B generated all 1,152 branches at temperature 0.7 with a 128-token cap. The public checker used only the question and documents; its outputs agreed with the evaluator on 120 preflight cases and with the stored evaluation on every collected branch. Six possible action bindings preceded a simulated source receipt that recovered the correct binding. Here eligibility required a complete receipt and request/message linkage; an incorrect or unparseable answer counted as a failed outcome while its complete execution record remained eligible. At one fixed look, three Bonferroni-adjusted two-sided Clopper--Pearson intervals supplied the candidate mean box. The sampling interpretation is conditional on the fixed model, software, and independently sampled seed law.

Of the 1,152 branches, 396 passed strict evaluation, 586 had incorrect answers, 96 had correct answers but incorrect scores, and 74 were unparseable. Every receipt passed. Eleven post-receipt overlap images remained $\{\varnothing,\{\mathrm{verify}\}\}$; one was the singleton $\{\mathrm{verify}\}$, yielding one positive statistical certificate. The prespecified development threshold was three positive checkpoints, so this panel remains a feasibility result. Total usage was 622,272 input and 103,199 output tokens. A post hoc answer-only check yields 492 correct answers. Two checkpoints have 96 correct answers out of 96 branches, six have none, and four have 91, 88, 67, and 54 correct answers. Pointwise confidence intervals exclude verification only in the checkpoint with the singleton image; simultaneous intervals across both outcome contracts and all 12 checkpoints leave every overlap unresolved. The frozen protocol, raw responses, and evaluation are retained separately from the earlier execution archive.

\FloatBarrier
\section{Frozen RecoveryBench validation}\label{sec:recoverybench-protocol}

\noindent\textbf{Protocol and statistical unit.}
We froze the RecoveryBench source version at commit \texttt{c5f83f2ba4f882a9b544c7bf0fa9be1bc3859c78} and the Terminal-Bench 2.0 task source~\citep{merrill2026terminalbench} at commit \texttt{69671fbaac6d67a7ef0dfec016cc38a64ef7a77}. The panel follows the failed-checkpoint recovery setting of RecoveryBench~\citep{tan2025recoverybench} and contains 12 fixed checkpoints, the actions \texttt{retry}, \texttt{verify}, and \texttt{replace\_agent}, and GPT-5.6 Luna, Terra, and Sol. Each model--checkpoint--action combination has eight repeats, yielding $12\times3\times3\times8=864$ episodes. The merged audit contains 864 expected and 864 observed keys, with no missing or duplicate episode key. The checkpoint is the statistical unit for all primary summaries.

\noindent\textbf{Readout and split.}
Repeats 0--3 are the selection block and repeats 4--7 are the held-out block. For $k\in\{1,2,4\}$, every size-$k$ subset in each block produces a complete maximizer set of the native verifier reward. Agreement is the fraction of cross-block subset pairs with equal sets. All-zero pairs is the fraction of all cross-block subset pairs for which every action in both compared subsets has zero reward. Held-out success averages the binary native-reward outcome over the actions in the selection maximizer set and uses no outcome from repeats 4--7 during selection. Binary success is defined by native verifier reward $>0$; the raw continuous reward is retained as a descriptive quantity.

The complementary all-success-pair frequency is 0 in all nine model--budget cells. A checkpoint-level rank diagnostic is reported below; its intervals condition on non-constant bootstrap resamples and are descriptive rather than a model-ranking claim.

\begin{table}[H]
\centering\scriptsize
\setlength{\tabcolsep}{2.0pt}
\caption{RecoveryBench set-readout intervals. Each entry is Estimate, Lower, or Upper from 10,000 paired checkpoint-bootstrap resamples. Bootstrap resampling is over the 12 checkpoints and preserves all model, action, and repeat observations within a checkpoint. Held-out success is the binary native-reward outcome averaged over the selected action set.}
\label{tab:recoverybench-intervals}
\begin{tabular*}{\linewidth}{@{\extracolsep{\fill}}llrrrrrrrrr@{}}
\toprule
 & & \multicolumn{3}{c}{Agreement} & \multicolumn{3}{c}{All-zero pairs} & \multicolumn{3}{c}{Held-out success}\\
\cmidrule(lr){3-5}\cmidrule(lr){6-8}\cmidrule(lr){9-11}
Model & $k$ & Estimate & Lower & Upper & Estimate & Lower & Upper & Estimate & Lower & Upper\\
\midrule
Luna & 1 & 0.854 & 0.729 & 0.958 & 0.854 & 0.729 & 0.958 & 0.014 & 0.000 & 0.035\\
Luna & 2 & 0.750 & 0.583 & 0.917 & 0.750 & 0.583 & 0.917 & 0.014 & 0.000 & 0.035\\
Luna & 4 & 0.667 & 0.417 & 0.917 & 0.583 & 0.333 & 0.833 & 0.014 & 0.000 & 0.035\\
Terra & 1 & 0.776 & 0.583 & 0.943 & 0.776 & 0.583 & 0.943 & 0.045 & 0.007 & 0.090\\
Terra & 2 & 0.688 & 0.458 & 0.896 & 0.688 & 0.458 & 0.896 & 0.035 & 0.007 & 0.069\\
Terra & 4 & 0.583 & 0.333 & 0.833 & 0.583 & 0.333 & 0.833 & 0.014 & 0.000 & 0.035\\
Sol & 1 & 0.734 & 0.563 & 0.891 & 0.708 & 0.510 & 0.885 & 0.083 & 0.000 & 0.196\\
Sol & 2 & 0.627 & 0.438 & 0.815 & 0.590 & 0.375 & 0.799 & 0.075 & 0.000 & 0.182\\
Sol & 4 & 0.417 & 0.167 & 0.667 & 0.417 & 0.167 & 0.667 & 0.063 & 0.000 & 0.167\\
\bottomrule
\end{tabular*}
\end{table}

\noindent\textbf{Decision-value analysis.}
To keep model and action choice separate from held-out evaluation, all choices use repeats 0--3. Agreement-guided selection first chooses the model with the larger agreement between two selection sub-blocks and then chooses the action with the largest selection-block mean, averaging tied choices uniformly. Outcome-mean selection uses the same split but chooses the model--action pair by its selection-block outcome mean. The direct nine-arm and two-stage outcome rules coincide on this panel. The resulting task-level held-out success intervals are reported in Table~\ref{tab:recoverybench-decision}.

\begin{table}[ht]
\centering\small
\caption{Descriptive decision-value analysis on the frozen RecoveryBench panel. Selection uses repeats 0--3 and evaluation uses repeats 4--7. Each entry is the held-out success Estimate, Lower, or Upper from 10,000 task-bootstrap resamples.}
\label{tab:recoverybench-decision}
\begin{tabular*}{\linewidth}{@{\extracolsep{\fill}}lrrr@{}}
\toprule
Policy & Estimate & Lower & Upper\\
\midrule
Agreement-guided, $k=1$ & 0.013 & 0.000 & 0.028\\
Agreement-guided, $k=2$ & 0.030 & 0.003 & 0.066\\
Outcome-mean direct & 0.065 & 0.000 & 0.171\\
Fixed verify, Luna & 0.000 & 0.000 & 0.000\\
Fixed verify, Terra & 0.021 & 0.000 & 0.063\\
Fixed verify, Sol & 0.146 & 0.000 & 0.313\\
\bottomrule
\end{tabular*}
\end{table}

\noindent\textbf{Execution provenance.}
The official reward is available for every merged row, while the execution status records how that reward was obtained. Native verifier observations account for 364 episodes and contain 46 positive rewards; model-protocol failures account for 487 episodes and contain one positive reward; invalid adapter attempts account for 13 episodes and contain no positive reward. These categories are provenance fields rather than additional statistical units. The primary table follows the declared reward contract, and the status breakdown below keeps semantic outcome evidence separate from execution-protocol evidence.

\begin{table}[ht]
\centering\small
\caption{Execution provenance for all 864 RecoveryBench episodes. Positive rate is the number of reward-positive official receipts divided by the number of episodes in the row.}
\label{tab:recoverybench-status}
\begin{tabular*}{\linewidth}{@{\extracolsep{\fill}}lrrr@{}}
\toprule
Execution status & Episodes & Reward-positive & Positive rate\\
\midrule
Native verifier observed & 364 & 46 & 0.126\\
Model protocol failure & 487 & 1 & 0.002\\
Invalid adapter attempt & 13 & 0 & 0.000\\
All episodes & 864 & 47 & 0.054\\
\bottomrule
\end{tabular*}
\end{table}

\noindent\textbf{Rank concordance.}
We correlate checkpoint-level agreement with held-out success for each model and budget. Spearman estimates are negative in all nine cells; Kendall's $\tau_b$ has the same sign. The table reports the Spearman intervals.

\begin{table}[H]
\centering\small
\caption{Checkpoint-level rank concordance between exact-set agreement and held-out success on the 12 RecoveryBench checkpoints. Each entry is Estimate, Lower, or Upper from 10,000 checkpoint-bootstrap resamples; resamples with a constant vector are omitted.}
\label{tab:recoverybench-rank}
\begin{tabular*}{\linewidth}{@{\extracolsep{\fill}}lrrrr@{}}
\toprule
Model & $k$ & Estimate & Lower & Upper\\
\midrule
Luna & 1 & -0.443 & -1.000 & -0.094\\
Luna & 2 & -0.443 & -1.000 & -0.094\\
Luna & 4 & -0.632 & -1.000 & -0.357\\
Terra & 1 & -0.783 & -1.000 & -0.283\\
Terra & 2 & -0.741 & -1.000 & -0.266\\
Terra & 4 & -0.529 & -1.000 & -0.255\\
Sol & 1 & -0.796 & -0.936 & -0.535\\
Sol & 2 & -0.776 & -0.916 & -0.535\\
Sol & 4 & -0.376 & -0.693 & -0.174\\
\bottomrule
\end{tabular*}
\end{table}

\end{document}